\documentclass[12pt]{article}

\usepackage[T1]{fontenc}
\usepackage{booktabs}
\usepackage{amsfonts}
\usepackage{graphicx}%
\usepackage{url}
\usepackage{xcolor}%
\usepackage{amsmath, amsthm}%
\usepackage[round]{natbib}
\usepackage{newtxtext}%
\usepackage[smallerops]{newtxmath}%

\usepackage{microtype}%
\usepackage{algorithm}%
\usepackage[noend]{algcompatible}%

\usepackage{lastpage}%

\newtheorem{theorem}{Theorem}%
\newtheorem{lemma}[theorem]{Lemma}%
\newtheorem{proposition}[theorem]{Proposition}%

\DeclareMathOperator*{\argmin}{argmin}%
\newcommand{\bs}{\boldsymbol}%
\allowdisplaybreaks[4]%

\newcommand{\blind}{0}

\begin{document}

\def\spacingset#1{\renewcommand{\baselinestretch}%
{#1}\small\normalsize} \spacingset{1}

\if0\blind%
{
  \title{\bf fastkqr: A Fast Algorithm for Kernel Quantile Regression}
  \author{Qian Tang  \\
       Department of Statistics and Actuarial Science, 
       University of Iowa\\[4pt]
       Yuwen Gu \\
       Department of Statistics, 
       University of Connecticut \\[4pt]
    Boxiang Wang\thanks{The corresponding author. \\
    To appear on \textit{Journal of Computational and Graphical Science.}} \\
    Department of Statistics and Actuarial Science,
       University of Iowa}
  \date{}
  \maketitle
} \fi

\if1\blind%
{
  \bigskip
  \bigskip
  \bigskip
  \begin{center}
    {\LARGE\bf fastkqr: A Fast Algorithm for Kernel Quantile Regression}
\end{center}
  \medskip
} \fi

\bigskip
\begin{abstract}
Quantile regression is a powerful tool for robust and heterogeneous learning that has seen applications in a diverse range of applied areas. However, its broader application is often hindered by the substantial computational demands arising from the non-smooth quantile loss function. In this paper, we introduce a novel algorithm named \texttt{fastkqr}, which significantly advances the computation of quantile regression in reproducing kernel Hilbert spaces. The core of \texttt{fastkqr} is a finite smoothing algorithm that magically produces exact regression quantiles, rather than approximations. To further accelerate the algorithm, we equip \texttt{fastkqr} with a novel spectral technique that carefully reuses matrix computations. In addition, we extend \texttt{fastkqr} to accommodate a flexible kernel quantile regression with a data-driven crossing penalty, addressing the interpretability challenges of crossing quantile curves at multiple levels. We have implemented \texttt{fastkqr} in a publicly available R package on CRAN. Extensive simulations and real applications show that \texttt{fastkqr} matches the accuracy of state-of-the-art algorithms but can operate up to an order of magnitude faster.
\end{abstract}

\noindent%
 \vfill

\newpage
\spacingset{1.75} 
\section{Introduction}\label{sec:intro}
Quantile regression \citep{KB78} is a popular tool in statistics and
econometrics. The method extends median regression from fitting the conditional
median to modeling a suite of conditional quantile functions, providing a more
comprehensive and nuanced view of the relationship between a response variable
and its predictors. One of the key advantages of quantile regression, also
rooted in median regression, is its robustness against outliers in the response
direction. Since its introduction, quantile regression has been adapted in
various research areas, including survival analysis
\citep{peng2008survival,wang2009locally}, longitudinal data modeling
\citep{koenker2004quantile}, machine learning
\citep{meinshausen2006quantile,fakoor2023flexible}, and so on, and has seen
widespread applications in fields such as finance, ecology, healthcare, and
engineering. For detailed introductions and the latest developments in quantile
regression, see \citet{koenker2017quantile} and \citet{koenker2018handbook}.




Despite its popularity, one primary limitation of quantile regression is its
high computational cost, which is also inherited from its median regression
origins. This computational burden is mainly due to its non-smooth loss
function. To address this, linear quantile regression is often formulated as a
linear program and solved using the simplex method \citep{koenker94} or the
interior point algorithm \citep{portnoyandkoenker1997}. However, computation
becomes more challenging when it comes to kernel quantile regression
\citep[KQR,][]{TS06, LL07}, the method that is essential for estimating
non-linear conditional quantile functions. KQR is typically solved using the
interior point method, which has been implemented in the state-of-the-art R
package \texttt{kernlab} \citep{AA04}, but in principle, the algorithm only
provides approximate solutions to the original problem. A seminal work finding
the exact solution of linear quantile regression was developed by
\citet{chen07}. However, Chen's algorithm works only for linear quantile
regression, and extending it to KQR is not straightforward. Alternatively, one
can consider the least angle regression (LARS) algorithm for computing the exact
solution paths of KQR~\citep{hastie2004, LL07, takeuchi2009}, but it is
empirically not as fast as \texttt{kernlab}. Recently, the kernel convolution
technique has been used to smooth the quantile regression to efficiently find
approximate solutions \citep{fernandes2021smoothing, tan2022high,
  he2023smoothed}.

The first main contribution of this work is the development of a fast algorithm
called \texttt{fastkqr} to alleviate the computational burden of KQR.\ Our core
strategy involves smoothing the original problem and recovering the exact
solution by leveraging some unique properties of the quantile loss. To solve the
smoothed problem, we introduce a novel spectral technique that builds upon the
accelerated proximal gradient descent algorithm. With this technique, the
algorithm operates at a complexity of only $\mathcal{O}(n^{2})$ after an initial
eigen-decomposition of the kernel matrix. This efficient implementation makes
our algorithm scalable for the KQR computation that involves numerous tuning
parameters with different quantile levels.

\begin{figure}[p]
  \centering
  \includegraphics[width=0.9\textwidth]{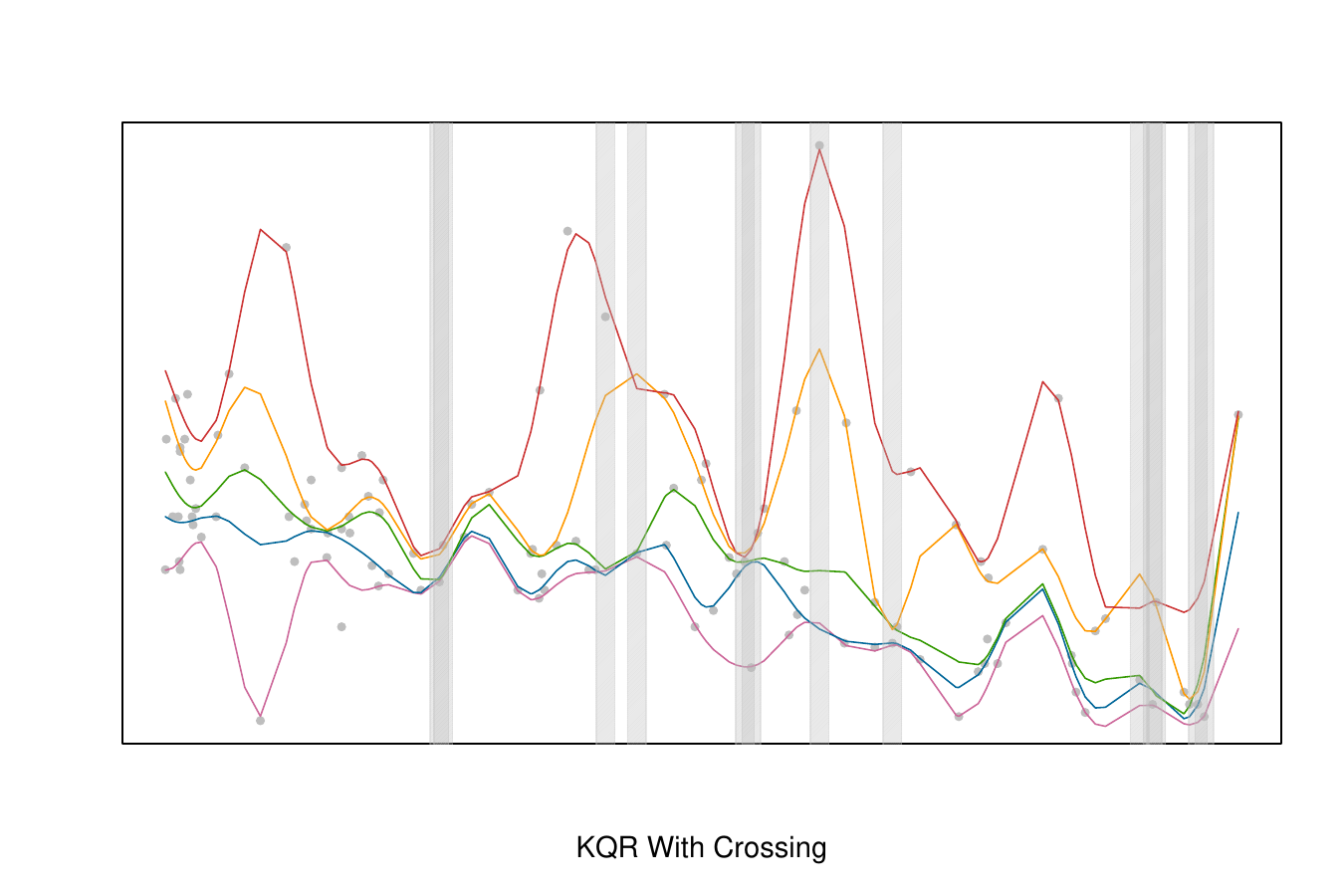}
  \vspace{-0.7cm}

  \includegraphics[width=0.9\textwidth]{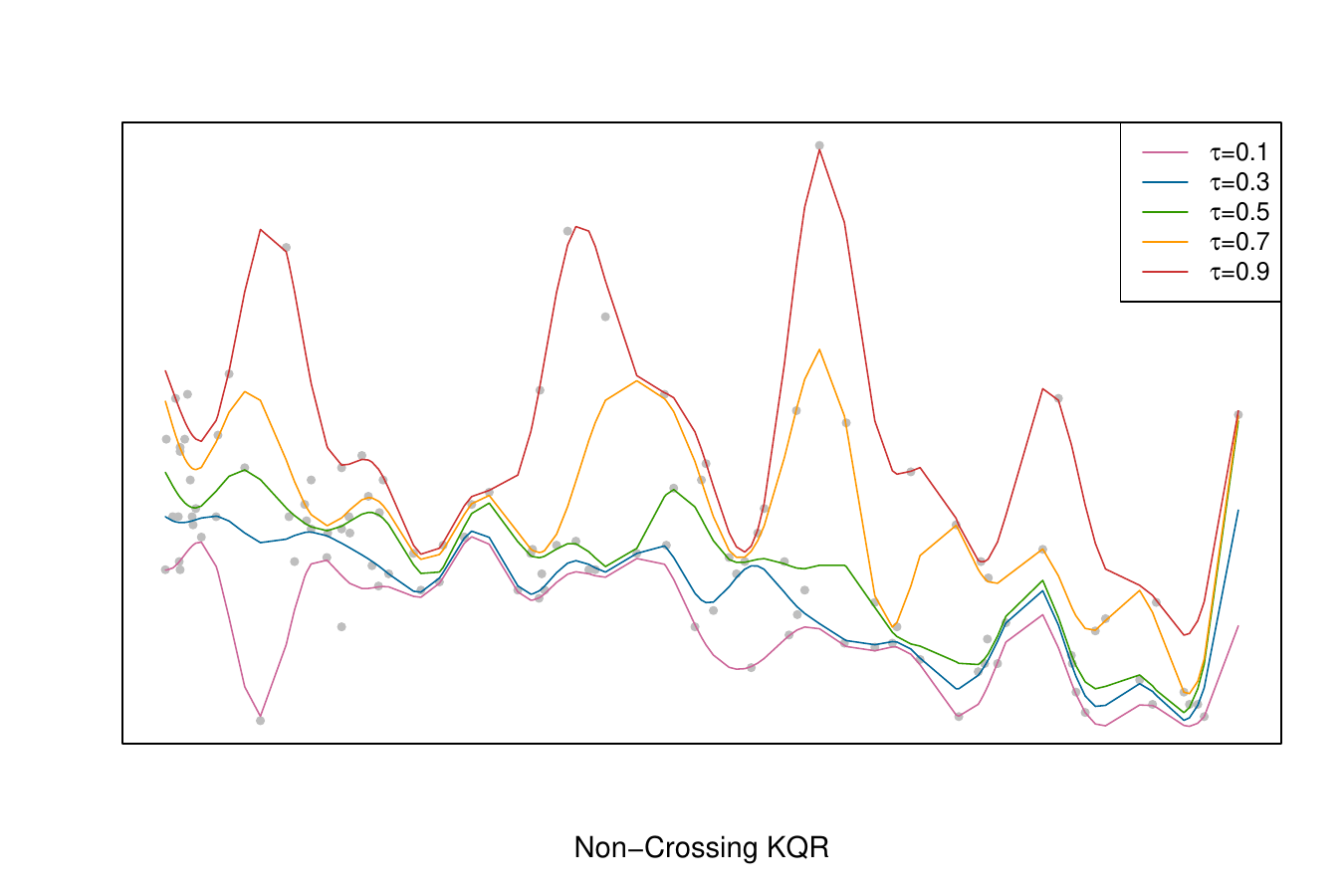}
  \caption{Fitting quantile curves at various levels using the \texttt{GAGurine}
    data. In both panels, gray dots represent the data points. The quantile curves
    are fitted at levels $\tau = 0.1, 0.3, 0.5, 0.7$, and $0.9$.
    The top panel shows quantile curves fitted individually at each level and
    the gray bands highlight the zones where crossings occur. The bottom panel
    displays the quantile curves fitted simultaneously using the NCKQR model,
    where no crossings are present.}\label{fig:GAGurine}
\end{figure}

In addition to the computational challenges mentioned above, quantile regression
also poses a notable interpretability difficulty, which arises when multiple
quantile functions estimated at different levels cross each other
\citep{cole1988fitting, He97}. This is a situation commonly encountered in
practice due to finite data samples. The issue can be exemplified in a benchmark
data set \texttt{GAGurine} from the R package \texttt{MASS}
\citep{venables2013modern}. This data set records the concentration of urinary
glycosaminoglycans (GAGs) for 314 children aged 0 to 17 years, with the age of
the children as the predictor. As depicted in the top panel of
Figure~\ref{fig:GAGurine}, five quantile curves are fitted at various levels,
with crossings highlighted by gray zones where they occur. Several strategies
have been proposed in the literature to address the crossing issue; examples
include location-shift modeling \citep{He97}, heteroscedastic location-scale
modeling \citep{shim2009non}, rearranging \citep{chernozhukov2010}, joint
estimation \citep{sangnier2016joint}, deep learning \citep{brando2022deep,
  shen2024nonparametric}, and imposition of non-crossing constraints
\citep{TS06, bondell2010noncrossing, liu2011simultaneous}, among others.

Our second major contribution addresses the crossing issue in KQR by enhancing
the capabilities of \texttt{fastkqr} to produce non-crossing kernel quantile
regression (NCKQR). Following the approach of a non-crossing constraint
\citep{TS06, bondell2010noncrossing, liu2011simultaneous}, we adopt a soft
non-crossing penalty. This soft penalty not only eases computation but also adds
flexibility, allowing the degree of crossing to be adjusted according to
specific scenario tolerances. To manage the increased computational load, we fit
NCKQR using a specialized majorization-minimization algorithm featuring two
majorization steps. The resulting NCKQR algorithm efficiently tackles the
non-crossing penalty, maintaining the same level of computational complexity as
fitting a single-level KQR.\ To give a quick demonstration, the bottom panel of
Figure~\ref{fig:GAGurine} depicts the five quantile curves fitted using NCKQR,
showing no crossings.

Our numerical studies demonstrate that \texttt{fastkqr} can be significantly
faster than the state-of-the-art solver, \texttt{kernlab}. For example, we fit
KQR using \texttt{fastkqr} and \texttt{kernlab} on simulated data considered
by~\cite{Yuan2006}. With a sample size $n = 1000$, \texttt{fastkqr} completed
the computation in 30 seconds. In contrast, \texttt{kernlab}, while maintaining
comparable accuracy with \texttt{fastkqr}, took about 400 seconds. We have
implemented our algorithms in an R package \texttt{fastkqr}, which is publicly
available on the Comprehensive R Archive Network (CRAN) at
\url{https://CRAN.R-project.org/package=fastkqr}.

The rest of the paper is organized as follows. In
Section~\ref{sec:quikquan_algo}, we review the KQR and introduce the
\texttt{fastkqr} algorithm to solve it. In Section~\ref{sec:nckqr}, we study the
non-crossing KQR and extend \texttt{fastkqr} to efficiently solve the
non-crossing problem. In Section~\ref{sec:simu}, we present extensive numerical
studies to demonstrate the competitive performance of our proposed algorithms.
All technical proofs and additional experiments are provided in the online
supplementary material.

\section{Fast Finite Smoothing Algorithm}\label{sec:quikquan_algo}

In this section, we propose our \texttt{fastkqr} algorithm. We begin with a
smooth surrogate for the quantile loss function in Section~\ref{sec:finite-kqr}
and develop the exact finite smoothing algorithm in Section~\ref{sec:exact-kqr}.
We propose an accelerated proximal gradient descent algorithm to solve the
smoothed problem in Section~\ref{sec:algo}, which is further accelerated by a
fast spectral technique introduced in Section~\ref{sec:fast-kqr}.

\subsection{A Smooth surrogate for kernel quantile regression}\label{sec:finite-kqr}

Given $\tau\in(0,1)$, KQR estimates the $\tau$th quantile function of a
response $y \in \mathbb{R}$ with covariates $\mathbf{x} \in \mathbb{R}^p$ through
\begin{equation}\label{eq:original_RKHS}
\min_{b\in\mathbb{R},f\in\mathcal{H}_{K}}
\frac{1}{n}\sum_{i=1}^{n}\rho_{\tau}(y_{i}-b-f(\mathbf{x}_{i}))+\frac{\lambda}{2}
\Vert{}f\Vert_{\mathcal{H}_{K}}^{2},
\end{equation}
where $\rho_{\tau}(t)=t(\tau-I(t<0))$ is the quantile check loss,
$\mathcal{H}_{K}$ is the reproducing kernel Hilbert space (RKHS) induced by a
kernel function $K$, and $\lambda$ is a tuning parameter governing the model
complexity. A commonly used kernel function is the radial basis kernel,
$K\left(\mathbf{x},
  \mathbf{x}^{\prime}\right)=\exp(-\frac{\|\mathbf{x}-\mathbf{x}^{\prime}\|^{2}}{2
  \sigma^{2}})$, where $\sigma$ is the kernel bandwidth parameter. By the
representer theorem \citep{Wahba90}, $\hat{f}$ has a finite-dimensional
representation in terms of the kernel function, that is, $\hat{f}(\mathbf{x})=
\sum_{i=1}^{n}\hat{\alpha}_{i}K \left(\mathbf{x}_{i},\mathbf{x}\right)$. Thus,
the estimation of the function in problem~\eqref{eq:original_RKHS} can be
transformed into a finite-dimensional optimization problem,
\begin{equation}\label{eq:representer}
\begin{aligned}
  (\hat{b},\hat{\bs\alpha})
=\argmin_{b\in\mathbb{R}, \bs{\alpha}\in\mathbb{R}^{n}}G(b, \bs{\alpha}),
\end{aligned}
\end{equation}
where
$\hat{\bs{\alpha}}={(\hat{\alpha}_{1},\hat{\alpha}_{2},\ldots,\hat{\alpha}_n)}^{\top}$,
\[
  G(b, \bs{\alpha}) = \frac{1}{n}
  \sum_{i=1}^{n}\rho_{\tau}(y_{i}-b-\mathbf{K}_{i}^{\top}\bs{\alpha})
  +\frac{\lambda}{2} \bs{\alpha}^{\top}\mathbf{K}\bs{\alpha},
\]
and each $\mathbf{K}_{i}^{\top}$ is the $i$th row of $\mathbf{K}$, the
$n\times{}n$ symmetric kernel matrix whose $(i,j)$th entry is
$K(\mathbf{x}_{i},\mathbf{x}_{j})$.

Solving problem~\eqref{eq:representer} is challenging primarily because the
check loss function is non-smooth. To efficiently compute KQR, we propose to
first smooth the original problem. Given $\tau \in (0, 1)$, we consider a
$\gamma$-smoothed check loss function,
\begin{equation}\label{eq:huber-check}
  H_{\gamma,\tau}(t)=
  \begin{cases}
    (\tau-1)t
    & \text{if}\enskip{}t<-\gamma,\\
    \frac{t^{2}}{4\gamma}+t(\tau-\frac{1}{2})+\frac{\gamma}{4}
    & \text{if}\enskip{}-\gamma\leq t\leq\gamma,\\
    \tau{}t
    & \text{if}\enskip{}t>\gamma,
  \end{cases}
\end{equation}
where $\gamma>0$ is the smoothing parameter. Similar smoothing strategies for support vector machine have previously been employed in \citet{wang2022fast}. A plot of the function \(H_{\gamma,\tau}(t)\) is provided in Figure S1 in the online supplement. 

We then introduce the following
smooth surrogate of the objective function in problem~\eqref{eq:representer}:
\begin{equation}\label{eq:smoothed_kqr_obj}
G^{\gamma}(b,\bs{\alpha}) = \frac1n
  \sum_{i=1}^{n}H_{\gamma,\tau}(y_{i}-b-\mathbf{K}_{i}^{\top}\bs{\alpha})
  +\frac{\lambda}{2}\bs{\alpha}^{\top}\mathbf{K}\bs{\alpha}.
\end{equation}

Note that the above $\gamma$-smoothed function matches the original check loss
function in the regions where $t < -\gamma$ and $t > \gamma$, while in the
intermediate region $-\gamma \leq t \leq \gamma$, it is smoothed using a
quadratic function to ensure continuity. This smoothing approach is natural and
simple and thus has been chosen in our implementation. Note that the
$\gamma$-smoothed function is not the only option. In
Section~\ref{sec:alternative}, we shall discuss other smoothing approaches.


\subsection{An exact finite smoothing algorithm}\label{sec:exact-kqr}
We now elaborate on how the exact KQR solution of
problem~\eqref{eq:representer}, $(\hat{b},\hat{\bs{\alpha}})
=\argmin_{b\in\mathbb{R},\bs{\alpha}\in\mathbb{R}^{n}}G(b,\bs{\alpha})$, can be
obtained based on our smoothing surrogate.

We first define
$S_{0}=\{i\colon{}y_{i}=\hat{b}+\mathbf{K}_{i}^{\top}\hat{\bs{\alpha}}\}
\subseteq \{1, 2, \ldots, n\}$ to be the \textit{singular set} of
problem~\eqref{eq:representer}. Note that if the singular set
$S_0$ is known, the following proposition shows that the exact KQR solution can
be obtained from a smooth optimization problem with simple linear constraints.


\begin{proposition}
\label{thm:S0given}
Suppose $S_0$, the singular set of problem~\eqref{eq:representer}, is known. Define
\begin{equation*}
  \begin{split}
    &(\hat{b}^{\gamma},\hat{\bs{\alpha}}^{\gamma})=
      \argmin_{b\in\mathbb{R}, \bs{\alpha}\in\mathbb{R}^{n}}
      G^{\gamma}(b,\bs{\alpha}), \enskip{}\text{subject to }
      y_{i}=b+\mathbf{K}_{i}^{\top}\bs{\alpha},\,i\in{}S_0.
  \end{split}
\end{equation*}
Then, $(\hat{b}^{\gamma},\hat{\bs{\alpha}}^{\gamma})
=(\hat{b},\hat{\bs{\alpha}})$ holds, where $(\hat{b},\hat{\bs{\alpha}})$ is the
solution to problem~\eqref{eq:representer}.
\end{proposition}

Although Proposition~\ref{thm:S0given} suggests that the exact KQR problem can
be efficiently solved through a smoothed optimization problem, it is not
practically useful since the singular set $S_0$ is unknown before
$(\hat{b},\hat{\bs{\alpha}})$ is actually obtained. To address this issue, we
present \textit{a set expansion method}. Specifically, for a given $\gamma > 0$,
we use a set $S\subseteq{}{}S_{0}$ as input and solve the following problem,
\begin{equation*}
  (\tilde{b}^{\gamma},\tilde{\bs{\alpha}}^{\gamma})
  =\argmin_{b\in\mathbb{R},\bs{\alpha}\in\mathbb{R}^{n}}
  G^{\gamma}(b,\bs{\alpha}), \enskip{}\text{subject to }
  y_{i}=b+\mathbf{K}_{i}^{\top}\bs{\alpha},\, i\in{}S.
\end{equation*}
Based on solution $(\tilde{b}^{\gamma},\tilde{\bs{\alpha}}^{\gamma})$, the set
expansion method outputs a set
\[
  \mathcal{E}(S) \equiv \{i\colon -\gamma\leq{}y_{i}
  -\tilde{b}^{\gamma}-\mathbf{K}_{i}^{\top}\tilde{\bs{\alpha}}^{\gamma}
  \leq\gamma\}.
\]

Define $\gamma_{0}=\min_{i\notin{}S_0}|y_{i}-\hat{b}-\mathbf{K}_{i}^{\top}
\hat{\bs{\alpha}}|$, \(D_{\gamma_{0}/2}=\{(b,\bs{\alpha})\colon\Vert{}
b\mathbf{1}_{n}+\mathbf{K}\bs{\alpha}-\hat{b}\mathbf{1}_{n}
-\mathbf{K}\hat{\bs{\alpha}}\Vert_{\infty}\geq\gamma_{0}/2\}\),
$\rho=\inf_{(b,\bs{\alpha})\in{}D_{\gamma_0/2}}
[G(b,\bs{\alpha})-G(\hat{b},\hat{\bs{\alpha}})]>0$,
and $\gamma^{*}=\min\{\gamma_{0}/2,4\rho\}$. The following theorem shows that
the output set $\mathcal{E}(S)$ is bounded between the input $S$ and the
singular set $S_0$.

\begin{theorem}\label{thm:update_set}
  For any set $S\subseteq{}S_{0}$ and $\gamma<\gamma^{*}$, if the set
  expansion method outputs a set $\mathcal{E}(S)$ from $S$, then we have
  $S\subseteq \mathcal{E}(S)\subseteq{}S_{0}$.
\end{theorem}

Theorem~\ref{thm:update_set} establishes that, by the set expansion method, any
set $S\subseteq{}{}S_{0}$ will either remain unchanged or expand toward the
singular set $S_0$. Therefore, we can initiate the process with the empty set
$\hat{S} = \emptyset \subseteq S_{0}$ and continue expanding $\hat{S}$ by
iteratively applying the set expansion method until $\hat{S}$ remains unchanged,
i.e., $\hat{S} = \mathcal{E}(\hat{S})$. Since the sample size is finite,
$\hat{S}$ becomes unchanged in finite steps. We therefore name the whole
procedure \textit{the finite smoothing algorithm}.

The following theorem indicates that the exact KQR solution can be obtained
based on $\hat{S}$. The uniqueness of the KQR solution naturally gives $\hat{S}
= S_0$. Hence, the singular set $S_0$ can be constructed using the finite
smoothing algorithm.

\begin{theorem}
  For any $\gamma\in(0,\gamma^{*})$, if there exists a set
  $\hat{S}\subseteq{}S_{0}$ such that $\mathcal{E}(\hat{S}) = \hat{S}$, then
  $(\hat{b}^{\gamma},\hat{\bs{\alpha}}^{\gamma})=(\hat{b},\hat{\bs{\alpha}})$
  holds, where
  \begin{equation}\label{eq:smoothed-kqr}
    \begin{split}
      &(\hat{b}^{\gamma},\hat{\bs{\alpha}}^{\gamma}) =\argmin_{b\in\mathbb{R},\bs{\alpha}\in\mathbb{R}^{n}}
        G^{\gamma}(b,\bs{\alpha}), \enskip{}\text{subject to } y_{i}=b+\mathbf{K}_{i}^{\top}\bs{\alpha},\,
        i\in{}\hat{S},
    \end{split}
  \end{equation}
  and $(\hat{b},\hat{\bs{\alpha}})$ is the solution to
  problem~\eqref{eq:representer}.
  \label{thm:subset}
\end{theorem}

Therefore, with the finite smoothing algorithm, the exact solution of
problem~\eqref{eq:representer} can be obtained by iteratively solving
problem~\eqref{eq:smoothed-kqr} and augmenting the set $\hat{S}$. Since
$\gamma^{*}$ is still unknown in practice, we handle this by repeatedly
implementing the above procedure with a decreasing sequence of values of
$\gamma$. The algorithm is terminated once a solution satisfies the
Karush–Kuhn–Tucker (KKT) conditions of problem~\eqref{eq:representer}. In our
implementation, we initiate this process with $\gamma = 1$ and iteratively
update it by reducing $\gamma$ to a quarter of its previous value, i.e.,
$\gamma \leftarrow \gamma / 4$. We observe that this approach typically
converges within only three or four iterations of updating $\gamma$.

\subsection{Solving the smoothed kernel quantile regression}\label{sec:algo}
In this section, we develop an accelerated proximal gradient descent (APGD)
algorithm to solve problem~\eqref{eq:smoothed-kqr}. We first consider the
unconstrained optimization, say, problem~\eqref{eq:smoothed-kqr} with $\hat{S} =
\emptyset$.

First, note that $H_{\gamma,\tau}'(t)$ is Lipschitz continuous, that is,
\[
  |H_{\gamma,\tau}'(c_{1})-H_{\gamma,\tau}'(c_{2})|\leq
  \frac{1}{2\gamma}|c_{1}-c_{2}|, \ \forall c_1, c_2 \in \mathbb{R}.
\]
Let $(b^{(1)},\bs{\alpha}^{(1)})$ be the initial value of \((b,\bs{\alpha})\).
For each $k=1,2,\ldots$, the proximal gradient method updates
$(b^{(k+1)},\bs{\alpha}^{(k+1)})$ by the majorization-minimization principle
\citep{hunter00_quant_regres_via_mm_algor},
\begin{equation}\label{eq:update0}
  \begin{split}
    \left(
    \begin{array}{c}
      b^{(k+1)} \\
      \bs{\alpha}^{(k+1)}
    \end{array}
    \right)
    =\argmin_{b\in\mathbb{R},\bs{\alpha}\in\mathbb{R}^{n}}
    &\frac\lambda2\bs{\alpha}^{\top}\mathbf{K}\bs{\alpha}
      +\frac1n\sum_{i=1}^{n}H_{\gamma,\tau}(y_{i}-b^{(k)}
      -\mathbf{K}_{i}^{\top}\bs{\alpha}^{(k)})\\
    &+\frac1n\sum_{i=1}^{n}H_{\gamma,\tau}'
      (y_{i}-b^{(k)}-\mathbf{K}_{i}^{\top}\bs{\alpha}^{(k)})
      (b^{(k)}+\mathbf{K}_{i}^{\top}\bs{\alpha}^{(k)}
      -b-\mathbf{K}_{i}^{\top}\bs{\alpha})\\
    &+\frac{1}{4n\gamma}
      \lVert{}b^{(k)}\mathbf{1}+\mathbf{K}\bs{\alpha}^{(k)}
      -b\mathbf{1}-\mathbf{K}\bs{\alpha}\rVert_{2}^{2}\\
    =\argmin_{b\in\mathbb{R},\bs{\alpha}\in\mathbb{R}^{n}}
    &\frac{1}{4n\gamma}\left\Vert\mathbf{K}\bs{\alpha}+b\mathbf{1}
      -(\mathbf{K}\bs{\alpha}^{(k)}+b^{(k)}\mathbf{1}
      {}+{}\gamma\mathbf{z}^{(k)})
      \right\Vert_{2}^{2}+\frac{\lambda}{2}\alpha^{\top}\mathbf{K}\alpha,
  \end{split}
\end{equation}
where $\mathbf{z}^{(k)}$ is an $n$-vector whose $i$th element is
$H'_{\gamma,\tau}(y_{i}-\mathbf{K}_{i}^{\top}\bs{\alpha}^{(k)}-b^{(k)})$. Then,
we have
\begin{equation}\label{eq:algo-upd0}
  \left(
    \begin{array}{c}
      b^{(k+1)}\\
      \bs{\alpha}^{(k+1)}
    \end{array}
  \right)=
  \left(
    \begin{array}{c}
      b^{(k)} \\
      \bs{\alpha}^{(k)}
    \end{array}
  \right)
  +\gamma\mathbf{P}_{\gamma,\lambda}^{-1}\bs\zeta,
\end{equation}
where
\begin{equation*}
  \bs{\zeta}=\left(
    \begin{array}{c}
      \mathbf{1}^{\top}\mathbf{z}^{(k)}\\
      \mathbf{K}^{\top}\mathbf{z}^{(k)}-n\lambda\mathbf{K}\bs{\alpha}^{(k)}
    \end{array}\right),
 \mathbf{P}_{\gamma,\lambda}=\left(
    \begin{array}{cc}
      n & \mathbf{1}^{\top}\mathbf{K}\\
      \mathbf{K}^{\top}\mathbf{1}
        & \mathbf{K}^{\top}\mathbf{K}+n\gamma\lambda\mathbf{K}
    \end{array}\right).
\end{equation*}

We further apply Nesterov's acceleration~\citep{nesterov1983method,beck2009fast}
to boost the algorithm. Given a sequence $\{c_k\}_{k\geq1}$, such that $c_{1}=1$
and $c_{k+1}=1/2+(1+4c_{k}^{2})^{1/2}/2$ for $k\geq1$, let
$(b^{(0)},{\bs{\alpha}^{(0)}})$ and $(b^{(1)},{\bs{\alpha}^{(1)}})$ be the first
two iterates. For each $k=1,2, \ldots$, we solve
$(b^{(k+1)},\bs{\alpha}^{(k+1)})$ from the following problem
\begin{equation}\label{eq:APGD_updt}
  \begin{split}
    \left(
    \begin{array}{c}
      b^{(k+1)}\\
      \bs{\alpha}^{(k+1)}
    \end{array}
    \right)
    &=\argmin_{b\in\mathbb{R},\bs{\alpha}\in\mathbb{R}^{n}}
      \frac{1}{4n\gamma}
      \left\|b\mathbf{1}+\mathbf{K}\bs{\alpha}-(\bar{b}^{(k)}\mathbf{1}+\mathbf{K}\bs{\bar\alpha}^{(k)}
      +2\gamma\bar{\mathbf{z}}^{(k)})\right\|_{2}^{2} + \frac{\lambda}{2}\bs{\alpha}^{\top}\mathbf{K}\bs{\alpha}\\
    &=\left(
      \begin{array}{c}
        \bar{b}^{(k)}\\
        \bar{\bs{\alpha}}^{(k)}
      \end{array}
      \right)+2\gamma\mathbf{P}_{\gamma,\lambda}^{-1}\bar{\bs{\zeta}},
  \end{split}
\end{equation}
where
\begin{equation*}
  \left(
    \begin{array}{c}
      \bar b^{(k)}\\
      \bs{\bar \alpha}^{(k)}
    \end{array}\right)
  =\left(
    \begin{array}{c}
      b^{(k)}\\
      \bs{\alpha}^{(k)}
    \end{array}
  \right)+\left(\frac{c_{k}-1}{c_{k+1}}\right)\left(
    \begin{array}{c}
      b^{(k)}-b^{(k-1)} \\
      \bs{\alpha}^{(k)}-\bs{\alpha}^{(k-1)}
    \end{array}\right)
\end{equation*}
and
\begin{equation*}
  \bar{\bs{\zeta}}=
  \left(
    \begin{array}{c}
      \mathbf{1}^{\top}\bar{\mathbf{z}}^{(k)}\\
      \mathbf{K}^{\top}\bar{\mathbf{z}}^{(k)}
      -n\lambda\mathbf{K}\bar{\bs{\alpha}}^{(k)}
    \end{array}
  \right)\enskip\text{with}\enskip{}
  \bar{z}_{i}^{(k)}=H_{\gamma,\tau}'(y_{i}
  -\bar{b}^{(k)}-\mathbf{K}_{i}^{\top}\bs{\bar \alpha}^{(k)}),
  \,i=1,\ldots,n.
\end{equation*}
The standard theory of the APGD algorithm gives the following convergence analysis.
\begin{proposition}
  Suppose $G^{\gamma}(b, \alpha)$ is defined in
  equation~\eqref{eq:smoothed_kqr_obj}, $(b^{*}, \bs\alpha^{*})$ is the global
  minimizer, and $(b^{(k)}, \bs\alpha^{(k)})$ is the solution at the $k$th
  iteration of the APGD algorithm. It holds that
  \[
    G^{\gamma}(b^{(k)}, \bs\alpha^{(k)}) - G^{\gamma}(b^{*}, \bs\alpha^{*}) \le
    \dfrac{1}{\gamma k^2} \left((b^{(0)} - b^{*})^2 + \Vert\mathbf{K}
      (\bs\alpha^{(0)} - \bs\alpha^{*}) \Vert_2^2\right).
  \]
  \label{thm:conv_pgd}
\end{proposition}

We now address the constraint in problem~\eqref{eq:smoothed-kqr}, for which we
consider the projected gradient descent algorithm. Specifically, after obtaining
$(b^{(k)}, \bs{\alpha}^{(k)})$ for each $k$, we project the solution onto the
feasible region associated with the constraint by solving the following
optimization problem,
\begin{equation}\label{eq:proj}
  \begin{aligned}
    &(\tilde{b}, \tilde{\bs\alpha})
      = \argmin_{b\in\mathbb{R},\bs{\alpha}\in\mathbb{R}^{n}} (b - b^{(k)})^2
      + \Vert \mathbf{K} (\bs\alpha - \bs\alpha^{(k)}) \Vert_2^2, \enskip{}\text{subject to } y_i = b + \mathbf{K}_i^\top \bs\alpha, i \in \hat{S}.
  \end{aligned}
\end{equation}
It can be shown that the solution is
$\tilde{b} = b^{(k)} + \frac{1}{|\hat{S}|+1}\sum_{i \in \hat{S}}[ y_i - \mathbf{K}_i^\top\bs\alpha^{(k)}]$
and $\tilde{\bs\alpha} = \mathbf{K}^{-1}\bs\theta$, where $|\hat{S}|$ denotes
the number of elements in the set $\hat{S}$, and $\bs\theta\in \mathbb{R}^n$
with $\theta_i = y_i -\tilde{b}$ if $i \in \hat{S}$ and
$\theta_i= \mathbf{K}_i^\top \bs\alpha^{(k)}$ otherwise. Subsequently, we use
$(\tilde{b}, \tilde{\bs\alpha})$ in place of $(b^{(k)}, \bs{\alpha}^{(k)})$ in
problem~\eqref{eq:update0} to move the APGD algorithm forward to obtain
$(b^{(k+1)}, \bs{\alpha}^{(k+1)})$. In practice, we find that the performance is
often nearly identical if the above projection is applied only once to the
unconstrained solution of problem~\eqref{eq:smoothed-kqr}. This implementation
is effective mainly because the solution of the unconstrained problem barely
violates the constraint when $\gamma$ is sufficiently small.



\subsection{A fast spectral technique}\label{sec:fast-kqr}
We note that the computational bottleneck of the APGD algorithm discussed in
Section \ref{sec:algo} is the inversion of $\mathbf{P}_{\gamma,\lambda}$, which
typically has a computational complexity of $\mathcal{O}(n^{3})$. The whole KQR
algorithm can become very expensive because the matrix inversion must be
repeated for every $\mathbf{P}_{\gamma,\lambda}$ as the smoothing parameter
$\gamma$ and the tuning parameter $\lambda$ vary. Although we could consider
alternative algorithms, such as gradient descent or quasi-Newton methods, to
circumvent matrix inversion, our empirical studies indicate that the precision
of these alternatives is generally inferior to that of the APGD algorithm,
unless they are executed for an excessive number of iterations.

To accelerate the APGD algorithm, we develop a spectral technique, which begins
with the eigen-decomposition of the kernel matrix,
$\mathbf{K}=\mathbf{U}\bs{\Lambda}\mathbf{U}^{\top}$, where $\bs{\Lambda}$ is
diagonal and $\mathbf{U}$ is orthogonal. Note that this step is free of the
parameters $\gamma$ and $\lambda$. Define
$\bs{\Pi}_{\gamma,\lambda}=\bs{\Lambda}^{2}+2n\gamma\lambda\bs{\Lambda}$,
$\mathbf{v}=\mathbf{U}\bs{\Lambda}\Pi_{\gamma,\lambda}^{-1}\mathbf{U}^{\top}\mathbf{1}$,
and
$g=1/\bigl(n\mathbf{1}^{\top}\mathbf{U}\bs{\Lambda}\Pi_{\gamma,\lambda}^{-1} \bs{\Lambda}\mathbf{U}^{\top}\mathbf{1}\bigr)$.
Using the Woodbury matrix identity, we obtain \vspace{-0.1in}
\begin{equation}\label{eq:invmat}
  \begin{split}
  \mathbf{P}_{\gamma,\lambda}^{-1}&=\left(
    \begin{array}{cc}
      n & \mathbf{1}^{\top}\mathbf{U}\bs{\Lambda}\mathbf{U}^{\top} \\
      \mathbf{U}\bs{\Lambda}\mathbf{U}^{\top}\mathbf{1}
        & \mathbf{U}\Pi_{\gamma,\lambda}\mathbf{U}^{\top}
    \end{array}\right)^{-1}
    =g\Biggl(
    \begin{array}{c}
      1 \\
      -\mathbf{v}
    \end{array}\Biggr)\Bigl(
    \begin{array}{ll}
      1 & -\mathbf{v}^{\top}
    \end{array}\Bigr)+\left(
    \begin{array}{cc}
      0 & \mathbf{0}^{\top} \\
      \mathbf{0} & \mathbf{U}\bs{\Pi}_{\lambda}^{-1}\mathbf{U}^{\top}
    \end{array}\right).
    \end{split}
\end{equation}
Although it may initially appear that the computation of
$\mathbf{P}_{\gamma,\lambda}^{-1}$ using equation~\eqref{eq:invmat} still has
the complexity of $\mathcal{O}(n^3)$, as matrix multiplications are still
involved, it is important to note that the APGD update requires only the direct
computation of $\mathbf{P}_{\gamma,\lambda}^{-1}\bs\zeta$ rather than the matrix
inversion itself. Thus equation~\eqref{eq:invmat} gives
\begin{equation}\label{eq:algo-upd}
  \mathbf{P}_{\gamma,\lambda}^{-1}\bar{\bs \zeta}
  =g\left\{
    \mathbf{1}^{\top}\bar{\mathbf{z}}-\mathbf{v}^{\top}\mathbf{K}
    \left(\bar{\mathbf{z}}+n\lambda\bs{\alpha}\right)\right\}\Biggl(
  \begin{array}{c}
    1 \\
    -\mathbf{v}
  \end{array}\Biggr)
  +\left(
    \begin{array}{c}
      0 \\
      \mathbf{U}\bs\Pi_{\gamma,\lambda}^{-1}\bs{\Lambda}
      \mathbf{U}^{\top}\left(\bar{\mathbf{z}}+n\lambda\bs{\alpha}\right)
    \end{array}\right),
\end{equation}
where $\bar{\bs \zeta}$ was defined in equation~\eqref{eq:APGD_updt}. The
computational complexity is only $\mathcal{O}\left(n^{2}\right)$ when computed
from right to left, that is, by only performing matrix-vector multiplications.

In our implementation, we use a warm-start strategy to further amplify the
effect of the spectral technique. Specifically, we solve
problem~\eqref{eq:representer} with a sequence of tuning parameters
$\lambda^{[1]}, \lambda^{[2]}, \ldots, \lambda^{[L]}$ to obtain the
corresponding solutions
$(\hat{b}^{[1]}, \hat{\bs\alpha}^{[1]}), (\hat{b}^{[2]}, \hat{\bs\alpha}^{[2]}),
\ldots, (\hat{b}^{[L]}, \hat{\bs\alpha}^{[L]})$.
With $l > 1$, $(\hat{b}^{[l-1]}, \hat{\bs\alpha}^{[l-1]})$ is summoned to
initialize the finite smoothing and APGD algorithms to solve for
$(\hat{b}^{[l]}, \hat{\bs\alpha}^{[l]})$. Therefore, thanks to both the warm
start and the spectral technique, except for the only step of
eigen-decomposition, which costs an $\mathcal{O}(n^3)$ complexity, the rest of
the entire KQR algorithm to solve problem~\eqref{eq:representer} takes only
$\mathcal{O}(n^2)$, hence the speed is significantly enhanced.

The algorithm \texttt{fastkqr} is summarized in Algorithm 1 in the online
supplemental material.

\subsection{Alternative smoothing surrogates}\label{sec:alternative}
Now that we have shown the exact KQR solution to problem~\eqref{eq:representer}
can be recovered through a smoothing surrogate, the proposed $\gamma$-smoothed
check loss is, however, not the only option that leads to the exact solution.
The following theorem provides a broader perspective.
\begin{theorem}\label{thm:S0given_general}
  Suppose $S_0$, the singular set of problem~\eqref{eq:representer}, is known.
  Define
  \begin{equation*}
    G^{\gamma}(b,\bs{\alpha}) = \frac1n
    \sum_{i=1}^{n}H_{\gamma,\tau}(y_{i}-b-\mathbf{K}_{i}^{\top}
    \bs{\alpha}) +\frac{\lambda}{2}\bs{\alpha}^{\top}\mathbf{K}\bs{\alpha},
  \end{equation*}
  where $H_{\gamma,\tau}(t)$ is a function satisfying the following constraints,
  \begin{equation}
    \begin{split}
      & 1.\enskip{} H_{\gamma, \tau}^{\prime}\left(t\right) \in \partial \rho_\tau\left(t\right) \text { if } t=0, \\
      & 2.\enskip{} H_{\gamma, \tau}^{\prime}\left(t\right) =\partial \rho_\tau\left(t\right) \text { if } t \neq 0.
    \end{split}
    \label{eq:smoothcondition}
  \end{equation}
  Let
  \begin{equation*}
    (\hat{b}^{\gamma},\hat{\bs{\alpha}}^{\gamma})=
    \argmin_{b\in\mathbb{R}, \bs{\alpha}\in\mathbb{R}^{n}}
    G^{\gamma}(b,\bs{\alpha}), \enskip{}\text{subject to }
    y_{i}=b+\mathbf{K}_{i}^{\top}\bs{\alpha},\,i\in{}S_0.
  \end{equation*}
  Then,
  $(\hat{b}^{\gamma},\hat{\bs{\alpha}}^{\gamma})=(\hat{b},\hat{\bs{\alpha}})$
  holds, where $(\hat{b},\hat{\bs{\alpha}})$ is the solution to problem (2).
\end{theorem}

Theorem~\ref{thm:S0given_general} provides a general condition for the smooth
surrogate to yield the exact solution to the original non-smooth problem, once
the singular set \(S_0\) is known. Using the same set expansion technique
introduced in Section~\ref{sec:exact-kqr}, we can identify the singular set
\(S_0\) in practice. The general condition can be satisfied by several popular
smoothing techniques, including the Moreau
envelope~\citep{moreau1965proximite,chen07}, Nesterov's
smoothing~\citep{nesterov2005smooth}, Huber
approximation~\citep{yi2017semismooth} and kernel density
convolution~\citep{tan2022high,he2023smoothed}; further details can be found in
the supplementary material. Our experiments indicate that their performance is
highly comparable, so we adopt the $\gamma$-smoothed check loss in our
implementation for simplicity.

\section{Non-crossing Kernel Quantile Regression}\label{sec:nckqr}
In this section, we propose a non-crossing kernel quantile regression (NCKQR)
method to address the crossing issue of the quantile curves fitted at various
levels.

\subsection{Methodology}
When KQR is fitted at multiple quantile levels individually, say, for $0 <
\tau_1 < \tau_2 < \cdots < \tau_T < 1$,
\begin{equation*}
  \begin{aligned}
    \min_{b\in\mathbb{R}, \bs{\alpha}\in\mathbb{R}^{n}} \frac{1}{n} \sum_{i=1}^{n} \rho_{\tau_t}
    \left(y_{i}-b_{\tau_t}-\mathbf{K}_{i}^{\top} \bs{\alpha}_{\tau_t} \right)
    + \frac{\lambda}{2} \bs\alpha_{\tau_t}^\top \mathbf{K} \bs{\alpha}_{\tau_t},
    \ t = 1, 2, \ldots, T,
  \end{aligned}
\end{equation*}
the fitted curves may cross each other. To avoid the occurrence of
crossing,~\cite{TS06},~\cite{bondell2010noncrossing},
and~\cite{liu2011simultaneous} consider fitting all the quantile curves
simultaneously with a \textit{hard} non-crossing constraint,
\begin{equation}
  \begin{aligned}
    \min_{{\{ b_{\tau_t}, \bs\alpha_{\tau_t} \}}_{t=1}^T }
    &\sum_{t=1}^{T} \left[ \frac{1}{n} \sum_{i=1}^{n} \rho_{\tau_t}
      \left(y_{i}-b_{\tau_t}-\mathbf{K}_{i}^{\top} \bs{\alpha}_{\tau_t} \right)
      + \frac{\lambda}{2} \bs\alpha_{\tau_t}^\top \mathbf{K} \bs{\alpha}_{\tau_t} \right]\\
    \text{subject to}\quad\;
    &b_{\tau_{t_1}}+\mathbf{K}_{i}^{\top}\bs{\alpha}_{\tau_{t_1}}<b_{\tau_{t_2}}
      +\mathbf{K}_{i}^{\top}\bs{\alpha}_{\tau_{t_2}}\text{ for all }
      t_1 < t_2 \text{ and }i = 1, 2, \ldots, n,
  \end{aligned}
\label{eq:NCKQR_LW11}
\end{equation}
where the notation ${\{ b_{\tau_t}, \bs\alpha_{\tau_t} \}}_{t=1}^T$ represents the
collection of $b_{\tau_1}, \bs{\alpha}_{\tau_1}, b_{\tau_2},
\bs{\alpha}_{\tau_2}, \ldots, b_{\tau_T}, \bs{\alpha}_{\tau_T}$ for ease of
presentation. In the sequel, we extend this notation to ${\{ \hat{b}_{\tau_t},
\hat{\bs\alpha}_{\tau_t} \}}_{t=1}^T$ and ${\{ \tilde{b}_{\tau_t},
\tilde{\bs\alpha}_{\tau_t} \}}_{t=1}^T$ to represent their respective
counterparts.

With the hard constraint imposed in problem~\eqref{eq:NCKQR_LW11}, the quantile
regression curves fitted on finite-sample data do not cross. However, the
inequality constraints introduced by this formulation may largely increase the
computational cost. As such, we propose using a \textit{soft} crossing penalty.
This approach does not have any inequality constraint. Moreover, it provides
practitioners with some flexibility to tolerate a certain level of crossing.
Specifically, our NCKQR is defined as
\begin{equation}\label{eq:nckqr}
  {\{ \hat{b}_{\tau_t}, \hat{\bs\alpha}_{\tau_t} \}}_{t=1}^T
  =\argmin_{ {\{b_{\tau_t}, \bs\alpha_{\tau_t} \}}_{t=1}^T }
  Q\left({\{b_{\tau_t}, \bs\alpha_{\tau_t} \}}_{t=1}^T\right),
\end{equation}
where
\begin{equation*}
  \begin{aligned}
    Q\left({\{b_{\tau_t}, \bs\alpha_{\tau_t} \}}_{t=1}^T \right) =
    &\sum_{t=1}^{T} \left[ \frac{1}{n} \sum_{i=1}^{n} \rho_{\tau_t}
      \left(y_{i}-b_{\tau_t}-\mathbf{K}_{i}^{\top} \bs{\alpha}_{\tau_t} \right)
      \right] +\sum_{t=1}^{T} \frac{\lambda_2}{2} \bs{\alpha}^{\top}_{\tau_t}
      \mathbf{K} \bs{\alpha}_{\tau_t}\\
    &+\lambda_1 \sum_{t=1}^{T-1} \left[ \sum_{i=1}^{n}V
      \left( b_{\tau_t}+\mathbf{K}_{i}^{\top} \bs{\alpha}_{\tau_t}-b_{\tau_{t+1}}
      -\mathbf{K}_{i}^{\top} \bs{\alpha}_{\tau_{t+1}}\right)\right],\\
  \end{aligned}
\end{equation*}
and $V$ is the soft crossing penalty that is non-decreasing and $V(0) = 0$. When
$\lambda_1 = 0$, NCKQR reduces to fitting KQR individually at different levels;
when $\lambda_1 \to \infty$, the soft crossing penalty functions as the hard
constraint to restrict the solution to the feasible set of
problem~\eqref{eq:NCKQR_LW11}.

A potential choice of $V$ is a rectified linear unit (ReLU), i.e.,
$V(t)=\max\{t, 0\}$, to enforce the regularization on crossing. However, the
ReLU penalty is non-smooth, which makes the computation unnecessarily
challenging. Although it seems to be a possible solution that employing the
smoothing algorithm developed in Section~\ref{sec:finite-kqr} to simultaneously
smooth both the functions $\rho_{\tau_t}$ and $V$, additional efforts are
expected to ensure the exact solution is obtained.

Yet, we may alleviate the computational burden brought by the ReLU penalty from
a different angle. The use of $\rho_\tau$ and the pursuit of an exact solution
are imperative due to the fundamental nature of quantile regression; deviating
from this would essentially lead to a different statistical problem. In
contrast, the use of a non-smooth ReLU penalty is not obligatory, as its
smoothed counterpart can function alone as a soft crossing penalty. Thus, we opt
to define NCKQR directly by employing a smooth ReLU penalty in
Problem~\eqref{eq:nckqr}, formulated as follows:
\begin{equation*}
  V(t)=
  \begin{cases}
    0 & \text{if}\enskip{}t< -\eta,\\
    \frac{t^{2}}{4\eta}+\frac{t}{2}+\frac{\eta}{4}
      & \text{if}\enskip{} -\eta\leq t\leq\eta,\\
    t & \text{if}\enskip{}t>\eta,
  \end{cases}
\end{equation*}
where $\eta=10^{-5}$ is used in our implementation. In the next section, we
shall focus on finding an algorithm to obtain the exact solution of
problem~\eqref{eq:nckqr} in which $V$ is the smooth ReLU penalty.

Note that the primary focus of this paper is on the fast computation of KQR.\
Although the soft crossing penalty could be advocated for its additional
flexibility, a theoretical investigation of this penalty is beyond the scope of
this paper. Our main rationale for proposing the soft non-crossing penalty is
due to computational considerations. Practitioners may directly use a large
$\lambda_1$ value to approximate the results that would be obtained with a hard
non-crossing constraint.

\subsection{Exact finite smoothing algorithm for NCKQR}
We now extend \texttt{fastkqr} to solve NCKQR.\ With the check loss replaced by
the $\gamma$-smoothed check loss $H_{\gamma, \tau}$, we formulate a smooth
surrogate objective function,
\begin{equation}\label{eq:Q_gamma_nckqr}
  \begin{aligned}
    Q^{\gamma}\left({\{ b_{\tau_t}, \bs\alpha_{\tau_t} \}}_{t=1}^T \right) =
    &\sum_{t=1}^{T} \left[ \frac{1}{n} \sum_{i=1}^{n} H_{\gamma, \tau_t}
      \left(y_{i}-b_{\tau_t}-\mathbf{K}_{i}^{\top} \bs{\alpha}_{\tau_t} \right)\right]
      +\sum_{t=1}^{T} \frac{\lambda_2}{2} \bs{\alpha}^{\top}_{\tau_t}
      \mathbf{K} \bs{\alpha}_{\tau_t}\\
    & + \lambda_1 \sum_{t=1}^{T-1} \left[ \sum_{i=1}^{n}
      V\left( b_{\tau_t}+\mathbf{K}_{i}^{\top} \bs{\alpha}_{\tau_t}-b_{\tau_{t+1}}
      -\mathbf{K}_{i}^{\top} \bs{\alpha}_{\tau_{t+1}}\right)\right].\\
  \end{aligned}
\end{equation}

According to ${\{\hat{b}_{\tau_t}, \hat{\bs\alpha}_{\tau_t} \}}_{t=1}^T$, the
solution of problem~\eqref{eq:nckqr}, we first construct the singular sets,
$S_{0,t} = \{i : y_{i} = \hat{b}_{\tau_t}+ \mathbf{K}_{i}^{\top}\hat{\bs{\alpha}}_{\tau_t}\} \subseteq \{1, \ldots, n \}$,
for each quantile level $\tau_1, \tau_2, \ldots, \tau_T$. In the following
proposition, we demonstrate that
${\{\hat{b}_{\tau_t}, \hat{\bs\alpha}_{\tau_t} \}}_{t=1}^T$ can be obtained by
solving the above smooth optimization problem with linear constraints associated
with the singular sets. When $T = 1$, the result reduces to
Proposition~\ref{thm:S0given} for a single-level KQR problem.

\noindent
\begin{proposition}\label{thm:givenS02}%
  {\it Suppose $S_{0,t}$ is known for each $t = 1, 2, \ldots ,T$.
    Define \begin{equation*} \begin{aligned} &{\{\hat{b}^{\gamma}_{\tau_t}, \hat{\bs{\alpha}}^{\gamma}_{\tau_t}\}}_{t=1}^T =\argmin_{{\{ b_{\tau_t}, \bs\alpha_{\tau_t} \}}_{t=1}^T} Q^{\gamma}\left({\{ b_{\tau_t}, \bs\alpha_{\tau_t} \}}_{t=1}^T\right), \enskip{}\text{subject
          to}\enskip{} y_{i}=b_{\tau_t}+\mathbf{K}_{i}^{\top}\bs{\alpha}_{\tau_t},\, \forall i\in{}S_{0,t}, 1\leq t\leq T.
   \end{aligned}
  \end{equation*}
  It holds that
  ${\{\hat{b}^{\gamma}_{\tau_t}, \hat{\bs{\alpha}}^{\gamma}_{\tau_t}\}}_{t=1}^T = {\{\hat{b}_{\tau_t}, \hat{\bs{\alpha}}_{\tau_t}\}}_{t=1}^T$,
  where ${\{\hat{b}_{\tau_t}, \hat{\bs{\alpha}}_{\tau_t}\}}_{t=1}^T$ is the
  solution to problem~\eqref{eq:nckqr}.}
\end{proposition}

Similar to Proposition~\ref{thm:S0given} for the single-level KQR,
Proposition~\ref{thm:givenS02} cannot be applied in practice, since none of the
$S_{0, t}$'s are known. We hereby develop a \textit{multiple-level set expansion
  method}. For a given $\gamma > 0$, we use a collection of sets $S_1, S_2,
\ldots, S_T$ as inputs and proceed to solve the following optimization
problem,
\begin{equation*}
  \begin{aligned}
    &{\{ \tilde{b}_{\tau_t}^{\gamma}, \tilde{\bs\alpha}_{\tau_t}^{\gamma} \}}_{t=1}^T
      =\argmin_{{\{ b_{\tau_t}, \bs\alpha_{\tau_t} \}}_{t=1}^T}
      Q^{\gamma}\left({\{ b_{\tau_t}, \bs\alpha_{\tau_t} \}}_{t=1}^T\right), \enskip{}\text{subject to}\enskip{}
      y_{i}=\mathbf{K}_{i}^{\top}\bs{\alpha}_{\tau_t}+b_{\tau_t},\,i\in{}S_t, 1\leq t\leq T.
  \end{aligned}
\end{equation*}
Then the multiple-level set expansion method outputs a collection of the
following $T$ sets,
\begin{equation*}
  \mathcal{E}_t(S_1, S_2, \ldots, S_T) \equiv \{i: -\gamma{}\leq{}
  y_{i}-\tilde{b}_{\tau_t}^{\gamma}-\mathbf{K}_{i}^{\top}
  \tilde{\bs{\alpha}}_{\tau_t}^{\gamma} \leq \gamma \}, \ t = 1, 2, \ldots, T.
\end{equation*}
To bound these output sets, some quantities are required. For each quantile
level $\tau_t$, we define
\begin{equation*}
  \begin{aligned}
    \gamma_{0, t}
    &= \min_{i\notin{}S_{0,t}} |y_{i}-\hat{b}_{\tau_t}
      - \mathbf{K}_{i}^{\top}\hat{\bs{\alpha}}_{\tau_t}|, \\
    D_{\gamma_{0}, t}
    &= \{ (b_{\tau_t}, \bs{\alpha}_{\tau_t}) \colon\Vert b_{\tau_t}\mathbf{1}_{n}
      + \mathbf{K}\bs{\alpha}_{\tau_t} -\hat{b}_{\tau_t}\mathbf{1}_{n}
      -\mathbf{K}\hat{\bs{\alpha}}_{\tau_t} \Vert_{\infty}\geq\gamma_{0,t}/2 \}.
  \end{aligned}
\end{equation*}
Denote by $\mathcal{D}$ the collection of ${\{(b_{\tau_t},
\bs{\alpha}_{\tau_t})\}}_{t=1}^T$ such that $(b_{\tau_t}, \bs{\alpha}_{\tau_t})
\in D_{\gamma_0, t}$ for each $t = 1, 2, \ldots, T$. Define
$\gamma^{*}=\min\{\frac{1}{2}\gamma_{0,1}, \frac{1}{2}\gamma_{0,2}, \ldots,
\frac{1}{2}\gamma_{0,T}, \frac{4}{T}\rho\}$, where
\begin{equation*}
  \rho = \inf_{{\{b_{\tau_t}, \bs{\alpha}_{\tau_t}\}}_{t=1}^T \in \mathcal{D}}
  Q\left({\{b_{\tau_t}, \bs{\alpha}_{\tau_t}\}}_{t=1}^T \right) -
  Q\left({\{\hat{b}_{\tau_t}, \hat{\bs{\alpha}}_{\tau_t}\}}_{t=1}^T \right) > 0.
\end{equation*}
The following theorem bounds the output sets from the multiple-level set
expansion method.

\begin{theorem}\label{thm:find-subset2}
  With some $\gamma < \gamma^{*}$, if $S_t \subseteq S_{0, t}$ holds for every $t
  = 1, 2, \ldots, T$, then the output sets from the multiple-level set expansion
  method must satisfy
  \begin{equation*}
    S_t \subseteq \mathcal{E}_t(S_1, S_2, \ldots, S_T) \subseteq S_{0, t},
    \ \forall t = 1, 2, \ldots, T.
  \end{equation*}
\end{theorem}

Knowing both the lower and upper bounds of the output sets, we have the
following theorem showing the exact solution of problem~\eqref{eq:nckqr}
can be obtained by the finite smoothing algorithm.

\begin{theorem}\label{thm:subset2}
  For any $\gamma \in (0, \gamma^{*})$, if there exist $T$ sets, $\hat{S}_1,
  \hat{S}_2, \ldots, \hat{S}_T$, satisfying
  \begin{enumerate}
  \item[(1)] $\hat{S}_t \in S_{0, t}, \ \forall t$,
  \item[(2)] $\mathcal{E}_t(\hat{S}_1, \hat{S}_2, \ldots, \hat{S}_T) = \hat{S}_t, \ \forall t$,
  \end{enumerate}
  then the solution to problem~\eqref{eq:nckqr} can be obtained from the
  following optimization problem,
  \begin{equation}\label{eq:NCKQR_smooth}
    \begin{aligned}
      &{\{\hat{b}_{\tau_t}^{\gamma}, \hat{\bs{\alpha}}_{\tau_t}^{\gamma}\}}_{t=1}^T
        =\argmin_{\{ b_{\tau_t}, \bs\alpha_{\tau_t} \}_{t=1}^T}
        Q^{\gamma}\left({\{ b_{\tau_t}, \bs\alpha_{\tau_t} \}}_{t=1}^T\right), \enskip{}\text{subject to}\enskip{}
        y_{i}=b_{\tau_t}+\mathbf{K}_{i}^{\top}\bs{\alpha}_{\tau_t},
        \, \forall i\in{}\hat{S}_{t}, 1\leq t\leq T.
    \end{aligned}
  \end{equation}
  In other words, we have ${\{\hat{b}_{\tau_t},
  \hat{\bs{\alpha}}_{\tau_t}\}}_{t=1}^T = {\{\hat{b}_{\tau_t}^{\gamma},
  \hat{\bs{\alpha}}_{\tau_t}^{\gamma}\}}_{t=1}^T$.
\end{theorem}

Therefore, similar to the algorithm for the single-level KQR,
Theorem~\ref{thm:subset2} enables one to apply the finite smoothing algorithm to
derive the exact NCKQR solution of problem~\eqref{eq:nckqr}. Starting with
$\hat{S}_t = \emptyset \subseteq S_{0, t}$, for each $t = 1, 2, \ldots, T$, the
multiple-level set expansion method is iteratively applied on these $T$ sets
until no further changes occur. According to Theorem~\ref{thm:subset2}, the
NCKQR solution is then obtained, provided that $\gamma < \gamma^{*}$. As
$\gamma^{*}$ is unknown, to ensure that $\gamma$ is adequately small, the finite
smoothing algorithm is repeated with a decreasing sequence of $\gamma$ values
until a solution satisfying the KKT conditions of problem~\eqref{eq:nckqr} is
eventually identified.

\subsection{Computation}
In the previous section, we extended the finite smoothing algorithm to address
the NCKQR problem. However, even when NCKQR is smoothed into
problem~\eqref{eq:NCKQR_smooth}, the non-crossing penalty increases the
computational demands of NCKQR compared to those of the single-level KQR.\ There
are three main reasons. First, the smooth ReLU and $\gamma$-smoothed quantile
loss functions in problem~\eqref{eq:NCKQR_smooth} have different Lipschitz
constants, hence it is challenging to determine the step size for the
optimization algorithm. Second, the different Lipschitz constants in the two
functions alter the update formula of the proximal gradient descent algorithm,
making the direct implementation of the fast spectral technique introduced in
Section~\ref{sec:fast-kqr} impractical. Third, even if the fast spectral
technique could be utilized, the counterpart of the matrix
$\mathbf{P}_{\gamma, \lambda}$ in equation~\eqref{eq:algo-upd} would be $L$
times larger in NCKQR, making the matrix operations exceedingly costly.

We first solve problem~\eqref{eq:NCKQR_smooth} without the linear constraints,
say, $\hat{S}_t = \emptyset$, for $t = 1, 2, \ldots, T$. To circumvent the three
challenges discussed above, we propose a specialized MM algorithm with two
majorization steps. This MM algorithm can calibrate the Lipschitz constants,
making the fast spectral technique feasible; it can also transform the
counterpart of $\mathbf{P}_{\gamma, \lambda}$ in NCKQR into a block diagonal
matrix, which effectively makes the computation scalable.

The first majorization manages the different Lipschitz constants in $V$ and $H_{\gamma, \tau}$.

Recall that $H_{\gamma,\tau}'$ is
Lipschitz continuous with constant $\gamma^{-1}$, so for any $c_1 \neq c_2$, we
have
\begin{equation}\label{eq:H_inequality}
  H_{\gamma, \tau}(c_1)\leq{}H_{\gamma, \tau}(c_2)+H_{\gamma, \tau}^{\prime}(c_2)(c_1-c_2)
  +\frac{1}{2 \gamma}{(c_1-c_2)}^2.
\end{equation}
Also by the definition of $V$, we can see that $V'$ is Lipschitz continuous with
constant $\eta^{-1}$. To calibrate the two different Lipschitz constants, we
require $\gamma \le \eta$, which gives
\begin{equation}\label{eq:V_inequality}
  \begin{aligned}
    V(c_1)&\leq{}V(c_2)+V^{\prime}(c_2)(c_1-c_2)+\frac{1}{2 \eta}{(c_1-c_2)}^2\\
          &\le V(c_2)+V^{\prime}(c_2)(c_1-c_2)+\frac{1}{2 \gamma}{(c_1-c_2)}^2.
  \end{aligned}
\end{equation}

Let ${\{b_{\tau_t}^{(1)}, \bs\alpha_{\tau_t}^{(1)}\}}_{t=1}^{T}$ be the initial
value. For each $k = 1, 2, \ldots$ and each $t = 1, 2, \ldots, T$, let
\begin{equation*}
  \begin{aligned}
    \bs{\zeta}_{\tau_t, \lambda_1}^{(k)}
    &=\left(
      \begin{array}{c}
        \lambda_1 \mathbf{1}^{\top}(\mathbf{q}_{\tau_{t-1}}^{(k)}-\mathbf{q}_{\tau_{t}}^{(k)})\\
        \lambda_1\mathbf{K}^{\top}(\mathbf{q}_{\tau_{t-1}}^{(k)}-\mathbf{q}_{\tau_{t}}^{(k)})
      \end{array}\right) \enskip{\text{and}}\enskip
    \bs{\zeta}_{\tau_t, \lambda_2}^{(k)}
    =\left(
      \begin{array}{c}
        \mathbf{1}^{\top}\mathbf{z}_{\tau_t}^{(k)}\\
        \mathbf{K}^{\top}\mathbf{z}_{\tau_t}^{(k)}-n\lambda_2\mathbf{K}\bs{\alpha}_{\tau_t}^{(k)}
      \end{array}\right),
  \end{aligned}
\end{equation*}
where $\mathbf{z}_{\tau_t}^{(k)}$ is an $n$-vector whose $i$th element is
$H'_{\gamma,\tau}(y_{i}-b_{\tau_t}^{(k)}-\mathbf{K}_{i}^{\top}\bs{\alpha}_{\tau_t}^{(k)})$,
$\mathbf{q}_{\tau_{t}}^{(k)}$ is an $n$-vector whose $i$th element is
$V'(b_{\tau_t}^{(k)} + \mathbf{K}_{i}^{\top}
\bs{\alpha}_{\tau_t}^{(k)}-b_{\tau_{t+1}}^{(k)}
-\mathbf{K}_{i}^{\top}\bs{\alpha}_{\tau_{t+1}}^{(k)})$, and
$\mathbf{q}_{\tau_{t}}^{(k)} = \mathbf{0}$ if $t=0$ or $t>T$. Define
\begin{equation*}
  \begin{aligned}
    R\left( {\{b_{\tau_t}, \bs\alpha_{\tau_t}\}}_{t=1}^{T} \right) =
    & \sum_{t=1}^{T} \left[ \frac{1}{n} \sum_{i=1}^{n} H_{\gamma, \tau_t}
      \left(y_{i}-b_{\tau_t}-\mathbf{K}_{i}^{\top}
      \bs{\alpha}_{\tau_t} \right)\right]\\
    &+ \lambda_1 \sum_{t=1}^{T-1} \left[\sum_{i=1}^{n}
      V\left( b_{\tau_t}+\mathbf{K}_{i}^{\top}
      \bs{\alpha}_{\tau_t}-b_{\tau_{t+1}}-\mathbf{K}_{i}^{\top}
      \bs{\alpha}_{\tau_{t+1}}\right)\right].
  \end{aligned}
\end{equation*}

With $Q^{\gamma}$ defined in equation~\eqref{eq:Q_gamma_nckqr}, according to
inequalities~\eqref{eq:H_inequality} and~\eqref{eq:V_inequality}, we have
\begin{equation}\label{eq:MM01}
  \begin{aligned}
    & Q^{\gamma}\left( {\{b_{\tau_t}, \bs\alpha_{\tau_t}\}}_{t=1}^{T} \right)
      = \sum_{t=1}^{T} \frac{\lambda_2}{2} \bs{\alpha}^{\top}_{\tau_t}
      \mathbf{K} \bs{\alpha}_{\tau_t}
      + R\left( {\{b_{\tau_t}, \bs\alpha_{\tau_t}\}}_{t=1}^{T} \right)\\
    \leq &\sum_{t=1}^{T} \frac{\lambda_2}{2} \bs{\alpha}^{\top}_{\tau_t}
      \mathbf{K} \bs{\alpha}_{\tau_t}
      +R\left( {\{b_{\tau_t}^{(k)}, \bs\alpha_{\tau_t}^{(k)}\}}_{t=1}^{T} \right)\\
    & + \sum_{t=1}^{T} \left[- \bs\zeta_{\tau_t, \lambda_2}^{\top}\left(
      \begin{array}{c}
        b_{\tau_{t}} - b^{(k)}_{\tau_{t}}\\
        \bs{\alpha}_{\tau_{t}} - \bs\alpha^{(k)}_{\tau_{t}}
      \end{array}\right)
      + \frac{1}{4\gamma}{\left(
      \begin{array}{c}
        b_{\tau_{t}}-b^{(k)}_{\tau_{t}} \\
        \bs{\alpha}_{\tau_{t}}-\bs\alpha^{(k)}_{\tau_{t}}
      \end{array}\right)}^{\top}\mathbf{P}_{\gamma, \lambda_2}
      \left(
      \begin{array}{c}
        b_{\tau_{t}}-b^{(k)}_{\tau_{t}} \\
        \bs{\alpha}_{\tau_{t}}-\bs\alpha^{(k)}_{\tau_{t}}
      \end{array}\right)
      \right] \\
    & + \sum_{t=1}^{T} \left[\bs\zeta_{\tau_t, \lambda_1}^{\top}\left(
      \begin{array}{c}
        b_{\tau_{t}}-b^{(k)}_{\tau_{t}} \\
        \bs{\alpha}_{\tau_{t}}-\bs\alpha^{(k)}_{\tau_{t}}
      \end{array}\right)\right]
      + \frac{1}{4\gamma}
      {\left(\begin{array}{c}
        b_{\tau_1}- b^{(k)}_{\tau_1} \\
        \bs{\alpha}_{\tau_1}- {\bs\alpha}^{(k)}_{\tau_1}\\
        \ldots\\
        b_{\tau_T}- b^{(k)}_{\tau_T} \\
        \bs{\alpha}_{\tau_T}- {\bs\alpha}^{(k)}_{\tau_T}
      \end{array}
      \right)}^{\top} {\bs\Phi}_{\gamma, \lambda_1, \lambda_2}
      \left(\begin{array}{c}
        b_{\tau_1}- b^{(k)}_{\tau_1} \\
        \bs{\alpha}_{\tau_1}- {\bs\alpha}^{(k)}_{\tau_1}\\
        \ldots\\
        b_{\tau_T}- b^{(k)}_{\tau_T} \\
        \bs{\alpha}_{\tau_T}- {\bs\alpha}^{(k)}_{\tau_T}\\
      \end{array}
      \right)\\
     \equiv & \tilde{Q}_{\mathrm{M}}^{\gamma}\left({\{b_{\tau_t}, \bs\alpha_{\tau_t}\}}_{t=1}^{T} \right),\\
  \end{aligned}
\end{equation}
where
\begin{equation*}
 \mathbf{P}_{\gamma,\lambda_2}=\left(
    \begin{array}{cc}
      n & \mathbf{1}^{\top}\mathbf{K}\\
      \mathbf{K}^{\top}\mathbf{1}
        & \mathbf{K}^{\top}\mathbf{K}+2n\gamma\lambda_2\mathbf{K}
    \end{array}\right)
\end{equation*}
and a Block Toeplitz matrix ${\bs\Phi}_{\gamma, \lambda_1, \lambda_2} \in
\mathbb{R}^{(n+1)(T-1) \times (n+1)(T-1)}$ such that
\begin{equation*}
  {\bs\Phi}_{\gamma, \lambda_1, \lambda_2}=
  \left(
    \begin{array}{ccccc}
      \mathbf{B}&-\mathbf{B}&& \\
      -\mathbf{B}&2\mathbf{B}&-\mathbf{B}&\\
                &\ddots& \ddots& \ddots\\
                &&-\mathbf{B}&2\mathbf{B}&-\mathbf{B}\\
                &&&-\mathbf{B}&\mathbf{B}
    \end{array}\right)\enskip\text{with}\enskip
  \mathbf{B}=\left(
    \begin{array}{cc}
      n\lambda_1 & \lambda_1\mathbf{1}^{\top}\mathbf{K}\\
      \lambda_1\mathbf{K}^{\top}\mathbf{1}
                 & \lambda_1\mathbf{K}^{\top}\mathbf{K}
    \end{array}\right).
\end{equation*}
In order to efficiently solve ${\{b_{\tau_t}, \bs\alpha_{\tau_t}\}}_{t=1}^{T}$
using inequality~\eqref{eq:MM01}, we note the matrix operations directly
involving ${\bs\Phi}_{\gamma, \lambda_1, \lambda_2}$ can be computationally
prohibitive. We thus propose to employ a second majorization
to craft a block diagonal ${\bs\Psi}_{\gamma, \lambda_1, \lambda_2}$ that
 majorizes ${\bs\Phi}_{\gamma, \lambda_1, \lambda_2}$,
\begin{equation*}
{\bs\Psi}_{\gamma, \lambda_1, \lambda_2}=
{\bs\Phi}_{\gamma, \lambda_1, \lambda_2}+
\left(
    \begin{array}{ccccc}
      \mathbf{C}&\mathbf{B}&& \\
      \mathbf{B}&\mathbf{C}&\mathbf{B}&\\
      &\ddots& \ddots& \ddots&\\
      &&\mathbf{B}&\mathbf{C}&\mathbf{B}\\
      &&&\mathbf{B}&\mathbf{C}
\end{array}\right) = \left(
    \begin{array}{ccccc}
      \mathbf{B+C}&&& \\
      &\mathbf{2B + C}&&\\
      && \ddots& &\\
      &&&\mathbf{2B + C}&\\
      &&&&\mathbf{B + C}
\end{array}\right),
\end{equation*}
where, with $\varepsilon$ set to $10^{-3}$,
\begin{equation*}
  \mathbf{C}=\left(
    \begin{array}{cc}
      2n\lambda_1+\varepsilon\lambda_1 & 2\lambda_1\mathbf{1}^{\top}\mathbf{K}\\
      2\lambda_1\mathbf{K}^{\top}\mathbf{1}
        & 2\lambda_1\mathbf{K}^{\top}\mathbf{K} + \varepsilon \mathbf{I}
    \end{array}\right).
\end{equation*}

We define ${Q}^{\gamma}_{\mathrm{M}}({\{b_{\tau_t},
\bs\alpha_{\tau_t}\}}_{t=1}^{T})$ by replacing the term ${\bs\Phi}_{\gamma,
  \lambda_1, \lambda_2}$ in $\tilde{Q}^{\gamma}_{\mathrm{M}}({\{b_{\tau_t},
\bs\alpha_{\tau_t}\}}_{t=1}^{T})$ with ${\bs\Psi}_{\gamma, \lambda_1,
  \lambda_2}$. We then derive the MM algorithm to obtain ${\{b_{\tau_t}^{(k+1)},
\bs\alpha_{\tau_t}^{(k+1)}\}}_{t=1}^{T}$ by minimizing
${Q}_{\mathrm{M}}^{\gamma}({\{b_{\tau_t}, \bs\alpha_{\tau_t}\}}_{t=1}^{T})$.
Setting the gradients of ${Q_{\mathrm{M}}^{\gamma}}({\{b_{\tau_t},
\bs\alpha_{\tau_t}\}}_{t=1}^{T})$ to be $\mathbf{0}$, we have
\begin{equation} \label{eq:nckqr-update}
        \left(
    \begin{array}{c}
      b^{(k+1)}_{\tau_t}\\
      \bs{\alpha}^{(k+1)}_{\tau_t}
    \end{array}
  \right)=
  \left(
    \begin{array}{c}
      b^{(k)}_{\tau_t} \\
      \bs{\alpha}^{(k)}_{\tau_t}
    \end{array}
  \right)
  +2\gamma \bs\Sigma_{\gamma,\lambda_1,\lambda_2}^{-1} \bs\varrho^{(k)},
\end{equation}
where
\begin{equation*}
  \bs\Sigma_{\gamma,\lambda_1,\lambda_2}=\left(
    \begin{array}{cc}
      n+4\lambda_1 n^2+ \varepsilon \lambda_1 n & (4 \lambda_1 n + 1 ) \mathbf{1}^{\top}\mathbf{K} \\
      (4 \lambda_1 n + 1 ) \mathbf{K}^{\top}\mathbf{1}
        & (4 \lambda_1 n + 1 ) \mathbf{K}^{\top}\mathbf{K}+2n\gamma\lambda_2\mathbf{K} + \lambda_1 \varepsilon n
    \end{array}\right)
\end{equation*}
and
\begin{equation*}
\bs\varrho^{(k)} =
    \left\{
      \begin{array}{ll}
  \left(
    \begin{array}{c}
      \mathbf{1}^{\top}\mathbf{z}^{(k)}_{\tau_t} - n\lambda_1\mathbf{1}^{\top}\mathbf{q}^{(k)}_{\tau_t}\\
      \mathbf{K}^{\top}\mathbf{z}^{(k)}_{\tau_t}-n\lambda_2\mathbf{K}\bs{\alpha}^{(k)}_{\tau_t}-n\lambda_1\mathbf{K}\mathbf{q}^{(k)}_{\tau_t}
    \end{array}\right), &t=1, \\
  \left(
    \begin{array}{c}
      \mathbf{1}^{\top}\mathbf{z}^{(k)}_{\tau_t} - n\lambda_1\mathbf{1}^{\top}\mathbf{q}^{(k)}_{\tau_t}+n\lambda_1\mathbf{1}^{\top}\mathbf{q}^{(k)}_{\tau_{t-1}}\\
      \mathbf{K}^{\top}\mathbf{z}^{(k)}_{\tau_t}-n\lambda_2\mathbf{K}\bs{\alpha}^{(k)}_{\tau_t}-n\lambda_1\mathbf{K}\mathbf{q}^{(k)}_{\tau_t}+n\lambda_1\mathbf{K}\mathbf{q}^{(k)}_{\tau_{t-1}}
    \end{array}\right), & 2 \le t \le T - 1, \\
   \left(
    \begin{array}{c}
      \mathbf{1}^{\top}\mathbf{z}^{(k)}_{\tau_t} + n\lambda_1\mathbf{1}^{\top}\mathbf{q}^{(k)}_{\tau_{t-1}}\\
      \mathbf{K}^{\top}\mathbf{z}^{(k)}_{\tau_t}-n\lambda_2\mathbf{K}\bs{\alpha}^{(k)}_{\tau_t}+n\lambda_1\mathbf{K}\mathbf{q}^{(k)}_{\tau_{t-1}}
    \end{array}\right), &t=T.
    \end{array}
    \right.
  \end{equation*}

  The fast spectral technique can be extended to address the repeated
  computation of inverting
  $\bs\Sigma_{\gamma,\lambda_1, \lambda_2} \in \mathbb{R}^{(n+1) \times (n+1)}$,
  for varying values of $\gamma$, $\lambda_1$ and $\lambda_2$. Similar to the
  single-level KQR, we begin with the eigendecomposition of
  $\mathbf{K}=\mathbf{U} \bs{\Lambda} \mathbf{U}^{\top}$, which does not vary
  with tuning parameters. After this step, the entire algorithm of NCKQR only
  involves $\mathcal{O}(n^2)$ operations. Further details are provided in
  Section B of the online supplemental material.
  
  When the linear constraints in problem~\eqref{eq:NCKQR_smooth} are present, we
  use the same projection in problem~\eqref{eq:proj} to obtain
  ${\{\tilde{b}_{\tau_t}, \tilde{\bs\alpha}_{\tau_t}\}}_{t=1}^T$ as
\begin{equation}\label{eq:proj_sol_NCKQR}
  \begin{aligned}
    \tilde{b}_{\tau_t}
    = b_{\tau_t}^{(k)} + \frac{1}{|\hat{S}_t|+1}\sum_{i \in \hat{S}_t}
      [ y_i - \mathbf{K}_i^\top\bs\alpha_{\tau_t}^{(k)}], \ 
    \tilde{\bs\alpha}_{\tau_t}
    = \mathbf{K}^{-1}\bs\theta,
  \end{aligned}
\end{equation}
where $|\hat{S}_t|$ is the cardinality of $\hat{S}_t$, and
$\bs\theta\in \mathbb{R}^n$ with $\theta_i = y_i -\tilde{b}_{\tau_t}$ if $i \in
\hat{S}$ and $\theta_i= \mathbf{K}_i^\top \bs\alpha_{\tau_t}^{(k)}$ otherwise.
We then use ${\{\tilde{b}_{\tau_t}, \tilde{\bs\alpha}_{\tau_t}\}}_{t=1}^T$ in
place of ${\{{b}^{[k]}_{\tau_t}, {\bs\alpha}^{[k]}_{\tau_t}\}}_{t=1}^T$ in
problem~\eqref{eq:MM01} to proceed with the algorithm.

We also note that the MM algorithm requires $\gamma \le \eta$, where
$\eta = 10^{-5}$. This condition guarantees the proper majorization within the
MM algorithm but it may also drive the algorithm overly conservative, slowing
progress toward the solution. In our implementation, we begin with
$\gamma = \eta = 1$ and iteratively reduce both parameters to a quarter of their
previous values, terminating the algorithm if the KKT conditions of
problem~\eqref{eq:nckqr} are met. If the algorithm does not terminate when we
reach some $\gamma = \eta < 10^{-5}$, we maintain $\eta = 10^{-5}$ and continue
to decrease $\gamma$. Similar to the single-level KQR, the algorithm typically
stops after three or four iterations of updating $\gamma$.

The NCKQR algorithm is summarized in Algorithm 2 in the online supplemental
material.


\section{Numerical Studies}\label{sec:simu}

We now use simulation experiments to showcase the performance of our
\texttt{fastkqr} algorithm for the single-level KQR and multi-level NCKQR, in
Sections~\ref{sec:simu-kqr} and~\ref{sec:nckqr02}, respectively. Additional
simulations and benchmark data applications are present in the online
supplemental materials.

\subsection{Kernel quantile regression}\label{sec:simu-kqr}

We compare our \texttt{fastkqr} algorithm with the Alternating Direction Method
of Multipliers (ADMM) algorithm \citep{boyd2011distributed}, the optimizer
\texttt{kqr} in the R package \texttt{kernlab} \citep{AA04}, the R packages
\texttt{clarabel} \citep{goulart2024clarabel} and \texttt{osqp}
\citep{stellato2020osqp}, and the two generic optimizers \texttt{nlm} and
\texttt{optim} in the R package \texttt{stats}. We choose the radial basis
kernel and explore various combinations of the sample size $n$ and dimension
$p$. We consider three different quantile levels: $\tau = 0.1$, $0.5$, and
$0.9$. For each training data, we apply the seven solvers to fit KQR over 50
$\lambda$ values. The optimal tuning parameter $\lambda$ is selected using
five-fold cross-validation, and the whole run time is recorded. The selected
$\lambda$ is then used to compute the objective value of
problem~\eqref{eq:representer}. The reported results represent averages from 20
independent repetitions. All computations are carried out on an Apple M1 (16GB)
processor.

   \begin{table}[p]
  \centering
  \resizebox{1.0\textwidth}{!}{%
  \begin{tabular}{rrrrrrrrrrr}
    \toprule
    $\tau$ & $n$ & &\texttt{fastkqr} & \texttt{kernlab} & \texttt{clarabel} & \texttt{ADMM} & \texttt{osqp} & \texttt{nlm} & \texttt{optim} \\
    \midrule
    0.1 & 200 & \text{obj} & $0.365_{(0.044)}$ & $0.365_{(0.044)}$ & $0.365_{(0.044)}$ & $0.365_{(0.044)}$ & $0.365_{(0.044)}$ & $0.367_{(0.041)}$ & $0.384_{(0.033)}$  \\
    & & \text{time} & 2.27 & 5.78 & 23.69 & 150.95 & 8.71 & 214.60 & 450.95 \\
    & 500 & \text{obj} & $0.356_{(0.019)}$ & $0.356_{(0.019)}$ & $0.356_{(0.019)}$ & $0.356_{(0.019)}$ & $0.356_{(0.019)}$ & $0.358_{(0.019)}$ & $0.368_{(0.020)}$  \\
    & & \text{time} & 9.46 & 57.32 & 363.70 & 2479.34 & 149.68 & 3263.86 & 6671.74 \\
    & 1000 & \text{obj} & $0.360_{(0.012)}$ & $0.360_{(0.012)}$ & $0.360_{(0.012)}$ & $0.360_{(0.012)}$ & $0.360_{(0.012)}$ & $0.361_{(0.012)}$ & $0.366_{(0.011)}$ \\
    & & \text{time} & 28.20 & 408.48 & 3802.69 & 10762.72 & 646.65 & 26906.18 & 54851.15 \\
    0.5 & 200 & \text{obj} & $0.813_{(0.063)}$ & $0.813_{(0.063)}$ & $0.813_{(0.063)}$ & $0.813_{(0.063)}$ & $0.813_{(0.063)}$ & $0.822_{(0.057)}$ & $0.844_{(0.053)}$ \\
    & & \text{time} & 2.56 & 5.53 & 19.65 & 218.50 & 8.76 & 221.58 & 449.92 \\
    & 500 & \text{obj} & $0.807_{(0.044)}$ & $0.807_{(0.044)}$ & $0.807_{(0.044)}$ & $0.807_{(0.044)}$ & $0.807_{(0.044)}$ & $0.813_{(0.041)}$ & $0.827_{(0.040)}$  \\
    & & \text{time} & 9.56 & 54.83 & 285.14 & 2654.29 & 141.39 & 3328.60 & 6804.10 \\
    & 1000 & \text{obj} & $0.811_{(0.024)}$ & $0.811_{(0.024)}$ & $0.811_{(0.024)}$ & $0.811_{(0.024)}$ & $0.812_{(0.024)}$ & $0.816_{(0.022)}$ & $0.829_{(0.020)}$  \\
    & & \text{time} & 28.44 & 368.38 & 2711.26 & 10784.24 & 661.21 & 27568.66 & 55697.12 \\
    0.9 & 200 & \text{obj} & $0.380_{(0.041)}$ & $0.380_{(0.041)}$ & $0.380_{(0.041)}$ & $0.380_{(0.041)}$ & $0.380_{(0.041)}$ & $0.384_{(0.038)}$ & $0.403_{(0.034)}$ &\\
    & & \text{time} & 2.29 & 5.09 & 26.36 & 121.39 & 7.35 & 224.17 & 455.74 \\
    & 500 & \text{obj} & $0.377_{(0.032)}$ & $0.377_{(0.032)}$ & $0.377_{(0.032)}$ & $0.377_{(0.032)}$ & $0.377_{(0.032)}$ & $0.378_{(0.031)}$ & $0.394_{(0.037)}$  \\
    & & \text{time} & 9.57 & 53.47 & 377.31 & 2344.44 & 125.50 & 3326.35 & 
    6709.23 \\
    & 1000 & \text{obj} & $0.365_{(0.010)}$ & $0.365_{(0.010)}$ & $0.365_{(0.010)}$ & $0.365_{(0.010)}$ & $0.365_{(0.010)}$ & $0.366_{(0.010)}$ & $0.403_{(0.046)}$ \\
    & & \text{time} & 27.98 & 386.04 & 3785.58 & 10346.09 & 622.64 & 27715.45 & 55714.28 \\
  \bottomrule
  \end{tabular}}
  \caption{Objective values and computation time of seven kernel quantile regression solvers for simulation data \citep{Yuan2006} with $p=2$, $n=\{200, 500, 1000\}$, and $\tau=\{0.1,  0.5, 0.9\}$. The numbers are the average quantities over 20 independent runs
    and the standard errors are presented in the parentheses.}
  \label{tab1:sim-p=2}
\end{table}

\begin{table}[t]
  \centering
  \begin{tabular}{lllllll}
    \toprule
     $n$ & & \texttt{fastkqr} & \texttt{cvxr} & \texttt{nlm} & \texttt{optim}  \\
    \midrule
     200 & \text{obj} & $2.527_{(0.108)}$ & $2.630_{(0.105)}$ & $3.048_{(0.105)}$ & 
    $3.127_{(0.135)}$ \\
    & \text{time}  & 3.37 & 1149.93 & 2938.42 & 6392.16\\
    500 & \text{obj} & $2.730_{(0.112)}$ & 
    $2.998_{(0.162)}$ & $3.423_{(0.132)}$ & * \\
    & \text{time} & 17.06 & 3439.39 & 41266.57 &$>24$h \\
    1000 & \text{obj} & $3.380_{(0.114)}$ & * & * & * \\
    & \text{time} &  57.56 & 20171.01 &$>24$h & $>24$h \\
    \bottomrule
  \end{tabular}
  \caption{Objective values and computation time of four NCKQR solvers for simulation data \citep{Yuan2006} with $p=2$, $n=\{200, 500, 1000\}$, and $\tau=\{0.1, 0.5, 0.9\}$.
    The average quantities over 20 independent runs are displayed, and
    standard errors are presented in parentheses. A result shown as a star
    ``$*$'' means the corresponding solver cannot output a solution due to
    numerical issues.}
  \label{tab:sim-obj-nckqr}
\end{table}

Following \cite{Yuan2006}, two-dimensional data are
generated based on
\begin{equation}
  Y=\frac{40\exp\left[8\left\{(X_{1}-.5)^{2}+(X_{2}-.5)^{2}\right\}\right]}
  {\exp\left[8\left\{(X_{1}-.2)^{2}+(X_{2}-.7)^{2}\right\}\right]
    +\exp\left[8\left\{(X_{1}-.7)^{2}+(X_{2}-.2)^{2}\right\}\right]}+\epsilon,
\end{equation}
where $X_1$ and $X_2$ are drawn from Uniform $(0,1)$, and the error term
$\epsilon$ is from the standard normal distribution.

Tables \ref{tab1:sim-p=2} displays the average
objective values and computation time for the seven solvers, with all the run time including both model training and tuning parameter
selection. Our algorithm \texttt{fastkqr} consistently outperforms the other solvers in speed across all the examples. For instance, when $n = 1000$, our fastkqr algorithm
was more than an order of magnitude faster than \texttt{kernlab}, and notably, more than 400 times faster than \texttt{nlm} and \texttt{optim}. As the sample size $n$ grows, the computational advantages of our algorithm
over the others become more significant. Additionally, apart from \texttt{nlm} and \texttt{optim}, the other five solvers exhibit roughly the same objective values, which are notably lower than those of the two generic optimizers.

\subsection{Non-crossing kernel quantile regression}\label{sec:nckqr02}

In this section, we compare \texttt{fastkqr} with the R package \texttt{cvxr} \citep{JSS20}, and \texttt{nlm} and \texttt{optim} for fitting NCKQR. 
In each scenario, we simultaneously fit three quantile curves with $\tau = 0.1, 0.5$, and $0.9$ using the four solvers. With 20 independent runs, we calculate the average objective values of problem~\eqref{eq:nckqr} and the total run time across 50 $\lambda_2$
values. 

Table~\ref{tab:sim-obj-nckqr} presents the average
objective values and computation time. Our algorithm is the fastest and the most accurate. For example, when $n=500$, our algorithm took only seventeen seconds, in contrast to \texttt{cvxr}
which spent about one hour. Furthermore, when $n$ is increased to 1000, our algorithm spent just 57 seconds, while \texttt{nlm} and \texttt{optim} took over 24 hours.



\section{Discussions}
In this paper, we have developed \texttt{fastkqr}, a fast algorithm for
computing the exact solution of kernel quantile regression. Our approach is
based on a finite smoothing algorithm and accelerated proximal gradient descent,
further enhanced by a fast spectral technique that optimizes matrix operations.
Notably, \texttt{fastkqr} is faster than \texttt{kernlab} while
maintaining nearly identical accuracy.

Furthermore, we have addressed the crossing of
quantile curves. We have introduced the non-crossing kernel quantile regression
with a soft non-crossing penalty, and have
expanded \texttt{fastkqr} with a specialized MM algorithm featuring two
majorization steps. We show that \texttt{fastkqr} significantly
outperforms \texttt{cvxr} in both computational speed and accuracy.

To broaden the applicability of \texttt{fastkqr} to large-scale data analysis,
we propose to integrate various kernel approximation techniques into our
existing algorithmic framework. Methods such as random features \citep{Rahimi07}
or Nystr\"om subsampling \citep{rudi2015less} could be employed within the exact update formula of kernel quantile regression to create a cost-effective
surrogate of the kernel matrix. These approximation
strategies are expected to further enhance the efficiency of our algorithm. We leave a full investigation of this direction for future work.


\section{Acknowledgments}
The authors extend their sincere gratitude to the Editor, the Associate Editor, and the two anonymous Referees for their insightful and constructive comments, which have greatly improved the quality of our work.

\section{Disclosure Statement}
The authors report there are no competing interests to declare.
\bibliographystyle{abbrvnat}
\bibliography{ref}


\spacingset{1.25} 
\newpage

\setcounter{page}{1}
\appendix

\makeatletter
\setcounter{figure}{0}
\setcounter{table}{0}
\renewcommand \thesection{S\@arabic\c@section}
\renewcommand\thetable{S\@arabic\c@table}
\renewcommand \thefigure{S\@arabic\c@figure}
\renewcommand \theequation{S\@arabic\c@equation}
\makeatother

\bigskip
\begin{center}
{\large\bf SUPPLEMENTARY MATERIAL}
\end{center}

\section{Plot of $\gamma$-Smoothed Check Loss}

\begin{figure}[h] \label{fig:loss}
\centering
\includegraphics[width=0.8\textwidth]{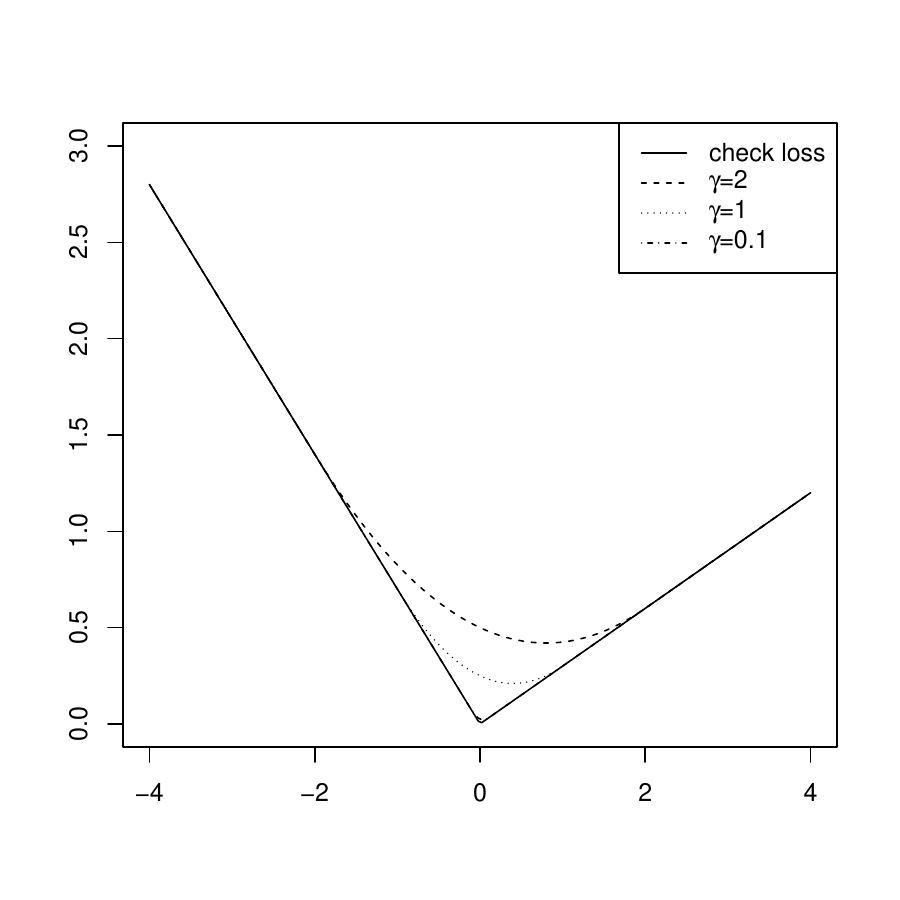}
\caption{check loss versus the$\gamma$-smoothed check loss}
\end{figure}

Figure~\ref{fig:loss} illustrates the $\gamma$-smoothed check loss for various values of $\gamma$, along with the original check loss.

\section{Algorithms}

The complete algorithm for solving the KQR in
Problem~\eqref{eq:representer} is detailed in Algorithm \ref{alg:kqr}. Algorithm~\ref{alg:nckqr} summarizes the entire algorithm for computing the NCKQR.

\begin{algorithm}[p]
  \caption{The \texttt{fastkqr} algorithm for solving the kernel quantile
    regression in Problem~\eqref{eq:representer}} \label{alg:kqr}
  \begin{algorithmic}[1]
    \REQUIRE $\mathbf{y}$, $\mathbf{K}$, $\tau$, and $\lambda^{[1]} > \lambda^{[2]} > \ldots > \lambda^{[L]}$.%
    \STATE Carry out the eigen-decomposition $\mathbf{K}=\mathbf{U}\bs{\Lambda}\mathbf{U}^{\top}$ \textit{only once}.%
    \STATE Initialize $\gamma = 1$, $\delta = 1/4$, and $(\tilde{b}, \tilde{\bs\alpha})$.%
    \FOR {$l = 1, 2, \ldots, L$}
    \STATE Set $\lambda \leftarrow \lambda^{[l]}$.
    \REPEAT {}%
    \STATE Compute
    $\mathbf{v}=\mathbf{U}\bs{\Lambda}\bs{\Pi}_{\gamma,\lambda}^{-1}
    \mathbf{U}^{\top}\mathbf{1}$ and $g=\bigl(n-\mathbf{1}^{\top}
    \mathbf{U}\bs{\Lambda}\bs{\Pi}_{\gamma,\lambda}^{-1}
    \bs{\Lambda}\mathbf{U}^{\top}\mathbf{1}\bigr)^{-1}$,
    where $\bs{\Pi}_{\gamma,\lambda}=\bs{\Lambda}^{2}
    +2n\gamma\lambda\bs{\Lambda}.$%
    \STATE Set $\hat{S} \leftarrow \emptyset$.
    \REPEAT {}%
    \STATE Set $k \leftarrow 1$ and $c_{1} \leftarrow 1$.%
    \STATE Initialize $({b}^{[0]}, {\bs\alpha}^{[0]}) = ({b}^{[1]}, {\bs\alpha}^{[1]}) \leftarrow (\tilde{b}, \tilde{\bs\alpha})$.
    \REPEAT {}%
    \STATE Compute $c_{k+1} \leftarrow \frac{1+\sqrt{1+4c_{k}^{2}}}{2}$.%
    \STATE Update
    \begin{equation*}
      \left(
        \begin{array}{c}
          \bar{b}^{(k+1)} \\
          \bar{\bs{\alpha}}^{(k+1)}
        \end{array}\right) \leftarrow \left(
      \begin{array}{c}
        b^{(k)} \\
        \bs{\alpha}^{(k)}
      \end{array}\right)+
      \left(\frac{c_{k}-1}{c_{k+1}}\right)\left(
        \begin{array}{c}
          b^{(k)}-b^{(k-1)} \\
          \bs{\alpha}^{(k)}-\bs{\alpha}^{(k-1)}
        \end{array}\right).
    \end{equation*}
    \STATE Update $\overline{{z}}_i \leftarrow \frac1n{}
    H_{\gamma,\tau}'(y_{i}-\bar{b}^{(k+1)}-\mathbf{K}_{i}^{\top}
    \bs{\bar\alpha}^{(k+1)}), \ i=1,\ldots,n.$%
    \STATE Calculate \emph{from right to left}
    $$
      \bs\mu \leftarrow g\{
      \mathbf{1}^{\top}\bar{\mathbf{z}}-\mathbf{v}^{\top}\mathbf{K}
        (\bar{\mathbf{z}}+n\lambda\bar{\bs{\alpha}}^{k+1})\}\left(
      \begin{array}{c}
        1 \\
        -\mathbf{v}
      \end{array}\right)+\left(
        \begin{array}{c}
          0 \\
          \mathbf{U}\bs\Pi_{\gamma,\lambda}^{-1}\bs{\Lambda}
          \mathbf{U}^{\top}(\bar{\mathbf{z}}+n\lambda\bar{\bs{\alpha}}^{k+1})
        \end{array}\right).
    $$
    \STATE Update
    $$
      \left(
        \begin{array}{c}
          b^{(k+1)} \\
          \bs{\alpha}^{(k+1)}
        \end{array}\right) \leftarrow \left(
        \begin{array}{c}
          \bar{b}^{(k+1)} \\
          \bs{\bar\alpha}^{(k+1)}
        \end{array}\right)
      +\bs\mu.
    $$
    \STATE Update $k\leftarrow{}k+1$.%
    \UNTIL {the convergence criterion is met.}%
    \STATE Update $(\tilde{b}, \tilde{\bs\alpha})$ from Problem~\eqref{eq:proj}.%
    \STATE Update $\hat{S} \leftarrow \{i: -\gamma \leq y_{i}-\tilde{b} - \mathbf{K}_{i}^{\top}\tilde{\bs{\alpha}} \leq \gamma\}$.
    \UNTIL {the set $\hat{S}$ is unchanged.}%
    \STATE Update $\gamma\leftarrow\delta\gamma$.%
    \UNTIL the KKT conditions of the KQR problem are satisfied.
    \STATE Set $(\hat{b}^{[l]}, \hat{\bs\alpha}^{[l]}) \leftarrow (\tilde{b}, \tilde{\bs\alpha})$.
    \ENDFOR
    \ENSURE KQR solution, $(\hat{b}^{[1]}, \hat{\bs\alpha}^{[1]}), (\hat{b}^{[2]}, \hat{\bs\alpha}^{[2]}), \ldots, (\hat{b}^{[L]}, \hat{\bs\alpha}^{[L]})$.
  \end{algorithmic}
\end{algorithm}

\begin{algorithm}[p]
  \caption{The \texttt{fastkqr} algorithm for solving NCKQR in Problem~\eqref{eq:nckqr}}   \label{alg:nckqr}
  \begin{algorithmic}[1]
    \REQUIRE $\mathbf{y}$, $\mathbf{K}$, $\tau_1<\tau_2<\ldots<\tau_T$, $\lambda_1^{[1]} > \lambda_1^{[2]} > \ldots > \lambda_1^{[L_1]}$, and $\lambda_2^{[1]} > \lambda_2^{[2]} > \ldots > \lambda_2^{[L_2]}$.
    \STATE Carry out the eigen-decomposition $\mathbf{K}=\mathbf{U}\bs{\Lambda}\mathbf{U}^{\top}$ \textit{only once}.
    \STATE Initialize $\gamma=1$, $\delta=1/4$, and $\{\tilde{b}_{\tau_t}, \tilde{\bs\alpha}_{\tau_t}\}_{t=1}^{T}$.
    \FOR {$l_1 = 1, 2, \ldots, L_1$}
    \FOR {$l_2 = 1, 2, \ldots, L_2$}
    \STATE Set $\lambda_1 \leftarrow \lambda_{1}^{[l_1]}$ and $\lambda_2 \leftarrow \lambda_{2}^{[l_2]}$.
    \REPEAT {}%
    \STATE Compute
    $\mathbf{v}=\left( 4 \lambda_2 n + 1\right) \mathbf{U} \bs{\Lambda} \bs{\Pi}_{\gamma, \lambda_2}^{-1} \mathbf{U}^{\top} \mathbf{1}$.
    \STATE Compute $\bs{\Pi}_{\gamma, \lambda_2}=\left( 4 \lambda_2 n + 1\right) \bs{\Lambda} \bs{\Lambda}+ \lambda_2 \varepsilon n + 2n \lambda_2 \gamma \bs{\Lambda}$.
    \STATE Compute $g=[\left(4 \lambda_2 n + 1\right)n+ \lambda_2 \varepsilon n-\left( 4 \lambda_2 n + 1\right)^2\mathbf{1}^{\top} \mathbf{U} \bs{\Lambda} \bs{\Pi}_{\gamma, \lambda_2}^{-1} \bs{\Lambda} \mathbf{U}^{\top} \mathbf{1}]^{-1}$.%
    \STATE Set $\hat{S}_1 = \hat{S}_2 = \ldots = \hat{S}_T \leftarrow \emptyset$.
    \REPEAT {}
    \STATE Set $k=1$.
    \STATE Initialize $\{b_{\tau_t}^{(k)}, \bs\alpha_{\tau_t}^{(k)}\}_{t=1}^T \leftarrow \{\tilde{b}_{\tau_t}, \tilde{\bs\alpha}_{\tau_t}\}_{t=1}^{T}$.
    \REPEAT {}%
    \STATE Update $(\mathbf{z}_{\tau_t})_i \leftarrow
    H_{\gamma,\tau_t}'(y_{i}-b^{(k+1)}_{\tau_t}-\mathbf{K}_{i}^{\top} \bs{\alpha}^{(k+1)}_{\tau_t})$ for each $t = 1, 2, \ldots, T$.
    \STATE Update $(\mathbf{q}_{\tau_t})_i \leftarrow V'(\mathbf{K}_{i}^{\top}\bs{\alpha}^{(k)}_{\tau_t}+b^{(k)}_{\tau_t}-\mathbf{K}_{i}^{\top}\bs{\alpha}^{(k)}_{\tau_{t+1}}-b^{(k)}_{\tau_{t+1}})$ for each $t = 1, 2, \ldots, T-1$.
    \STATE Calculate the update formulae \eqref{eq:nckqr-formula1},  \eqref{eq:nckqr-formula2} and \eqref{eq:nckqr-formula3} \emph{from right to left}.
    \STATE Update $\{b_{\tau_t}^{(k)}, \bs\alpha_{\tau_t}^{(k)}\}_{t=1}^T$ using formula \eqref{eq:nckqr-update}.
    \STATE Update $k\leftarrow{}k+1$.%
    \UNTIL {the convergence criterion is met.}%
    \STATE Update $\{\tilde{b}_{\tau_t}, \tilde{\bs\alpha}_{\tau_t}\}_{t=1}^{T}$ using formula~\eqref{eq:proj_sol_NCKQR}.
    \STATE Update $\hat{S}_{t} \leftarrow \{i: -\gamma{}\leq{}
    y_{i}-\tilde{b}_{\tau_t}-\mathbf{K}_{i}^{\top}
    \tilde{\bs{\alpha}}_{\tau_t} \leq \gamma \}$ for each $t = 1, 2, \ldots, T$.
    \UNTIL all the sets $\hat{S}_1, \hat{S}_2, \ldots, \hat{S}_T$ are unchanged.
    \STATE Update $\gamma\leftarrow\delta\gamma$.%
    \UNTIL the KKT conditions of the NCKQR problem are satisfied.
    \STATE Set $\{\hat{b}_{\tau_t}^{[l_1, \ l_2]}, \hat{\bs\alpha}_{\tau_t}^{[l_1, \ l_2]}\}_{t=1}^T \leftarrow \{\tilde{b}_{\tau_t}, \tilde{\bs\alpha}_{\tau_t}\}_{t=1}^{T}$.
    \ENDFOR
    \ENDFOR
    \ENSURE NCKQR solution, $\{\hat{b}_{\tau_t}^{[l_1, \ l_2]}, \hat{\bs\alpha}_{\tau_t}^{[l_1, \ l_2]}\}_{t=1}^T$, for each $l_1 = 1, 2, \ldots, L_1$ and $l_2 = 1, 2, \ldots, L_2$.
  \end{algorithmic}
\end{algorithm}

\section{The Fast Spectral Technique for NCKQR}
In this section, we study the fast spectral technique for the computation of NCKQR in Section 3.3. Specifically, the fast spectral technique is developed to address the repeated computation 
  of inverting $\bs\Sigma_{\gamma,\lambda_1, \lambda_2} \in \mathbb{R}^{(n+1)
    \times (n+1)}$, in equation~\eqref{eq:nckqr-update}, for varying values of $\gamma$, $\lambda_1$ and $\lambda_2$.
  We calculate
  $\bs{\Pi}_{\gamma, \lambda_1, \lambda_2}=\left( 4 \lambda_1 n +
    1\right) \bs{\Lambda} \bs{\Lambda}+ \lambda_1 \varepsilon n
  +2 n \lambda_2 \gamma \bs{\Lambda}$ for each $\lambda_1$ and $\gamma$.
  By employing $g=1 /[( 4 \lambda_1 n + 1)n+ \lambda_1 \varepsilon n-\left( 4
    \lambda_1 n + 1\right)^2\mathbf{1}^{\top} \mathbf{U} \bs{\Lambda}
  \bs{\Pi}_{\gamma, \lambda_1}^{-1} \bs{\Lambda}
  \mathbf{U}^{\top} \mathbf{1}]$ and $\mathbf{v}=\left( 4 \lambda_1 n + 1\right)
  \mathbf{U} \bs{\Lambda} \bs{\Pi}_{\gamma, \lambda_1,
    \lambda_2}^{-1} \mathbf{U}^{\top} \mathbf{1}$, we can readily attain the
  desired decomposition
\begin{equation*}
  \begin{aligned}
    \bs\Sigma_{\gamma,\lambda_1, \lambda_2}^{-1}
    & =
      \left(
      \begin{array}{cc}
        n+4 \lambda_1 n^2 + \lambda_1 \varepsilon n
        & \left(4 \lambda_1 n + 1\right) \mathbf{1}^{\top} \mathbf{U} \bs{\Lambda} \mathbf{U}^{\top} \\
        \left(4 \lambda_1 n + 1\right) \mathbf{U} \bs{\Lambda} \mathbf{U}^{\top} \mathbf{1}
        & \mathbf{U} \bs{\Pi}_{\gamma,\lambda_1, \lambda_2} \mathbf{U}^{\top}
      \end{array}\right)^{-1} \\
    &=g\left(\begin{array}{c}1 \\
      -\mathbf{v}\end{array}\right)
      \left(\begin{array}{ll}
        1 & -\mathbf{v}^{\top}
      \end{array}\right)
      +
      \left(\begin{array}{cc}
        0 & \mathbf{0}^{\top} \\
        \mathbf{0} & \mathbf{U} \bs{\Pi}_{\gamma, \lambda_1, \lambda_2}^{-1} \mathbf{U}^{\top}
      \end{array}\right).
  \end{aligned}
\end{equation*}
Rather than computing $\bs\Sigma_{\gamma,\lambda_1, \lambda_2}^{-1}$, we
directly compute the following matrix-vector multiplications.
\begin{itemize}
\item Case 1. When $t=1$, we have
  \begin{equation}
    \begin{aligned}\label{eq:nckqr-formula1}
      &\bs\Sigma_{\gamma,\lambda_1, \lambda_2}^{-1}
        \left(
        \begin{array}{c}
          \mathbf{1}^{\top}\mathbf{z}^{(k)}_{\tau_t} - n\lambda_1\mathbf{1}^{\top}\mathbf{q}^{(k)}_{\tau_t}\\
          \mathbf{K}^{\top}\mathbf{z}^{(k)}_{\tau_t}-n\lambda_2\mathbf{K}^{\top}\bs{\alpha}^{(k)}_{\tau_t}-n\lambda_1\mathbf{K}^{\top}\mathbf{q}^{(k)}_{\tau_t}
        \end{array}\right) \\
      &= g\left\{\mathbf{1}^{\top} \mathbf{z}^{(k)}_{\tau_t}-\lambda_1 \mathbf{1}^{\top}
        \mathbf{q}^{(k)}_{\tau_t}n-\mathbf{v}^{\top} \mathbf{K}\left(\mathbf{z}^{(k)}_{\tau_t}
        + n\lambda_2 \bs{\alpha}^{(k)}_{\tau_t}-n\lambda_1 \mathbf{q}^{(k)}_{\tau_t} \right)\right\}
         \left(\begin{array}{c}1 \\
           -\mathbf{v}\end{array}\right) \\
      &\quad +\left(\begin{array}{c}
          0 \\
        \mathbf{U} \bs{\Pi}_{\gamma, \lambda_1, \lambda_2}^{-1} \bs{\Lambda}
        \mathbf{U}^{\top}\left(\mathbf{z}^{(k)}_{\tau_t}+n\lambda_2
        \bs{\alpha}^{(k)}_{\tau_t}-n\lambda_1 \mathbf{q}^{(k)}_{\tau_t}\right)
        \end{array}\right).
    \end{aligned}
  \end{equation}
\item Case 2. When $2\le t \le T-1$, we have
  \begin{eqnarray}
    &&\bs\Sigma_{\gamma,\lambda_1, \lambda_2}^{-1}
      \left(
      \begin{array}{c}
        \mathbf{1}^{\top}\mathbf{z}^{(k)}_{\tau_t} - n\lambda_1\mathbf{1}^{\top}
        \mathbf{q}^{(k)}_{\tau_t}+ n\lambda_1\mathbf{1}^{\top}\mathbf{q}^{(k)}_{\tau_{t-1}}\\
        \mathbf{K}^{\top}\mathbf{z}^{(k)}_{\tau_t}-n\lambda_2\mathbf{K}^{\top}
        \bs{\alpha}^{(k)}_{\tau_t}-n\lambda_1\mathbf{K}^{\top}\mathbf{q}^{(k)}_{\tau_t}
        + n\lambda_1\mathbf{K}^{\top}\mathbf{q}^{(k)}_{\tau_{t-1}}
      \end{array}\right)\label{eq:nckqr-formula2} \\
    &&= g\left\{\mathbf{1}^{\top} \mathbf{z}^{(k)}_{\tau_t}- n\lambda_1\mathbf{1}^{\top}
      \mathbf{q}^{(k)}_{\tau_t}+ n\lambda_1\mathbf{1}^{\top}\mathbf{q}^{(k)}_{\tau_{t-1}}
      -\mathbf{v}^{\top} \mathbf{K}\left(\mathbf{z}^{(k)}_{\tau_t}+ n\lambda_2
      \bs{\alpha}^{(k)}_{\tau_t}- n\lambda_1\mathbf{q}^{(k)}_{\tau_t}
      + n\lambda_1\mathbf{q}^{(k)}_{\tau_{t-1}} \right)\right\}\nonumber  \\
    &&\quad\cdot\left(\begin{array}{c}1 \\
      -\mathbf{v}\end{array}\right)
      + \left(\begin{array}{c}
        0 \\
        \mathbf{U} \bs{\Pi}_{\gamma,\lambda_1, \lambda_2}^{-1} \bs{\Lambda}
        \mathbf{U}^{\top}\left(\mathbf{z}^{(k)}_{\tau_t}+n\lambda_2
        \bs{\alpha}^{(k)}_{\tau_t}- n\lambda_1\mathbf{q}^{(k)}_{\tau_t}
        + n\lambda_1\mathbf{q}^{(k)}_{\tau_{t-1}}\right)
      \end{array}\right).\nonumber
  \end{eqnarray}
\item Case 3. When $t=T$, we have
  \begin{equation}
    \begin{aligned}\label{eq:nckqr-formula3}
      &\bs\Sigma_{\gamma,\lambda_1, \lambda_2}^{-1}
        \left(
        \begin{array}{c}
          \mathbf{1}^{\top}\mathbf{z}^{(k)}_{\tau_t} + n\lambda_1
          \mathbf{1}^{\top}\mathbf{q}^{(k)}_{\tau_{t-1}}\\
          \mathbf{K}^{\top}\mathbf{z}^{(k)}_{\tau_t}-n\lambda_2\mathbf{K}^{\top}
          \bs{\alpha}^{(k)}_{\tau_t}+n\lambda_1\mathbf{K}^{\top}\mathbf{q}^{(k)}_{\tau_{t-1}}
        \end{array}\right)\\
      &= g\left\{\mathbf{1}^{\top} \mathbf{z}^{(k)}_{\tau_t}+n\lambda_1 \mathbf{1}^{\top}
        \mathbf{q}^{(k)}_{\tau_{t-1}}\mathbf{v}^{\top} \mathbf{K}\left(\mathbf{z}^{(k)}_{\tau_t}
        + n\lambda_2 \bs{\alpha}^{(k)}_{\tau_t}+n\lambda_1
        \mathbf{q}^{(k)}_{\tau_{t-1}} \right)\right\}
         \left(\begin{array}{c}1 \\
           -\mathbf{v}\end{array}\right)\\
      &\quad+
        \left(\begin{array}{c}
          0 \\
          \mathbf{U} \bs{\Pi}_{\gamma, \lambda_1, \lambda_2}^{-1} \bs{\Lambda}
          \mathbf{U}^{\top}\left(\mathbf{z}^{(k)}_{\tau_t}+n\lambda_2
          \bs{\alpha}^{(k)}_{\tau_t}+n\lambda_1 \mathbf{q}^{(k)}_{\tau_{t-1}}\right)
        \end{array}\right).
    \end{aligned}
  \end{equation}
\end{itemize}

\section{Alternative smoothing surrogates}
\textbf{Moreau envelope.}\enskip{}This approach smooths a function by
introducing a quadratic regularization term. Given a function $f(t)$, the Moreau
envelope $f_\gamma(t)$ is defined as
$f_\gamma(t)=\inf _y\left\{f(y)+\frac{1}{2 \gamma}\|t-y\|_2^2\right\}.$ Applying
this concept to the check loss function $\rho_\tau(t)$ yields
\begin{equation*}
  H_{\gamma,\tau}(t)  =\inf _y\left\{\rho_\tau(y)+\frac{1}{2 \gamma}\|t-y\|_2^2\right\}  = \begin{cases} t(\tau-1)-\frac{1}{2}(\tau-1)^2 \gamma & \text { if } \quad t < (\tau-1) \gamma, \\ \frac{t^2}{2 \gamma} & \text { if } \quad(\tau-1) \gamma \leq t \leq \tau \gamma, \\ t \tau-\frac{1}{2} \tau^2 \gamma & \text { if } t > \tau \gamma.\end{cases}
\end{equation*}

\noindent{}\textbf{Nesterov's smoothing.}\enskip{}
Consider a function $f(t)$, Nesterov's smoothing approach
\citep{nesterov2005smooth} constructs a smooth approximation
$f_\gamma(t)=\sup _{x \in \operatorname{dom}(g)}\langle t, x\rangle-(g(x)+\gamma d(x)),$
where $g(x) $ is the conjugate function of $f(x)$ and $d(x)$ is a prox-function.
    
Applying this to the check loss $\rho_\tau(t)$ with
$g(x)=\max _{z \in \mathbb{R}}\{x z-z(p-\mathbf{1}(z<0))\}$ and setting
$d(x)=\frac{1}{2}\|x\|_2^2$ yields
\begin{equation*}
  H_{\gamma,\tau}(t)=\left\{
    \begin{array}{ll}
      t(\tau-1)-\frac{1}{2}(\tau-1)^{2}\gamma&\text{if}\enskip{}t\leq(\tau-1)\gamma,\\
      \frac{t^{2}}{2\gamma}&\text{if}\enskip{}(\tau-1)\gamma\leq{t}\leq\tau\gamma,\\
      t\tau-\frac12\tau^{2}\gamma&\text{if}\enskip{}t\geq\tau\gamma.
    \end{array}\right.
\end{equation*}

\noindent{}\textbf{Huber approximation.}\enskip{}
Note that
$\rho_\tau(t)=(1-\tau) t_{-}+\tau t_{+}=\frac{1}{2}\{|t|+(2 \tau-1) t\}$ and the
Moreau envelope of $|t|$ is the Huber function
$h_\gamma(t)=\frac{t^2}{2 \gamma}\mathbb{I}(|t| \leq \gamma)+(|t|-\frac{\gamma}{2})\mathbb{I}(|t|>\gamma).$
Building on this relationship,~\cite{yi2017semismooth} introduce the Huberized
smooth loss function:
\begin{equation*}
  \begin{aligned}
    H_{\gamma,\tau}(t)  = \frac{1}{2}\left(h_\gamma(t)+(2\tau-1)t \right)  = \begin{cases}(\tau-1)t-\frac{\gamma}{4}  & \text { if } t<-\gamma, \\
      \frac{t^2}{4 \gamma}+t\left(\tau-\frac{1}{2}\right) & \text { if }-\gamma \leq t \leq \gamma, \\
      \tau t -\frac{\gamma}{4} & \text { if } t>\gamma,\end{cases}
  \end{aligned}
\end{equation*}
which coincides exactly with the proposed smooth check loss function up to a
constant shift.

\noindent{}\textbf{Kernel density convolution.}\enskip{}
Given a kernel function $K(\cdot)$ and bandwidth $\gamma >0$, the convolution
smoothed check loss $H_{\gamma,\tau}(t)$ is defined as
$H_{\gamma,\tau}(t)=\int_{-\infty}^{\infty} \rho_\tau(t-x) K(x) d x$. Several
common kernels illustrate this approach \citep{tan2022high, he2023smoothed}:
\begin{enumerate}
  \item (Uniform kernel) Consider the uniform kernel
        $K(u)=(1 / 2)I(|u| \leq 1)$, substituting this into the above integral
        yields
        \begin{equation*}
          \begin{aligned}
            H_{\gamma,\tau}(t)
                = \begin{cases}(\tau-1) t & \text { if } t<-\gamma, \\ \frac{t^2}{4 \gamma}+t\left(\tau-\frac{1}{2}\right)+\frac{\gamma}{4} & \text { if }-\gamma \leq t \leq \gamma, \\ \tau t & \text { if } t>\gamma,\end{cases}
          \end{aligned}
        \end{equation*}
        which is exactly the proposed smoothed check loss function.
  \item (Epanechnikov kernel) Consider the Epanechnikov kernel
        $K(u)=(3 / 4)\left(1-u^2\right)I(|u| \leq 1)$, then the smoothed check
        loss takes the form
        \begin{equation*}
          \begin{aligned}
            H_{\gamma,\tau}(t) = \begin{cases}(\tau-1) t & \text { if } t<-\gamma, \\
              -\frac{t^4}{16\gamma^3}+\frac{3t^2}{8 \gamma}+t\left(\tau-\frac{1}{2}\right)+\frac{3\gamma}{16} & \text { if }-\gamma \leq t \leq \gamma, \\ \tau t & \text { if } t>\gamma.\end{cases}
          \end{aligned}
        \end{equation*}
    \end{enumerate}
    It can be easily shown that all the aforementioned smoothed loss functions
    satisfy the gradient conditions stated in
    Theorem~\eqref{thm:S0given_general}; as such, they can recover the exact
    solution through the finite smoothing algorithm.

\section{Additional Simulation Results}

In this section, we use the same simulation model in
\cite{friedman2010regularization} to further demonstrate the performance of
\texttt{fastkqr}. Predictors are generated from $\mathrm{N}(0, 1)$, where each
pair is correlated with $\rho=0.1$. The response values were generated by
\begin{equation}\label{eq:11}
Y=\sum_{j=1}^{p} X_{j} \beta_{j}+ c Z,
\end{equation}
where $\beta_{j}=(-1)^{j} \exp (-\frac{j-1}{10})$, $Z \sim \mathrm{N}(0,1)$, and
$c$ is set so that the signal-to-noise ratio is 3.0. All computations were
carried out on an Apple M1 (16GB) processor.

Table~\ref{tab1:sim-p=100} and Table~\ref{tab1:sim-p=5000} present the objective
values and computation time using simulation data for $p=100$ and $5000$,
respectively. Notably, \texttt{fastkqr} consistently emerges as the fastest
solver, outperforming the other three solvers by at least an order of magnitude,
while also achieving the lowest objective value.

\begin{table}[p]
  \centering
  \resizebox{1.0\textwidth}{!}{%
  \begin{tabular}{rrrrrrrrrrr}
    \toprule
    $\tau$ & $n$ & &\texttt{fastkqr} & \texttt{kernlab} & \texttt{clarabel} & \texttt{ADMM} & \texttt{osqp} & \texttt{nlm} & \texttt{optim} \\
    \midrule
    0.1 & 200 & \text{obj} & $0.601_{(0.054)}$ & $0.601_{(0.054)}$ & $0.601_{(0.054)}$ & $0.601_{(0.054)}$ & $0.601_{(0.054)}$ & $0.601_{(0.054)}$ & $0.611_{(0.051)}$ \\
    & & \text{time} & 0.31 & 5.87 & 21.43 & 174.99 & 8.92 & 217.40 & 452.91 \\
    & 500 & \text{obj} & $0.563_{(0.033)}$ & $0.563_{(0.033)}$ & $0.563_{(0.033)}$ & $0.563_{(0.033)}$ & $0.563_{(0.033)}$ & $0.573_{(0.031)}$ & $0.593_{(0.032)}$ \\
    & & \text{time} & 1.87 & 53.58 & 328.30 & 2368.89 & 162.54 & 3492.46 & 6785.29 \\
    & 1000 & \text{obj} & $0.539_{(0.019)}$ & $0.539_{(0.019)}$ & $0.539_{(0.019)}$ & $0.539_{(0.019)}$ & $0.539_{(0.019)}$ & $0.561_{(0.016)}$ & $0.578_{(0.024)}$ \\
    & & \text{time} & 9.07 & 368.41 & 3264.13 & 10643.25 & 635.71 & 28046.38 & 56038.72 \\
    0.5 & 200 & \text{obj} & $0.939_{(0.224)}$ & $0.939_{(0.224)}$ & $0.939_{(0.224)}$ & $0.939_{(0.224)}$ & $0.939_{(0.224)}$ & $1.021_{(0.162)}$ & $1.157_{(0.131)}$ \\
    & & \text{time} & 0.28 & 6.40 & 17.07 & 294.33 & 10.43 & 212.05 & 450.62 \\
    & 500 & \text{obj} & $0.953_{(0.144)}$ & $0.953_{(0.144)}$ & $0.953_{(0.144)}$ & $0.953_{(0.144)}$ & $0.953_{(0.144)}$ & $1.089_{(0.100)}$ & $1.218_{(0.075)}$ \\
    & & \text{time} & 1.62 & 55.06 & 243.30 & 2441.78 & 175.92 & 3215.64 & 6732.79 \\
    & 1000 & \text{obj} & $1.009_{(0.087)}$ & $1.009_{(0.087)}$ & $1.009_{(0.087)}$ & $1.009_{(0.087)}$ & $1.009_{(0.087)}$ & $1.154_{(0.046)}$ & $1.232_{(0.055)}$ \\
    & & \text{time} & 7.93 & 362.85 & 2569.78 & 10827.59 & 770.64 & 27356.19 & 54737.99 \\
    0.9 & 200 & \text{obj} & $0.585_{(0.041)}$ & $0.585_{(0.041)}$ & $0.585_{(0.041)}$ & $0.585_{(0.041)}$ & $0.585_{(0.041)}$ & $0.586_{(0.040)}$ & $0.598_{(0.038)}$ \\
    & & \text{time} & 0.32 & 5.54 & 21.90 & 162.68 & 9.14 & 219.43 & 451.78 \\
    & 500 & \text{obj} & $0.554_{(0.029)}$ & $0.554_{(0.029)}$ & $0.554_{(0.029)}$ & $0.554_{(0.029)}$ & $0.554_{(0.029)}$ & $0.566_{(0.026)}$ & $0.591_{(0.037)}$  \\
    & & \text{time} & 1.87 & 51.74 & 324.96 & 2424.24 & 161.57 & 3363.38 & 6746.65 \\
    & 1000 & \text{obj} & $0.541_{(0.017)}$ & $0.541_{(0.017)}$ & $0.541_{(0.017)}$ & $0.541_{(0.017)}$ & $0.541_{(0.017)}$ & $0.565_{(0.016)}$ & $0.581_{(0.016)}$ \\
    & & \text{time} & 8.89 & 366.14 & 3599.66 & 10632.72 & 620.84 & 28308.27 & 55795.43 \\
  \bottomrule
  \end{tabular}}
  \caption{Objective values and computation time of seven kernel quantile regression solvers for simulation data \citep{friedman2010regularization} with $p=100$, $n=\{200, 500, 1000\}$, and $\tau=\{0.1,  0.5, 0.9\}$. The numbers are the average quantities over 20 independent runs
    and the standard errors are presented in the parentheses.}
  \label{tab1:sim-p=100}
\end{table}

\begin{table}[p]
  \centering
  \resizebox{1.0\textwidth}{!}{%
  \begin{tabular}{rrrrrrrrrrr}
    \toprule
    $\tau$ & $n$ & &\texttt{fastkqr} & \texttt{kernlab} & \texttt{clarabel} & \texttt{ADMM} & \texttt{osqp} & \texttt{nlm} & \texttt{optim} \\
    \midrule
    0.1 & 200 & \text{obj} & $0.634_{(0.039)}$ & $0.634_{(0.039)}$ & $0.634_{(0.039)}$ & $0.634_{(0.039)}$ & $0.634_{(0.039)}$ & $0.634_{(0.039)}$ & $0.693_{(0.166)}$ \\
    & & \text{time} & 1.41 & 42.73 & 85.60 & 138.07 & 66.34 & 227.24 & 523.11 \\
    & 500 & \text{obj} & $0.632_{(0.027)}$ & $0.631_{(0.027)}$ & $0.631_{(0.027)}$ & $0.672_{(0.194)}$ & $0.631_{(0.027)}$ & $0.632_{(0.027)}$ & $0.639_{(0.034)}$  \\
    & & \text{time} & 8.35 & 228.24 & 636.40 & 2298.51 & 497.47 & 2987.70 & 7117.52\\
    & 1000 & \text{obj} & $0.641_{(0.021)}$ & $0.641_{(0.021)}$ & $0.641_{(0.021)}$ & $0.641_{(0.021)}$ & $0.641_{(0.021)}$ & $0.641_{(0.021)}$ & $0.682_{(0.119)}$ \\
    & & \text{time} & 62.32 & 1013.36 & 6601.43 & 11948.91 & 3372.22 & 24770.94 & 59033.30\\
    0.5 & 200 & \text{obj} & $1.061_{(0.439)}$ & $1.061_{(0.439)}$ & $1.061_{(0.439)}$ & $1.061_{(0.439)}$ & $1.061_{(0.439)}$ & $1.077_{(0.416)}$ & $1.227_{(0.245)}$ \\
    & & \text{time} & 1.55 & 59.04 & 79.22 & 412.30 & 73.56 & 261.21 & 525.72 \\
    & 500 & \text{obj} & $0.978_{(0.335)}$ & $0.978_{(0.335)}$ & $0.978_{(0.335)}$ & $0.978_{(0.335)}$ & $0.978_{(0.335)}$ & $1.021_{(0.295)}$ & $1.252_{(0.164)}$ \\
    & & \text{time} & 8.25 & 283.25 & 563.51 & 2705.82 & 517.18 & 2897.41 & 6925.84 \\
    & 1000 & \text{obj} & $1.059_{(0.214)}$ & $1.059_{(0.213)}$ & $1.059_{(0.213)}$ & $1.059_{(0.213)}$ & $1.059_{(0.213)}$ & $1.103_{(0.184)}$ & $1.305_{(0.134)}$ \\
    & & \text{time} & 62.89 & 1181.56 & 4968.14 & 13406.95 & 3333.52 & 27473.93 & 59327.84\\
    0.9 & 200 & \text{obj} & $0.610_{(0.037)}$ & $0.610_{(0.037)}$ & $0.610_{(0.037)}$ & $0.610_{(0.037)}$ & $0.610_{(0.037)}$ & $0.610_{(0.037)}$ & $0.617_{(0.047)}$  \\
    & & \text{time} & 1.49 & 45.33 & 89.07 & 229.38 & 68.94 & 231.50 & 520.99  \\
    & 500 & \text{obj} & $0.638_{(0.032)}$ & $0.638_{(0.032)}$ & $0.638_{(0.032)}$ & $0.679_{(0.183)}$ & $0.638_{(0.032)}$ & $0.638_{(0.032)}$ & $0.645_{(0.036)}$ \\
    & & \text{time} & 8.31 & 224.15 & 647.39 & 1720.98 & 485.26 & 2921.25 & 7018.28 \\
    & 1000 & \text{obj} & $0.639_{(0.020)}$ & $0.639_{(0.019)}$ & $0.639_{(0.019)}$ & $0.682_{(0.201)}$ & $0.639_{(0.019)}$ & $0.639_{(0.019)}$ & $0.668_{(0.050)}$  \\
    & & \text{time} & 61.56 & 1017.84 & 5840.61 & 10796.91 & 3282.25 & 25235.31 & 58794.06 \\
  \bottomrule
  \end{tabular}}
  \caption{Objective values and computation time of seven kernel quantile regression solvers for simulation data \citep{friedman2010regularization} with $p=5000$, $n=\{200, 500, 1000\}$, and $\tau=\{0.1,  0.5, 0.9\}$. The numbers are the average quantities over 20 independent runs
    and the standard errors are presented in the parentheses.}
  \label{tab1:sim-p=5000}
\end{table}

\section{Benchmark Data Applications}\label{sec:real_data}

We used four benchmark data from the R packages \texttt{MASS} and
\texttt{mlbench} to further compare our \texttt{fastkqr} with the six
competitors: \texttt{kernlab}, \texttt{clarabel}, \texttt{ADMM}, \texttt{osqp}, \texttt{nlm}, and \texttt{optim}. The first data
set \texttt{crabs} includes five morphometric measurements from 50 crabs of the
Leptograpsus variegatus species, collected in Fremantle, Western Australia. For
our analysis, categorical variables were converted to dummy variables, with
carapace width as the response variable, and variable ``index'' was removed. The
second data set \texttt{GAGurine} (\texttt{GAG}) was introduced in
Section~\ref{sec:intro}. The third data set, \texttt{BostonHousing}
(\texttt{BH}), provides housing information in the Boston area with the median
value of owner-occupied homes as the predictor variable. The last data set,
\texttt{geyser}, records details of the ``Old Faithful" geyser in the
Yellowstone National Park, using eruption time as the predictor variable. 

For each benchmark data, we employed the five-fold cross-validation to select
the optimal tuning parameter, which was then used to calculate the objective
values for accuracy assessment. We also recorded the total computation time,
including both the model training and parameter tuning via the five-fold
cross-validation. All computations were carried out on an Apple M1 (16GB) processor. Notably, \texttt{fastkqr} consistently demonstrated superior
efficiency compared to the other six solvers. In addition, \texttt{fastkqr}, \texttt{kernlab}, \texttt{clarabel} and \texttt{osqp}
achieve the highest accuracy. The computational advantages of \texttt{fastkqr}
become increasingly evident with larger sample sizes. For instance, when fitting
the \texttt{mcycle} data with 133 observations, \texttt{fastkqr} required only
half the time needed by \texttt{kernlab}. For the larger \texttt{BostonHousing}
data consisting of 506 observations, \texttt{fastkqr} was ten times faster than
\texttt{kernlab}.

\begin{table}[t]
  \centering
  \resizebox{1.0\textwidth}{!}{%
  \begin{tabular}{rrrrrrrrrrr}
    \toprule
    $\text{data}_{(n, p)}$
    & $\tau$  && \texttt{fastkqr} & \texttt{kernlab} & \texttt{clarabel}  & \texttt{ADMM} & \texttt{osqp} & \texttt{nlm} & \texttt{optim}\\
    \midrule
    $\text{crabs}_{(200, 8)}$
    & 0.1 & \texttt{obj} & $0.204_{(0.034)}$ & $0.204_{(0.034)}$ & $0.204_{(0.034)}$ & $0.204_{(0.034)}$ & $0.205_{(0.034)}$ & $0.208_{(0.034)}$ & $1.091_{(0.292)}$ \\
    & & \texttt{time} & 1.54 & 4.39 & 23.68  & 116.94 & 7.67 & 224.93 & 458.51\\
    & 0.5 & \texttt{obj} & $0.267_{(0.049)}$ & $0.267_{(0.049)}$ & $0.267_{(0.049)}$ & $0.267_{(0.049)}$ & $0.268_{(0.049)}$ & $0.285_{(0.047)}$ & $1.668_{(0.828)}$ \\
    & & \texttt{time} & 1.45 & 4.83 & 19.15 & 214.45 & 6.95 & 218.69 & 453.75 \\
    & 0.9 & \texttt{obj} & $0.235_{(0.038)}$ & $0.235_{(0.038)}$ & $0.235_{(0.038)}$ & $0.235_{(0.038)}$ & $0.236_{(0.038)}$ & $0.238_{(0.038)}$ & $0.949_{(0.495)}$ \\
    & & \texttt{time} & 1.45 & 5.22 & 23.20 & 140.45 & 8.50 & 220.64 & 455.58 \\
    
    $\text{GAG}_{(314, 1)}$
    & 0.1 & \texttt{obj} & $0.540_{(0.013)}$ & $0.540_{(0.013)}$ & $0.540_{(0.013)}$ & $0.540_{(0.013)}$ & $0.540_{(0.013)}$ & $0.540_{(0.013)}$ & $0.578_{(0.029)}$  \\
    & & \texttt{time} & 4.77 & 24.51 & 65.32 & 845.47 & 32.98 & 639.42 & 1634.79 \\
    & 0.5 & \texttt{obj} & $1.460_{(0.047)}$ & $1.460_{(0.047)}$ & $1.460_{(0.047)}$ & $1.460_{(0.047)}$ & $1.460_{(0.047)}$ & $1.461_{(0.047)}$ & $1.500_{(0.039)}$ \\
    & & \texttt{time} & 5.21 & 19.90 & 52.99 & 895.03 & 28.07 & 720.77 & 1631.48 \\
    & 0.9 & \texttt{obj} & $0.235_{(0.038)}$ & $0.235_{(0.038)}$ & $0.235_{(0.038)}$ & $0.235_{(0.038)}$ & $0.236_{(0.038)}$ & $0.238_{(0.038)}$ & $0.949_{(0.495)}$ \\
    & & \texttt{time} & 4.72 & 21.69 & 72.08 & 729.15 & 33.98 & 562.52 & 1655.45 \\
    
    $\text{mcycle}_{(133, 1)}$
    & 0.1 & \texttt{obj} & $4.177_{(0.259)}$ & $4.177_{(0.259)}$ & $4.177_{(0.259)}$ & $4.177_{(0.259)}$ & $4.180_{(0.259)}$ & $4.196_{(0.254)}$ & $5.056_{(0.384)}$ \\
    & & \texttt{time} & 1.74 & 4.37 & 11.29 & 92.50 & 3.92 & 65.40 & 163.23 \\
    & 0.5 & \texttt{obj} & $8.565_{(0.534)}$ & $8.565_{(0.534)}$ & $8.565_{(0.534)}$ & $8.565_{(0.534)}$ & $8.567_{(0.533)}$ & $8.620_{(0.527)}$ & $9.254_{(0.557)}$ \\
    & & \texttt{time} & 1.56 & 2.45 & 8.63 & 68.33 & 2.71 & 77.05 & 163.75 \\
    & 0.9 & \texttt{obj} & $3.773_{(0.141)}$ & $3.773_{(0.141)}$ & $3.773_{(0.141)}$ & $3.773_{(0.141)}$ & $3.775_{(0.141)}$ & $3.781_{(0.140)}$ & $4.137_{(0.250)}$ \\
    & & \texttt{time} & 1.32 & 4.65 & 10.44 & 91.25 & 3.93 & 66.97 & 162.88 \\
    
    $\text{BH}_{(506, 14)}$
    & 0.1 & \texttt{obj} & $0.490_{(0.020)}$ & $0.490_{(0.020)}$ & $0.490_{(0.020)}$ & $0.490_{(0.020)}$ & $0.490_{(0.019)}$ & $0.498_{(0.019)}$ & $0.700_{(0.056)}$ \\
    & & \texttt{time} & 6.17 & 70.67 & 360.05 & 2427.07 & 136.88 & 3425.43 & 6988.44 \\
    & 0.5 & \texttt{obj} & $1.022_{(0.013)}$ & $1.022_{(0.013)}$ & $1.022_{(0.013)}$ & $1.022_{(0.013)}$ & $1.023_{(0.013)}$ & $1.165_{(0.058)}$ & $1.740_{(0.547)}$ \\
    & & \texttt{time} & 5.92 & 58.45 & 302.34 & 2657.78 & 90.87 & 3503.75 & 6963.22 \\
    & 0.9 & \texttt{obj} & $0.671_{(0.002)}$ & $0.671_{(0.002)}$ & $0.671_{(0.002)}$ & $0.671_{(0.002)}$ & $0.671_{(0.002)}$ & $0.761_{(0.010)}$ & $1.438_{(0.377)}$  \\
    & & \texttt{time} & 6.28 & 79.92 & 405.05 & 1150.37 & 135.20 & 3434.75 & 7038.91 \\
    \bottomrule
  \end{tabular}}
  \caption{Objective values of seven KQR solvers for benchmark data with
    $\tau=\{0.1, 0.5, 0.9\}$. The numbers are the average quantities
    over 20 independent runs and the standard errors are presented in the
    parentheses.}
  \label{tab1:real}
\end{table}

For the NCKQR problem, we employed the same four benchmark data to compare
\texttt{fastkqr} algorithm with its three competitors: \texttt{cvxr},
\texttt{nlm}, and \texttt{optim}. For each benchmark data, we presented the
average objective values and the total computation time across a range of
$\lambda_2$ values and five quantile levels: 0.1, 0.3, 0.5, 0.7, and 0.9.
Table~\ref{tab:realdata-obj-nckqr2} show
that our algorithm consistently delivers the best accuracy and the computation
is the fastest. In all the examples, \texttt{fastkqr} consistently outperforms
the other competitors, being at least 80 times faster.

\begin{table}[tbp]
\centering
\begin{tabular}{llllll}
\hline
$\text{data}_{(n, p)}$ &
 & \texttt{fastkqr}&  \texttt{kernlab} &  \texttt{nlm} &  \texttt{optim} \\ \hline
$\text{crabs}_{(200, 8)}$ & \text{obj}
& $0.920_{(0.005)}$ & $1.053_{(0.024)}$ & 
$7.071_{(0.317)}$ & $12.264_{(0.441)}$ \\
& \text{time} & 13.41 & 1137.84 & 2936.02 & 6420.14\\
$\text{GAG}_{(314, 1)}$ & \text{obj}
& $5.082_{(0.083)}$ & $5.400_{(0.083)}$ & 
$6.345_{(0.180)}$ & $6.051_{(0.181)}$ \\
& \text{time} & 5.82 & 1433.54 & 10385.86 & 22994.12\\
$\text{mcycle}_{(133, 1)}$ & \text{obj}
& $30.607_{(0.476)}$ & $31.849_{(0.973)}$ & 
$35.614_{(0.936)}$ & $38.751_{(2.107)}$ \\
& \text{time} & 2.29 & 1063.62 & 1043.04 & 2269.04 \\
$\text{BH}_{(506, 14)}$ & \text{obj}
& $4.001_{(0.032)}$ & $5.414_{(0.168)}$ & $11.025_{(0.388)}$ & *\\
& \text{time} &28.87 & 2856.31 & 39887.25 & $>24$h\\
      \hline
\end{tabular}
\caption{Objective values and computation time of four NCKQR solvers for benchmark data. The numbers
  are the average quantities over 20 independent runs and the standard errors
  are presented in the parentheses. A result shown as a star ``$*$" means the
  corresponding solver cannot output a solution due to some numerical issue.}
\label{tab:realdata-obj-nckqr2}
\end{table}


\section{Technical Proofs}
In this section, we provide all technical proofs for the results presented in
the main article.

\subsection{Proof of Proposition \ref{thm:S0given}}
 To establish the proposition, first we prove the following lemma:
\begin{lemma}
Let $(b^{\gamma}, \bs{\alpha}^{\gamma})=\argmin_{b\in\mathbb{R},
  \bs{\alpha}\in\mathbb{R}^{n}} G^{\gamma}(b, \bs{\alpha})=\frac1n\sum_{i=1}^{n}
H_{\gamma,\tau}(y_{i}-b-\mathbf{K}_{i}^{\top}\bs{\alpha})
+\frac{\lambda}{2}\bs{\alpha}^{\top}\mathbf{K}\bs{\alpha}$, then we have
\begin{equation*}
  G(b^{\gamma}, \bs{\alpha}^{\gamma})-\frac{\gamma}{4}\leq{}
  G(\hat{b}, \hat{\bs{\alpha}})\leq{}
  G(b^{\gamma}, \bs{\alpha}^{\gamma}) .
\end{equation*}
\end{lemma}

\begin{proof}
  Recall that
  \begin{equation*}
    H_{\gamma,\tau}(t)=
    \begin{cases}
      (\tau-1)t
      & \text{if}\enskip{}t< -\gamma,\\
      \frac{t^{2}}{4\gamma}+t(\tau-\frac{1}{2})+\frac{\gamma}{4}
      & \text{if}\enskip{} -\gamma\leq t\leq\gamma,\\
      \tau{}t
      & \text{if}\enskip{}t>\gamma,
    \end{cases}
  \end{equation*}
  and
  \begin{equation*}
    \rho_{\tau}(t)=
    \begin{cases}
      (\tau-1)t & \text{if}\enskip{}t\leq0,\\
      \tau{}t & \text{if}\enskip{}t>0.
    \end{cases}
  \end{equation*}
  Consider the following three cases:
  \begin{itemize}
  \item Case 1. When $t<-\gamma<0$ or $t>\gamma>0$, we have
    \[
      H_{\tau,\gamma}(t)-\rho_{\tau}(t)=0.
    \]
  \item Case 2. When $-\gamma\leq t\leq0$, we have
    \[
      0 {}\leq{}H_{\tau,\gamma}(t)-\rho_{\tau}(t)
      =\frac{1}{4\gamma}(t+\gamma)^{2}{}
      \leq{}\frac{\gamma}{4}.
    \]
  \item Case 3. When $0<t\leq \gamma$, we have
    \[
      0 {}\leq{}H_{\tau,\gamma}(t)-\rho_{\tau}(t)
      =\frac{1}{4\gamma}(t-\gamma)^{2}{}<{}\frac{\gamma}{4}.
    \]
  \end{itemize}
  Thus, for any $t\in\mathbb{R}$, $0\leq{}H_{\tau,\gamma}(t)-\rho_{\tau}(t)\leq
  \gamma/4$. It follows that
  \begin{equation}\label{eq:huber-approx}
    0\leq{}G^{\gamma}(b, \bs{\alpha})-G(b, \bs{\alpha})\leq \frac{\gamma}{4},
    \forall{} b\in\mathbb{R}, \bs{\alpha}\in \mathbb{R}^{n},
  \end{equation}
  which implies $G(b^{\gamma}, \bs{\alpha}^{\gamma})
  \leq{}G^{\gamma}(b^{\gamma}, \bs{\alpha}^{\gamma})$ and $G^{\gamma}(\hat{b},
  \hat{\bs{\alpha}})\leq{}G(\hat{b}, \hat{\bs{\alpha}})+\gamma/4$. By optimality
  of $(\hat{b},\hat{\bs{\alpha}})$ and $(b^{\gamma},\bs{\alpha}^{\gamma})$, we
  have
  \[
    G(\hat{b},\hat{\bs{\alpha}})\leq{}G(b^{\gamma},\bs{\alpha}^{\gamma})
    \leq{}G^{\gamma}(b^{\gamma}, \bs{\alpha}^{\gamma})\leq{}G^{\gamma}(\hat{b},
    \hat{\bs{\alpha}}) \leq{}G(\hat{b},\hat{\bs{\alpha}})+\frac{\gamma}{4}.
  \]
\end{proof}

Now consider the following constrained problem
\begin{equation}\label{eq:huberS0}
  \begin{split}
    &\min_{b \in \mathbb{R}, \bs{\alpha}\in \mathbb{R}^n}\enskip{}
      G^{\gamma}(b, \bs{\alpha})\\
    &\text{subject to}\enskip{}
      y_{i}=b+\mathbf{K}_{i}^{\top}\bs{\alpha},\,i\in S_0.
  \end{split}
\end{equation}
The Lagrangian of problem \eqref{eq:huberS0} is
\begin{equation}
  L(b,\bs{\alpha},\xi_{i})=\frac{1}{n}\sum_{i=1}^{n}
  H_{\gamma ,{}\tau}(y_{i}-b-\mathbf{K}_{i}^{\top}\bs{\alpha})
  +\frac{\lambda}{2}\bs{\alpha}^{\top}\mathbf{K}\bs{\alpha}
  +\sum_{i\in S_0}\xi_{i}
  (b+\mathbf{K}_{i}^{\top}\bs{\alpha}-y_{i}),
\end{equation}
where the $\xi_{i}$'s are the Lagrangian multipliers. By the optimality of
$(\hat{\bs{\alpha}}^{\gamma}, \hat{b}^{\gamma})$ in problem \eqref{eq:huberS0},
we have
\begin{equation}\label{eq:huber-kkt-S0}
  \left\{
    \begin{array}{l}
      -\frac{1}{n}\sum_{ i=1}^{ n}
      H^{\prime}_{\gamma ,{}\tau}
      (y_{i}-\hat{b}^{\gamma}-\mathbf{K}_{i}^{\top} \hat{\bs{\alpha}}^{\gamma}
      )\mathbf{K}_{i}+\lambda\mathbf{K}
      \hat{\alpha}^{\gamma}+\sum_{i\in S_0}
      \xi_{i}\mathbf{K}_{i}=\mathbf{0},\\
      -\frac{1}{n}\sum_{ i=1}^{ n}
      H^{\prime}_{\gamma ,{}\tau}
      (y_{i}-\hat{b}^{\gamma}-\mathbf{K}_{i}^{\top} \hat{\bs{\alpha}}^{\gamma}
      )+\sum_{i\in S_0}
      \xi_{i}=0,\\
      y_{i}=\hat{b}^{\gamma}+\mathbf{K}_{i}^{\top}\hat{\bs{\alpha}}^{\gamma}
      ,\,i\in S_0,
    \end{array}
  \right.
\end{equation}
where $H'_{\gamma,\tau}(\cdot)$ denotes the derivative of
$H_{\gamma,\tau}(\cdot)$. Note that the subdifferential of $\rho_{\tau}(\cdot)$
is
\[
  \partial\rho_{\tau}(t)=
  \begin{cases}
    \{\tau-1\}, & \text{if}\enskip{}t<0,\\
    [\tau-1,\tau], & \text{if}\enskip{}t=0,\\
    \{\tau\}, & \text{if}\enskip{}t>0.
  \end{cases}
\]
According to the definition of $H_{\gamma ,{}\tau}(t)$, one has
$H'_{\gamma,{}\tau}(y_{i}-
\hat{b}^{\gamma}-\mathbf{K}_{i}^{\top}\hat{\bs{\alpha}}^{\gamma})
=\partial\rho_{\tau}(y_{i}-\hat{b}^{\gamma}-\mathbf{K}_{i}^{\top}
\hat{\bs{\alpha}}^{\gamma})$ when $i\notin S_0$ and $H^{\prime}_{\gamma ,{}\tau}
(y_{i}-\hat{b}^{\gamma}-\mathbf{K}_{i}^{\top}
\hat{\bs{\alpha}}^{\gamma})=\tau-1/2 {}\in{}\partial\rho_{\tau}(y_{i}
-\hat{b}^{\gamma}-\mathbf{K}_{i}^{\top} \hat{\bs{\alpha}}^{\gamma} )$ when $i\in
S_0$. Therefore, it follows from expression \eqref{eq:huber-kkt-S0} that
\begin{equation}\label{eq:huber-kkt}
  \left\{
    \begin{array}{l}
      \mathbf{0} \in -\frac1n\sum_{i=1}^{n}\partial\rho_{\tau}
      (y_{i}-\hat{b}^{\gamma}-\mathbf{K}_{i}^{\top}\hat{\bs{\alpha}}^{\gamma}
      )\mathbf{K}_{i}+\lambda\mathbf{K}
      \hat{\alpha}^{\gamma}+\sum_{i\in S_0}
      \xi_{i}\mathbf{K}_{i},\\
      0 \in -\frac1n\sum_{i=1}^{n}\partial\rho_{\tau}
      (y_{i}-\hat{b}^{\gamma}-\mathbf{K}_{i}^{\top}
      \hat{\bs{\alpha}}^{\gamma})
      +\sum_{i\in S_0}\xi_{i}.
    \end{array}
  \right.
\end{equation}
We can see that $(\hat{b}^{\gamma}, \hat{\bs{\alpha}}^{\gamma})$ satisfies the
KKT conditions of the convex problem:
\begin{equation}\label{eq:constr-kqr-S0}
  \begin{split}
    &\min_{b \in \mathbb{R}, \bs{\alpha}\in \mathbb{R}^n}\enskip{}
      \frac1n\sum_{i=1}^{n}\rho_{\tau}(y_{i}-b-\mathbf{K}_{i}^{\top}\bs{\alpha}
      )+\frac{\lambda}{2}\bs{\alpha}^{\top}\mathbf{K}\bs{\alpha}\\
    &\text{subject to}\enskip{}
      y_{i}=b+\mathbf{K}_{i}^{\top}\bs{\alpha},\,i\in S_0.
  \end{split}
\end{equation}
Therefore, $(\hat{b}^{\gamma}, \hat{\bs{\alpha}}^{\gamma})$ is the optimal
solution of problem~\eqref{eq:constr-kqr-S0}. This implies that
\[
  G(\hat{b}^{\gamma}, \hat{\bs{\alpha}}^{\gamma})
  \leq{}G(\hat{b}, \hat{\bs{\alpha}})\leq{}
  G(\hat{b}^{\gamma}, \hat{\bs{\alpha}}^{\gamma}),
\]
The first inequality is justified because $(\hat{b}, \hat{\bs{\alpha}})$ is a feasible point of problem~\eqref{eq:constr-kqr-S0}. The second inequality follows since
$(\hat{b}, \hat{\bs{\alpha}})$ is the unconstrained minimizer of $G(b,
\bs{\alpha})$. Thus, we have $G(\hat{b}^{\gamma},
\hat{\bs{\alpha}}^{\gamma})=G(\hat{b}, \hat{\bs{\alpha}})$. Given the uniqueness
of $(\hat{b}, \hat{\bs{\alpha}})$, it follows that $(\hat{b}^{\gamma},
\hat{\bs{\alpha}}^{\gamma})=(\hat{b}, \hat{\bs{\alpha}})$.

\subsection{Proof of Theorem \ref{thm:update_set}}
Since $S\subseteq{}S_{0}$, we have
$y_{i}=\hat{b}+\mathbf{K}_{i}^{\top}\hat{\bs{\alpha}}$ for $i\in{}S$. Therefore,
$(\hat{b}, \hat{\bs{\alpha}})$ is a feasible point of the minimization problem
\[
  \begin{split}
    \min_{b \in \mathbb{R}, \bs{\alpha}\in \mathbb{R}^n}\enskip{}
    & G^{\gamma}(b, \bs{\alpha})\\
    \text{subject to}\enskip{}
    & y_{i}=b+ \mathbf{K}_{i}^{\top}\bs{\alpha},i\in{}S.
  \end{split}
\]
By the optimality of $(\tilde{b}^{\gamma}, \tilde{\bs{\alpha}}^{\gamma})$, we
have $G^{\gamma}(\tilde{b}^{\gamma}, \tilde{\bs{\alpha}}^{\gamma})\leq{}
G^{\gamma}(\hat{b}, \hat{\bs{\alpha}})$.
It then follows from inequality~\eqref{eq:huber-approx}  that
\begin{equation*}
  \begin{split}
    &G(\tilde{b}^{\gamma}, \tilde{\bs{\alpha}}^{\gamma})
      -G(\hat{b}, \hat{\bs{\alpha}})\\
    &=[G(\tilde{b}^{\gamma}, \tilde{\bs{\alpha}}^{\gamma})
      -G^{\gamma}(\tilde{b}^{\gamma}, \tilde{\bs{\alpha}}^{\gamma})]
      +[G^{\gamma}(\tilde{b}^{\gamma}, \tilde{\bs{\alpha}}^{\gamma})
      -G^{\gamma}(\hat{b}, \hat{\bs{\alpha}})]
      +[G^{\gamma}(\hat{b}, \hat{\bs{\alpha}})
      -G(\hat{b}, \hat{\bs{\alpha}})]\\
    &\leq 0+0+\frac{\gamma}{4}
      < \frac{\gamma^{*}}{4}
      \leq \rho.
  \end{split}
\end{equation*}
By the definition of $D_{\gamma_{0}/2}$, we know
$(\tilde{\bs{\alpha}}^{\gamma},\tilde{b}^{\gamma})\notin{}D_{\gamma_{0}/2}$, and
therefore,
\[
  |\tilde{b}^{\gamma}+\mathbf{K}_{ i}^{ \top}\tilde{\bs{\alpha}}^{\gamma}
  -\hat{b}-\mathbf{K}_{ i}^{ \top}
  \hat{\bs{\alpha}}|<\frac{\gamma_{0}}{2},\forall{}i=1,\ldots,n.
\]
 This implies that for any $i\in \mathcal{E}(S)$,
\begin{equation*}
  |y_{i}-\hat{b}-\mathbf{K}_{i}^{ \top}\hat{\bs{\alpha}}|\leq{}
  |y_{i}-\tilde{b}^{\gamma}-\mathbf{K}_{i}^{ \top}\tilde{\bs{\alpha}}^{\gamma}
  |+|\tilde{b}^{\gamma}+\mathbf{K}_{i}^{ \top}
  \tilde{\bs{\alpha}}^{\gamma}-\hat{b}
  -\mathbf{K}_{i}^{ \top}\hat{\bs{\alpha}}|
  <\gamma+\frac{\gamma_{0}}{2}<\gamma^{*}+\frac{\gamma_{0}}{2}<\gamma_{0},
\end{equation*}
which, by the definition of $\gamma_{0}$, further implies that $i\in{}S_{0}$. We
conclude that $S\subseteq\mathcal{E}(S)\subseteq{}S_{0}$.

\subsection{Proof of Theorem \ref{thm:subset}}
Note that since $S\subseteq{}S_{0}$ and $\gamma\in(0,\gamma^{*})$, by Theorem
\ref{thm:update_set}, we have
$S\subseteq\tilde{S}_{0}^{\gamma}\subseteq{}S_{0}$. Similarly, for any
$j\in\mathbb{N}^{+}$, if $\tilde{S}_{j-1}^{\gamma}\subseteq{}S_0$, then by
Theorem \ref{thm:update_set} again, we have $\tilde{S}_{j-1}^{\gamma}\subseteq
\tilde{S}_{j}^{\gamma}\subseteq{}S_{0}$. Therefore, by induction, we have
\[
  S\subseteq\tilde{S}_{0}^{\gamma}\subseteq\tilde{S}_{1}^{\gamma}
  \subseteq\tilde{S}_{2}^{\gamma}\subseteq\cdots\subseteq{}S_{0}.
\]
Since $S_{0}$ is a finite set, there exists a finite $j^{*}\in\mathbb{N}^{+}$
such that $\tilde{S}_{j^{*}-1}^{\gamma}=\tilde{S}_{j^{*}}^{\gamma}
{}\subseteq{}S_{0}.$ At this moment $\tilde{S}_{j^{*}}^{\gamma}$ is the set,
$\hat{S}$, we want to find.

Consider the following constrained problem
\begin{equation}\label{eq:constr-huber}
  \begin{split}
    \min_{b \in \mathbb{R}, \bs{\alpha} \in \mathbb{R}^n}\enskip{}
    &G^{\gamma}(b, \bs{\alpha})=\frac1n\sum_{i=1}^{n}
      H_{\gamma,\tau}(y_{i}-b-\mathbf{K}_{i}^{\top}\bs{\alpha})
      +\frac{\lambda}{2}\bs{\alpha}^{\top}\mathbf{K}\bs{\alpha}\\
    \text{subject to}\enskip{}
    &y_{i}=b+\mathbf{K}_{i}^{\top}\bs{\alpha},\,i\in \hat{S}.
  \end{split}
\end{equation}
Note that the minimizer of problem \eqref{eq:constr-huber} is
$(\hat{b}^{\gamma}, \hat{\bs{\alpha}}^{\gamma})$. Therefore,
$y_{i}=\hat{b}^{\gamma} + \mathbf{K}_{i}^{\top} \hat{\bs{\alpha}}^{\gamma} $ for
$i\in\hat{S}$. The Lagrangian of problem \eqref{eq:constr-huber} is
\begin{equation}
  L(b, \bs{\alpha},\xi_{i})=\frac{1}{n}\sum_{i=1}^{n}
  H_{\gamma ,{}\tau}(y_{i}-b-\mathbf{K}_{i}^{\top}\bs{\alpha})
  +\frac{\lambda}{2}\bs{\alpha}^{\top}\mathbf{K}\bs{\alpha}
  +\sum_{i\in \hat{S}}\xi_{i}
  (b+\mathbf{K}_{i}^{\top}\bs{\alpha}-y_{i}),
\end{equation}
where the $\xi_{i}$'s are the Lagrangian multipliers. Since $( \hat{b}^{\gamma},
\hat{\bs{\alpha}}^{\gamma})$ is the optimal solution of problem
\eqref{eq:constr-huber}, we have
\begin{equation}\label{eq:huber-kkt2}
  \left\{
    \begin{array}{l}
      -\frac{1}{n}\sum_{ i=1}^{ n}
      H^{\prime}_{\gamma ,{}\tau}
      (y_{i}-\hat{b}^{\gamma}-\mathbf{K}_{i}^{\top} \hat{\bs{\alpha}}^{\gamma}
      )\mathbf{K}_{i}+\lambda\mathbf{K}
      \hat{\alpha}^{\gamma}+\sum_{i\in \hat{S}}
      \xi_{i}\mathbf{K}_{i}=\mathbf{0},\\
      -\frac{1}{n}\sum_{ i=1}^{ n}
      H^{\prime}_{\gamma ,{}\tau}
      (y_{i}-\hat{b}^{\gamma}-\mathbf{K}_{i}^{\top} \hat{\bs{\alpha}}^{\gamma}
      )+\sum_{i\in \hat{S}}
      \xi_{i}=0,\\
      y_{i}=\hat{b}^{\gamma}+\mathbf{K}_{i}^{\top}\hat{\bs{\alpha}}^{\gamma}, i\in\hat{S},
    \end{array}
  \right.
\end{equation}
where $H'_{\gamma,\tau}(\cdot)$ denotes the derivative of
$H_{\gamma,\tau}(\cdot)$. Recall that
$H'_{\gamma,{}\tau}(y_{i}-\hat{b}^{\gamma}-\mathbf{K}_{i}^{\top}\hat{\bs{\alpha}}^{\gamma}
)=\partial\rho_{\tau}(y_{i}-\hat{b}^{\gamma}-\mathbf{K}_{i}^{\top}
\hat{\bs{\alpha}}^{\gamma})$ when $i\notin{\hat{S}}$ and $H^{\prime}_{\gamma
  ,{}\tau} (y_{i}-\hat{b}^{\gamma}-\mathbf{K}_{i}^{\top}
\hat{\bs{\alpha}}^{\gamma})=\tau-1/2 {}\in{}\partial\rho_{\tau}(y_{i}-
\hat{b}^{\gamma} -\mathbf{K}_{i}^{\top} \hat{\bs{\alpha}}^{\gamma} )$ when $i\in
\hat{S}$. Therefore, it follows from the first line of expression
\eqref{eq:huber-kkt2} that
\begin{equation}\label{eq:kqr-kkt1}
  \begin{split}
    \mathbf{0}
    &=-\frac1n\sum_{i\notin\hat{S}}
      H'_{\gamma,\tau}(y_{i}-\hat{b}^{\gamma}-\mathbf{K}_{i}^{\top}
      \hat{\bs{\alpha}}^{\gamma})\mathbf{K}_{i}-\frac1n\sum_{i\in\hat{S}}
      H'_{\gamma,\tau}(y_{i}-\hat{b}^{\gamma}-\mathbf{K}_{i}^{\top}
      \hat{\bs{\alpha}}^{\gamma})\mathbf{K}_{i}+\lambda\mathbf{K}
      \hat{\alpha}^{\gamma}+\sum_{i\in\hat{S}}\xi_{i}\mathbf{K}_{i}\\
    &\in-\frac1n\sum_{i\notin\hat{S}}
      \partial\rho_{\tau}(y_{i}-\hat{b}^{\gamma}-\mathbf{K}_{i}^{\top}
      \hat{\bs{\alpha}}^{\gamma})\mathbf{K}_{i}-\frac1n\sum_{i\in\hat{S}}
      \partial\rho_{\tau}(y_{i}-\hat{b}^{\gamma}-\mathbf{K}_{i}^{\top}
      \hat{\bs{\alpha}}^{\gamma})\mathbf{K}_{i}+\lambda\mathbf{K}
      \hat{\alpha}^{\gamma}+\sum_{i\in\hat{S}}\xi_{i}\mathbf{K}_{i}\\
    &=-\frac1n\sum_{i=1}^{n}\partial\rho_{\tau}
      (y_{i}-\hat{b}^{\gamma}-\mathbf{K}_{i}^{\top}
      \hat{\bs{\alpha}}^{\gamma})\mathbf{K}_{i}+\lambda\mathbf{K}
      \hat{\alpha}^{\gamma}+\sum_{i\in\hat{S}}\xi_{i}\mathbf{K}_{i}.
  \end{split}
\end{equation}
Similarly, it can be shown from the second line of display \eqref{eq:huber-kkt2}
that
\begin{equation}
  \label{eq:kqr-kkt2}
  0=-\frac{1}{n}\sum_{i=1}^{n}
  H'_{\gamma,\tau}(y_{i}-\hat{b}^{\gamma}-\mathbf{K}_{i}^{\top}
  \hat{\bs{\alpha}}^{\gamma})
  +\sum_{i\in \hat{S}}\xi_{i}
  \in-\frac1n\sum_{i=1}^{n}\partial\rho_{\tau}
  (y_{i}-\hat{b}^{\gamma}-\mathbf{K}_{i}^{\top}
  \hat{\bs{\alpha}}^{\gamma})
  +\sum_{i\in\hat{S}}\xi_{i}.
\end{equation}
Now consider the constrained problem
\begin{equation}\label{eq:constr-kqr}
  \begin{split}
    \min_{b \in \mathbb{R}, \bs{\alpha}  \in \mathbb{R}^n}\enskip{}
    & \frac1n\sum_{i=1}^{n}\rho_{\tau}(y_{i}-b-\mathbf{K}_{i}^{\top}\bs{\alpha})
      +\frac{\lambda}{2}\bs{\alpha}^{\top}\mathbf{K}\bs{\alpha}\\
    \text{subject to}\enskip{}
    & y_{i}=b+\mathbf{K}_{i}^{\top}\bs{\alpha}+b,\,i\in\hat{S}.
  \end{split}
\end{equation}
We can see from expressions~\eqref{eq:kqr-kkt1}, \eqref{eq:kqr-kkt2} and the third line of
expression~\eqref{eq:huber-kkt2} that $(\hat{b}^{\gamma},
\hat{\bs{\alpha}}^{\gamma})$ satisfies the KKT conditions of the convex problem~\eqref{eq:constr-kqr}. Therefore, $(\hat{b}^{\gamma},
\hat{\bs{\alpha}}^{\gamma})$ is the minimizer of problem~\eqref{eq:constr-kqr}.
Moreover, it can be easily seen that $(\hat{b}, \hat{\bs{\alpha}})$ is a
feasible point of problem~\eqref{eq:constr-kqr} since $\hat{S}\subseteq{}S_{0}$. This implies that
\[
  G(\hat{b}^{\gamma}, \hat{\bs{\alpha}}^{\gamma})
  \leq{}G(\hat{b}, \hat{\bs{\alpha}})\leq{}
  G(\hat{b}^{\gamma}, \hat{\bs{\alpha}}^{\gamma}),
\]
where the second inequality follows from the fact that $(\hat{b},
\hat{\bs{\alpha}})$ is the unconstrained minimizer of $G(b, \bs{\alpha})$. Thus, we have $G(\hat{b}^{\gamma}, \hat{\bs{\alpha}}^{\gamma})=G(\hat{b},
\hat{\bs{\alpha}})$. By the uniqueness of $(\hat{b}, \hat{\bs{\alpha}})$, we obtain $(\hat{b}^{\gamma}, \hat{\bs{\alpha}}^{\gamma})=(\hat{b},
\hat{\bs{\alpha}})$.

\subsection{Proof of Proposition \ref{thm:givenS02}}
 To establish the proposition, first we prove the following lemma:
 \begin{lemma}
     Let $\{b^{\gamma}_{\tau_t},
\bs{\alpha}^{\gamma}_{\tau_t}\}^T_{t=1}=\argmin_{\{b_{\tau_t},
  \bs{\alpha}_{\tau_t}\}^T_{t=1}\\}Q^\gamma\left(\{b_{\tau_t},
  \bs{\alpha}_{\tau_t}\}^T_{t=1}\right)$, then we have
\begin{equation*}
  Q\left(\{b^{\gamma}_{\tau_t},
    \bs{\alpha}^{\gamma}_{\tau_t}\}^T_{t=1}\right)
  -\frac{T}{4}\gamma\leq{}Q\left(\{\hat{b}_{\tau_t},
    \hat{\bs{\alpha}}_{\tau_t}\}^T_{t=1}\right)\leq{}
  Q\left(\{b^{\gamma}_{\tau_t},
    \bs{\alpha}^{\gamma}_{\tau_t}\}^T_{t=1}\right).
\end{equation*}
 \end{lemma}

\begin{proof}
  Based on the proof of Proposition \ref{thm:S0given}, we have
  \begin{equation}\label{eq:huber-relu-approx}
    0 \leq Q^{\gamma}\left(\{b_{\tau_t}, \bs{\alpha}_{\tau_t}\}^T_{t=1}\right)
    -Q\left(\{b_{\tau_t}, \bs{\alpha}_{\tau_t}\}^T_{t=1}\right)
    =\sum^{T}_{t=1} \left[G^{\gamma}(\bs{\alpha}_{\tau_t},b_{\tau_t})
      -G(\bs{\alpha}_{\tau_t},b_{\tau_t})\right] \leq \frac{T}{4} \gamma,
  \end{equation}
  which implies $$0 \leq Q^{\gamma}\left(\{b^{\gamma}_{\tau_t},
    \bs{\alpha}^{\gamma}_{\tau_t}\}^T_{t=1}\right)-
  Q\left(\{b^{\gamma}_{\tau_t}, \bs{\alpha}^{\gamma}_{\tau_t}\}^T_{t=1}\right)
  \leq \frac{T}{4} \gamma$$ and $$Q\left(\{\hat{b}_{\tau_t},
    \hat{\bs{\alpha}}_{\tau_t}\}^T_{t=1}\right) \leq
  Q^{\gamma}\left(\{\hat{b}_{\tau_t},
    \hat{\bs{\alpha}}_{\tau_t}\}^T_{t=1}\right).$$ By the optimality of
  $\{\hat{b}_{\tau_t}, \hat{\bs{\alpha}}_{\tau_t}\}^T_{t=1}$ and
  $\{b^{\gamma}_{\tau_t}, \bs{\alpha}^{\gamma}_{\tau_t}\}^T_{t=1}$, we have
  \begin{equation*}
    \begin{split}
      &Q\left(\{\hat{b}_{\tau_t}, \hat{\bs{\alpha}}_{\tau_t}\}^T_{t=1}\right)
      \leq{}Q\left(\{b^{\gamma}_{\tau_t},
      \bs{\alpha}^{\gamma}_{\tau_t}\}^T_{t=1}\right)
      \leq{}Q^{\gamma}\left(\{b^{\gamma}_{\tau_t},
      \bs{\alpha}^{\gamma}_{\tau_t}\}^T_{t=1}\right)\\
      &\leq{}Q^{\gamma}\left(\{\hat{b}_{\tau_t},
      \hat{\bs{\alpha}}_{\tau_t}\}^T_{t=1}\right)
      \leq{}Q\left(\{\hat{b}_{\tau_t},
      \hat{\bs{\alpha}}_{\tau_t}\}^T_{t=1}\right)+\frac{T}{4}\gamma.
    \end{split}
  \end{equation*}
\end{proof}

Consider the following constrained problem
\begin{equation}\label{eq:constr-huber-relu-S0}
  \begin{split}
    &\min_{\{b_{\tau_t}, \bs{\alpha}_{\tau_t}\}^T_{t=1}}\enskip{}
      Q^{\gamma}\left(\{b_{\tau_t}, \bs{\alpha}_{\tau_t}\}^T_{t=1}\right)\\
    & \text{subject to}\enskip{}
      y_{i}=b_{\tau_t}+\mathbf{K}_{i}^{\top}
      \bs{\alpha}_{\tau_t},\,i\in S_{0,t}, 1\leq t\leq T.
  \end{split}
\end{equation}
The Lagrangian of problem \eqref{eq:constr-huber-relu-S0} is
\begin{equation}
  \begin{split}
    L(b,\bs{\alpha},\xi)=
    & \sum^{T}_{t=1}\left[\frac{1}{n}\sum_{i=1}^{n}
      H_{\gamma, \tau}(y_{i}-b_{\tau_t}-\mathbf{K}_{i}^{\top}
      \bs{\alpha}_{\tau_t})+\frac{\lambda_2}{2}\bs{\alpha}^{\top}_{\tau_t}
      \mathbf{K}\bs{\alpha}_{\tau_t}\right]\\
    &+\sum^{T}_{t=1}\sum_{i\in S_{0,t}}\xi_{i,t}
      (b_{\tau_t}+\mathbf{K}_{i}^{\top}\bs{\alpha}_{\tau_t}-y_{i})\\
    &+\lambda_1 \sum^{T-1}_{t=1}\sum_{i=1}^{n} V(b_{\tau_t}
      +\mathbf{K}_{i}^{\top}\bs{\alpha}_{\tau_t}-b_{\tau_{t+1}}
      -\mathbf{K}_{i}^{\top}\bs{\alpha}_{\tau_{t+1}}),
  \end{split}
\end{equation}
where the $\xi_{i,t}$'s are the Lagrangian multipliers. Let
$H'_{\gamma,\tau}(\cdot)$ denote the derivative of $H_{\gamma,\tau}(\cdot)$ and
$V'(\cdot)$ denote the derivative of $V(\cdot)$. Consider the following three
cases, by the optimality of $\{\hat{b}^\gamma_{\tau_t},
\hat{\bs{\alpha}}^\gamma_{\tau_t}\}^T_{t=1}$ in problem
\eqref{eq:constr-huber-relu},
\begin{itemize}
\item Case 1. When $t=1$, we have
  \begin{equation}\label{eq:nckqr-kkt1-S0}
    \left\{
      \begin{array}{l}
        -\frac{1}{n}\sum_{i=1}^{n}
        H^{\prime}_{\gamma,{}\tau}
        (y_{i}-\hat{b}^{\gamma}_{\tau_t}-\mathbf{K}_{i}^{\top}
        \hat{\bs{\alpha}}^{\gamma}_{\tau_t})\mathbf{K}_{i}+\lambda_2\mathbf{K}
        \hat{\alpha}^{\gamma}_{\tau_t}+ \sum_{i\in S_{0,t}}
        \xi_{i,t}\mathbf{K}_{i}\\
        \quad
        +\lambda_1 \sum_{i=1}^{n} V^{\prime}(\hat{b}^{\gamma}_{\tau_t}+\mathbf{K}_{i}^{\top}
        \hat{\bs{\alpha}}^{\gamma}_{\tau_t}-\hat{b}^{\gamma}_{\tau_{t+1}}-\mathbf{K}_{i}^{\top}
        \hat{\bs{\alpha}}^{\gamma}_{\tau_{t+1}})\mathbf{K}_{i}=\mathbf{0},\\
        -\frac{1}{n}\sum_{i=1}^{n}
        H^{\prime}_{\gamma,{}\tau}
        (y_{i} -\hat{b}^{\gamma}_{\tau_t}-\mathbf{K}_{i}^{\top}\hat{\bs{\alpha}}^{\gamma}_{\tau_t}
       ) \\
        \quad +
        \lambda_2 \sum_{i=1}^{n} V^{\prime}(\hat{b}^{\gamma}_{\tau_{t}}+\mathbf{K}_{i}^{\top}
        \hat{\bs{\alpha}}^{\gamma}_{\tau_{t}}
        -\hat{b}^{\gamma}_{\tau_{t+1}}-\mathbf{K}_{i}^{\top}\hat{\bs{\alpha}}^{\gamma}_{\tau_{t+1}}
        )+\sum_{i\in S_{0,t}}\xi_{i,t}=0,\\
        y_{i}=\hat{b}^{\gamma}+\mathbf{K}_{i}^{\top}
        \hat{\bs{\alpha}}^{\gamma},\enskip{}i\in S_{0,t}.
      \end{array}
    \right.
  \end{equation}

\item Case 2. When $1<t<T$, we have

  \begin{equation}\label{eq:nckqr-kkt2-S0}
    \left\{
      \begin{array}{l}
        -\frac{1}{n}\sum_{i=1}^{n}
        H^{\prime}_{\gamma,{}\tau}
        (y_{i}-\hat{b}^{\gamma}_{\tau_t}-\mathbf{K}_{i}^{\top}\hat{\bs{\alpha}}^{\gamma}_{\tau_t}
        )\mathbf{K}_{i}+\lambda_2\mathbf{K}
        \hat{\alpha}^{\gamma}_{\tau_t}
        +\sum_{i\in S_{0,t}}
        \xi_{i,t}\mathbf{K}_{i} \\
        \quad - \lambda_1 \sum_{i=1}^{n} V^{\prime}(\hat{b}^{\gamma}_{\tau_{t-1}}
        +\mathbf{K}_{i}^{\top}\hat{\bs{\alpha}}^{\gamma}_{\tau_{t-1}}-\hat{b}^{\gamma}_{\tau_t}
        -\mathbf{K}_{i}^{\top}\hat{\bs{\alpha}}^{\gamma}_{\tau_t})\mathbf{K}_{i} \\
        \quad +\lambda_1 \sum_{i=1}^{n} V^{\prime}(\hat{b}^{\gamma}_{\tau_{t}}
        +\mathbf{K}_{i}^{\top}\hat{\bs{\alpha}}^{\gamma}_{\tau_{t}}-\hat{b}^{\gamma}_{\tau_{t+1}}
        -\mathbf{K}_{i}^{\top}\hat{\bs{\alpha}}^{\gamma}_{\tau_{t+1}})\mathbf{K}_{i}=\mathbf{0},\\
        -\frac{1}{n}\sum_{i=1}^{n}
        H^{\prime}_{\gamma,{}\tau}
        (y_{i}-\hat{b}^{\gamma}_{\tau_t}-\mathbf{K}_{i}^{\top}\hat{\bs{\alpha}}^{\gamma}_{\tau_t}
        )-
        \lambda_1 \sum_{i=1}^{n} V^{\prime}(\hat{b}^{\gamma}_{\tau_{t-1}}
        +\mathbf{K}_{i}^{\top}\hat{\bs{\alpha}}^{\gamma}_{\tau_{t-1}}
        -\hat{b}^{\gamma}_{\tau_{t}}-\mathbf{K}_{i}^{\top}\hat{\bs{\alpha}}^{\gamma}_{\tau_{t}}) \\
        \quad +
        \lambda_1 \sum_{i=1}^{n} V^{\prime}(\hat{b}^{\gamma}_{\tau_{t}}
        +\mathbf{K}_{i}^{\top}\hat{\bs{\alpha}}^{\gamma}_{\tau_{t}}
        -\hat{b}^{\gamma}_{\tau_{t+1}}-\mathbf{K}_{i}^{\top}
        \hat{\bs{\alpha}}^{\gamma}_{\tau_{t+1}})+\sum_{i\in S_{0,t}}\xi_{i,t}=0\\
        y_{i}=\hat{b}^{\gamma}+ \mathbf{K}_{i}^{\top}
        \hat{\bs{\alpha}}^{\gamma},\,i\in S_{0,t}.
      \end{array}
    \right.
  \end{equation}
\item When $t=T$, we have
  \begin{equation}\label{eq:nckqr-kkt3-S0}
    \left\{
      \begin{array}{l}
        -\frac{1}{n}\sum_{i=1}^{n} H^{\prime}_{\gamma,{}\tau}
        (y_{i}-\hat{b}^{\gamma}_{\tau_t}-\mathbf{K}_{i}^{\top}
        \hat{\bs{\alpha}}^{\gamma}_{\tau_t})\mathbf{K}_{i}
        +\lambda_2\mathbf{K}\hat{\alpha}^{\gamma}_{\tau_t}\\
        \quad-\lambda_1\sum_{i=1}^{n} V^{\prime}
        (\hat{b}^{\gamma}_{\tau_{t-1}}+\mathbf{K}_{i}^{\top}
        \hat{\bs{\alpha}}^{\gamma}_{\tau_{t-1}}-\hat{b}^{\gamma}_{\tau_t}
        -\mathbf{K}_{i}^{\top}\hat{\bs{\alpha}}^{\gamma}_{\tau_t})
        \mathbf{K}_{i}+\sum_{i\in S_{0,t}}\xi_{t,i}\mathbf{K}_{i}=\mathbf{0},\\
        -\frac{1}{n}\sum_{i=1}^{n}H^{\prime}_{\gamma,{}\tau}
        (y_{i}-\hat{b}^{\gamma}_{\tau_t}-\mathbf{K}_{i}^{\top}
        \hat{\bs{\alpha}}^{\gamma}_{\tau_t})\\
        \quad-\lambda_1\sum_{i=1}^{n} V^{\prime}
        (\hat{b}^{\gamma}_{\tau_{t-1}}+\mathbf{K}_{i}^{\top}
        \hat{\bs{\alpha}}^{\gamma}_{\tau_{t-1}}-\hat{b}^{\gamma}
        _{\tau_{t}}-\mathbf{K}_{i}^{\top}\hat{\bs{\alpha}}^{\gamma}_{\tau_{t}})
        +\sum_{i\in S_{0,t}}\xi_{i,t}=0,\\
        y_{i}=\hat{b}^{\gamma}+ \mathbf{K}_{i}^{\top}
        \hat{\bs{\alpha}}^{\gamma},\,i\in S_{0,t}.
      \end{array}
    \right.
  \end{equation}
\end{itemize}
Therefore, it follows from expression \eqref{eq:nckqr-kkt1-S0} that
\begin{equation*}
  \left\{
    \begin{array}{l}
     \mathbf{0}
      \in-\frac1n\sum_{i=1}^{n}\partial\rho_{\tau}
      (y_{i}-\hat{b}^{\gamma}_{\tau_t}-\mathbf{K}_{i}^{\top}\hat{\bs{\alpha}}^{\gamma}_{\tau_t}
      )\mathbf{K}_{i}+ \lambda_2\mathbf{K}\hat{\alpha}^{\gamma}_{\tau_t}+\sum_{i\in S_{0,t}}
      \xi_{i,t}\mathbf{K}_{i}\\
      \quad \quad +\lambda_1 \sum_{i=1}^{n} V^{\prime}(\hat{b}^{\gamma}_{\tau_t}
      +\mathbf{K}_{i}^{\top}\hat{\bs{\alpha}}^{\gamma}_{\tau_t}-\hat{b}^{\gamma}_{\tau_{t+1}}
      -\mathbf{K}_{i}^{\top}\hat{\bs{\alpha}}^{\gamma}_{\tau_{t+1}})\mathbf{K}_{i},\\

      0\in-\frac{1}{n}\sum_{i=1}^{n}\partial \rho_{\tau}
      (y_{i}-\hat{b}^{\gamma}_{\tau_t}-\mathbf{K}_{i}^{\top}\hat{\bs{\alpha}}^{\gamma}_{\tau_t})\\
      \quad\quad+\lambda_1 \sum_{i=1}^{n} V^{\prime}(\hat{b}^{\gamma}_{\tau_{t}}+\mathbf{K}_{i}^{\top}
      \hat{\bs{\alpha}}^{\gamma}_{\tau_{t}}-\hat{b}^{\gamma}_{\tau_{t+1}}
      -\mathbf{K}_{i}^{\top}\hat{\bs{\alpha}}^{\gamma}_{\tau_{t+1}})
      +\sum_{i\in S_{0,t}}\xi_{i,t},\\
      y_{i}=\hat{b}^{\gamma}_{\tau_t}+\mathbf{K}_{i}^{\top}\hat{\bs{\alpha}}^{\gamma}_{\tau_t}
      ,\,i\in S_{0,t},
    \end{array}
  \right.
\end{equation*}
Similarly, it can be shown for the rest two displays \eqref{eq:nckqr-kkt2-S0}
and \eqref{eq:nckqr-kkt3-S0} that
\begin{equation*}
  \left\{
    \begin{array}{l}
     \mathbf{0} \in
      -\frac{1}{n}\sum_{i=1}^{n}
      \partial \rho_{\tau}
      (y_{i}-\hat{b}^{\gamma}_{\tau_t}-\mathbf{K}_{i}^{\top}\hat{\bs{\alpha}}^{\gamma}_{\tau_t}
      )\mathbf{K}_{i}+\lambda_2\mathbf{K}
      \hat{\alpha}^{\gamma}_{\tau_t}
      +\sum_{i\in S_{0,t}}
      \xi_{i,t}\mathbf{K}_{i} \\
      \quad - \lambda_1 \sum_{i=1}^{n} V^{\prime}(\hat{b}^{\gamma}_{\tau_{t-1}}
      +\mathbf{K}_{i}^{\top}\hat{\bs{\alpha}}^{\gamma}_{\tau_{t-1}}
      -\hat{b}^{\gamma}_{\tau_t}-\mathbf{K}_{i}^{\top}\hat{\bs{\alpha}}^{\gamma}_{\tau_t})
      \mathbf{K}_{i}\\
      \quad+\lambda_1 \sum_{i=1}^{n} V^{\prime}
      (\hat{b}^{\gamma}_{\tau_{t}}+\mathbf{K}_{i}^{\top}
      \hat{\bs{\alpha}}^{\gamma}_{\tau_{t}}-\hat{b}^{\gamma}_{\tau_{t+1}}
      -\mathbf{K}_{i}^{\top}\hat{\bs{\alpha}}^{\gamma}_{\tau_{t+1}})\mathbf{K}_{i},\\
      0\in-\frac{1}{n}\sum_{i=1}^{n}\partial\rho_{\tau}
      (y_{i}-\hat{b}^{\gamma}_{\tau_t}-\mathbf{K}_{i}^{\top}
      \hat{\bs{\alpha}}^{\gamma}_{\tau_t})-\lambda_1 \sum_{i=1}^{n}
      V^{\prime}(\hat{b}^{\gamma}_{\tau_{t-1}}+\mathbf{K}_{i}^{\top}
      \hat{\bs{\alpha}}^{\gamma}_{\tau_{t-1}}-\hat{b}^{\gamma}_{\tau_{t}}
      -\mathbf{K}_{i}^{\top}\hat{\bs{\alpha}}^{\gamma}_{\tau_{t}})\\
      \quad+\lambda_1 \sum_{i=1}^{n} V^{\prime}(\hat{b}^{\gamma}_{\tau_{t}}
      +\mathbf{K}_{i}^{\top}\hat{\bs{\alpha}}^{\gamma}_{\tau_{t}}
      -\hat{b}^{\gamma}_{\tau_{t+1}}-\mathbf{K}_{i}^{\top}
      \hat{\bs{\alpha}}^{\gamma}_{\tau_{t+1}})+\sum_{i\in S_{0,t}}\xi_{i,t},\\
      y_{i}=\hat{b}^{\gamma}_{\tau_t}+\mathbf{K}_{i}^{\top}\hat{\bs{\alpha}}^{\gamma}_{\tau_t}
      ,i\in S_{0,t},
    \end{array}
  \right.
\end{equation*}
and
\begin{equation*}
  \left\{
    \begin{array}{l}
      \mathbf{0}\in -\frac{1}{n}\sum_{i=1}^{n} \partial \rho_{\tau}
      (y_{i}-\hat{b}^{\gamma}_{\tau_t}-\mathbf{K}_{i}^{\top}
      \hat{\bs{\alpha}}^{\gamma}_{\tau_t})\mathbf{K}_{i}+\lambda_2\mathbf{K}
      \hat{\alpha}^{\gamma}_{\tau_t}\\
      \quad \quad - \lambda_1 \sum_{i=1}^{n} V^{\prime}(\hat{b}^{\gamma}_{\tau_{t-1}}
      +\mathbf{K}_{i}^{\top}\hat{\bs{\alpha}}^{\gamma}_{\tau_{t-1}}-\hat{b}^{\gamma}_{\tau_t}
      -\mathbf{K}_{i}^{\top}\hat{\bs{\alpha}}^{\gamma}_{\tau_t})\mathbf{K}_{i}
      +\sum_{i\in S_{0,t}} \xi_{i,t}\mathbf{K}_{i}\\

      0\in -\frac{1}{n}\sum_{i=1}^{n} \partial \rho_{\tau}
      (y_{i}-\hat{b}^{\gamma}_{\tau_t}-\mathbf{K}_{i}^{\top}
      \hat{\bs{\alpha}}^{\gamma}_{\tau_t})\\
      \quad\quad-\lambda_1 \sum_{i=1}^{n}
      V^{\prime}(\hat{b}^{\gamma}_{\tau_{t-1}}+\mathbf{K}_{i}^{\top}
      \hat{\bs{\alpha}}^{\gamma}_{\tau_{t-1}}-\hat{b}^{\gamma}_{\tau_{t}}
      -\mathbf{K}_{i}^{\top}\hat{\bs{\alpha}}^{\gamma}_{\tau_{t}})
      +\sum_{i\in S_{0,t}} \xi_{i,t},\\
      y_{i}=\hat{b}^{\gamma}_{\tau_{t}}+\mathbf{K}_{i}^{\top}
      \hat{\bs{\alpha}}^{\gamma}_{\tau_{t}}, i \in S_{0,t}.
    \end{array}
  \right.
\end{equation*}
We can see that $\{\hat{b}^\gamma_{\tau_t},
\hat{\bs{\alpha}}^\gamma_{\tau_t}\}^T_{t=1}$ satisfies the KKT conditions of the
following convex problem:

\begin{equation}\label{eq:constr-nckqr-S0}
  \begin{split}
    \min_{\{b_{\tau_t}, \bs{\alpha}_{\tau_t}\}^T_{t=1}}\enskip{}
    & \sum^{T}_{t=1}\left[\frac1n\sum_{i=1}^{n}
      \rho_{\tau}(y_{i}-b_{\tau_t}-\mathbf{K}_{i}^{\top}\bs{\alpha}_{\tau_t})
      +\frac{\lambda_2}{2}\bs{\alpha}_{\tau_t}^{\top}\mathbf{K}\bs{\alpha}_{\tau_t}\right]\\
    & + \lambda_1 \sum^{T-1}_{t=1} V(b_{\tau_t}+\mathbf{K}_{i}^{\top}
      \bs{\alpha}_{\tau_t}-b_{\tau_{t+1}}-\mathbf{K}_{i}^{\top}\bs{\alpha}_{\tau_{t+1}})\\
    \text{subject to}\enskip{}
    & y_{i}=b_{\tau_t}+\mathbf{K}_{i}^{\top}\bs{\alpha}_{\tau_t},\,i\in S_{0,t}, t=1, \ldots, T.
  \end{split}
\end{equation}
Therefore, $\{\hat{b}^\gamma_{\tau_t},
\hat{\bs{\alpha}}^\gamma_{\tau_t}\}^T_{t=1}$ is the minimizer of problem
\eqref{eq:constr-nckqr-S0}. Moreover, since $\{\hat{b}_{\tau_t},
\hat{\bs{\alpha}}_{\tau_t}\}^T_{t=1}$ is a feasible point of
\eqref{eq:constr-nckqr-S0}, we have
\[
  Q\left(\{\hat{b}^\gamma_{\tau_t}, \hat{\bs{\alpha}}^\gamma_{\tau_t}\}^T_{t=1}\right)
  \leq{}Q\left(\{\hat{b}_{\tau_t}, \hat{\bs{\alpha}}_{\tau_t}\}^T_{t=1}\right)\leq{}
  Q\left(\{\hat{b}^\gamma_{\tau_t}, \hat{\bs{\alpha}}^\gamma_{\tau_t}\}^T_{t=1}\right),
\]
where the second inequality follows from the fact that $(\{\hat{b}_{\tau_t},
\hat{\bs{\alpha}}_{\tau_t}\}^T_{t=1})$ is the unconstrained minimizer of
$Q\left(\{b_{\tau_t}, \bs{\alpha}_{\tau_t}\}^T_{t=1}\right)$. Thus, we have
$Q\left(\{\hat{b}^\gamma_{\tau_t},
  \hat{\bs{\alpha}}^\gamma_{\tau_t}\}^T_{t=1}\right)=Q\left(\{\hat{b}_{\tau_t},
  \hat{\bs{\alpha}}_{\tau_t}\}^T_{t=1}\right)$ and by the uniqueness of
$\{\hat{b}_{\tau_t}, \hat{\bs{\alpha}}_{\tau_t}\}^T_{t=1}$, we obtain
$$
\{\hat{b}^\gamma_{\tau_t}, \hat{\bs{\alpha}}^\gamma_{\tau_t}\}^T_{t=1}
=\{\hat{b}_{\tau_t}, \hat{\bs{\alpha}}_{\tau_t}\}^T_{t=1},
$$
which completes the proof.

\subsection{Proof of Theorem \ref{thm:find-subset2}}

Since $S_t \subseteq{}S_{0, t}$, we have
$y_{i}=\hat{b}_{\tau_t}+\mathbf{K}_{i}^{\top}\hat{\bs{\alpha}}_{\tau_t}$ for
$i\in{}S_t$ and $1\leq t\leq T$. Therefore, $\{\hat{b}_{\tau_t},
\hat{\bs{\alpha}}_{\tau_t}\}^T_{t=1}$ is a feasible point of the minimization
problem
\[
  \begin{split}
    \min_{\{b_{\tau_t}, \bs{\alpha}_{\tau_t}\}^T_{t=1}}\enskip{}
    & Q^{\gamma}\left(\{b_{\tau_t}, \bs{\alpha}_{\tau_t}\}^T_{t=1}\right)\\
    \text{subject to}\enskip{}
    & y_{i}=b_{\tau_t}+\mathbf{K}_{i}^{\top}\bs{\alpha}_{\tau_t},i\in{}S_t,
      1\leq t\leq T.
  \end{split}
\]
By the optimality of $\{\tilde{b}^{\gamma}_{\tau_t},
\tilde{\bs{\alpha}}^{\gamma}_{\tau_t}\}^T_{t=1}$, we
have $$Q^{\gamma}\left(\{\tilde{b}^{\gamma}_{\tau_t},
  \tilde{\bs{\alpha}}^{\gamma}_{\tau_t}\}^T_{t=1}\right)\leq{}
Q^{\gamma}\left(\{\hat{b}_{\tau_t},
  \hat{\bs{\alpha}}_{\tau_t}\}^T_{t=1}\right).$$ It then follows from
\eqref{eq:huber-relu-approx} that
\begin{equation*}
  \begin{split}
    &Q\left(\{\tilde{b}^{\gamma}_{\tau_t}, \tilde{\bs{\alpha}}^{\gamma}_{\tau_t}\}^T_{t=1}\right)
      -Q\left(\{\hat{b}^{\gamma}_{\tau_t}, \hat{\bs{\alpha}}^{\gamma}_{\tau_t}\}^T_{t=1}\right) \\
    &=\left[Q\left(\{\tilde{b}^{\gamma}_{\tau_t}, \tilde{\bs{\alpha}}^{\gamma}_{\tau_t}\}^T_{t=1}\right)
      -Q^{\gamma}\left(\{\tilde{b}^{\gamma}_{\tau_t}, \tilde{\bs{\alpha}}^{\gamma}_{\tau_t}\}^T_{t=1}\right)\right]
      +\left[Q^{\gamma}\left(\{\tilde{b}^{\gamma}_{\tau_t}, \tilde{\bs{\alpha}}^{\gamma}_{\tau_t}\}^T_{t=1}\right)
      -Q^{\gamma}\left(\{\hat{b}^{\gamma}_{\tau_t}, \hat{\bs{\alpha}}^{\gamma}_{\tau_t}\}^T_{t=1}\right)\right] \\
    &\quad+\left[Q^{\gamma}\left(\{\hat{b}^{\gamma}_{\tau_t}, \hat{\bs{\alpha}}^{\gamma}_{\tau_t}\}^T_{t=1}\right)
      -Q\left(\{\hat{b}^{\gamma}_{\tau_t}, \hat{\bs{\alpha}}^{\gamma}_{\tau_t}\}^T_{t=1}\right)\right]\\
    &\leq 0+0+\frac{T}{4} \gamma<\frac{T}{4} \gamma^{*}\leq\rho.
  \end{split}
\end{equation*}
By the definition of $D_{\gamma_{0,t}/2}$, we know
$(\tilde{b}^{\gamma}_{\tau_t},
\tilde{\bs{\alpha}}^{\gamma}_{\tau_t})\notin{}D_{\gamma_{0,t}/2}$ for all
quantile levels, and therefore,
\[
  |\tilde{b}^{\gamma}_{\tau_t}+\mathbf{K}_{i}^{\top}\tilde{\bs{\alpha}}^{\gamma}_{\tau_t}
  -\hat{b}_{\tau_t}-\mathbf{K}_{i}^{\top}
  \hat{\bs{\alpha}}_{\tau_t}|<\frac{\gamma_{0,t}}{2},\forall{}i=1,\ldots,n, t=1, \ldots, T.
\]
This implies that for any $i\in \mathcal{E}_t(S_1,S_2,\cdots, S_T)$ and $1 \leq
t \leq T$,
\begin{equation*}
  \begin{split}
    &|y_{i}-\hat{b}_{\tau_t}-\mathbf{K}_{i}^{\top}\hat{\bs{\alpha}}_{\tau_t}|
    \leq{}
    |y_{i} -\tilde{b}^{\gamma}_{\tau_t}-\mathbf{K}_{i}^{\top}\tilde{\bs{\alpha}}^{\gamma}_{\tau_t}|
    +|\tilde{b}^{\gamma}_{\tau_t}+\mathbf{K}_{i}^{\top}
    \tilde{\bs{\alpha}}^{\gamma}_{\tau_t}
    -\hat{b}_{\tau_t}-\mathbf{K}_{i}^{\top}\hat{\bs{\alpha}}_{\tau_t}|\\
    &<\gamma+\frac{\gamma_{0,t}}{2}<\gamma^{*}+\frac{\gamma_{0,t}}{2}\leq\gamma_{0,t},
  \end{split}
\end{equation*}
which, by the definition of $\gamma_{0,t}$, further implies that
$i\in{}S_{0,t}$ . Thus, $S_t\subseteq\mathcal{E}_t(S_1,S_2,\cdots,
S_T)\subseteq{}S_{0,t}$.

\subsection{Proof of Theorem \ref{thm:subset2}}

Note that since $S_t\subseteq{}S_{0,t}$ and $\gamma\in(0,\gamma^{*})$, by
Theorem \ref{thm:find-subset2}, we have $S_t \subseteq\tilde{S}_{0,t}^{\gamma}
\subseteq{} S_{0,t}$ for all $t$. Similarly, for any $j\in\mathbb{N}^{+}$, if
$\tilde{S}_{j-1,t}^{\gamma}\subseteq{}S_{0,t}$, then by Theorem
\ref{thm:find-subset2} again, we have $\tilde{S}_{j-1,t}^{\gamma}\subseteq
\tilde{S}_{j,t}^{\gamma}\subseteq{}S_{0,t}$ for all $t$. Therefore, by
mathematical induction, we have
\[
  S_t \subseteq \tilde{S}_{0,t}^{\gamma} \subseteq \tilde{S}_{1,t}^{\gamma}
  \subseteq \tilde{S}_{2,t}^{\gamma} \subseteq\cdots\subseteq{}S_{0,t},
  \quad t=1, \ldots, T.
\]
Since $S_{0,t}$ is a finite set, there exists a finite $j^{*}\in\mathbb{N}^{+}$
such that $\tilde{S}_{j^{*}-1, t}^{\gamma}=\tilde{S}_{j^{*},t}^{\gamma}
{}\subseteq{}S_{0,t}.$ At this moment $\tilde{S}_{j^{*},t}^{\gamma}$ is the
target set $\hat{S}_t$.

Consider the following constrained problem
\begin{equation}\label{eq:constr-huber-relu}
  \begin{split}
    &\min_{\{b_{\tau_t}, \bs{\alpha}_{\tau_t}\}^T_{t=1}}\enskip{}
      Q^{\gamma}\left(\{b_{\tau_t}, \bs{\alpha}_{\tau_t}\}^T_{t=1}\right)\\
    & \text{subject to}\enskip{}
      y_{i}=b_{\tau_t}+\mathbf{K}_{i}^{\top}\bs{\alpha}_{\tau_t},\,i\in \hat{S}_t, 1\leq t\leq T.
  \end{split}
\end{equation}
The Lagrangian of problem \eqref{eq:constr-huber-relu} is
\begin{equation}
  \begin{split}
    L(b,\bs{\alpha},\xi)=
    & \sum^{T}_{t=1}\left[\frac{1}{n}\sum_{i=1}^{n}
      H_{\gamma, \tau}(y_{i}-b_{\tau_t}-\mathbf{K}_{i}^{\top}\bs{\alpha}_{\tau_t})
      +\frac{\lambda_2}{2}\bs{\alpha}^{\top}_{\tau_t}\mathbf{K}\bs{\alpha}_{\tau_t}\right]\\
    & + \sum^{T}_{t=1}\sum_{i\in \hat{S}_t^{\gamma}}\xi_{i,t}
      (b_{\tau_t}+\mathbf{K}_{i}^{\top}\bs{\alpha}_{\tau_t}-y_{i})\\
    &+ \lambda_1 \sum^{T-1}_{t=1}\sum_{i=1}^{n}
      V(b_{\tau_t}+\mathbf{K}_{i}^{\top}\bs{\alpha}_{\tau_t}
      -b_{\tau_{t+1}}-\mathbf{K}_{i}^{\top}\bs{\alpha}_{\tau_{t+1}}),
  \end{split}
\end{equation}
where the $\xi_{i,t}$'s are the Lagrangian multipliers. Consider the following three
cases, by the optimality of $\{\hat{b}^\gamma_{\tau_t}, \hat{\bs{\alpha}}^
\gamma_{\tau_t}\}^T_{t=1}$ in problem \eqref{eq:constr-huber-relu},
\begin{itemize}
\item Case 1. When $t=1$, we have
  \begin{equation}\label{eq:nckqr-kkt1}
    \left\{
      \begin{array}{l}
        -\frac{1}{n}\sum_{i=1}^{n}
        H^{\prime}_{\gamma,{}\tau}
        (y_{i}-\hat{b}^{\gamma}_{\tau_t}-\mathbf{K}_{i}^{\top}\hat{\bs{\alpha}}^{\gamma}_{\tau_t}
        )\mathbf{K}_{i}+\lambda_2\mathbf{K}
        \hat{\alpha}^{\gamma}_{\tau_t}+ \sum_{i\in \hat{S}_t}
        \xi_{i,t}\mathbf{K}_{i}\\
        \quad
        +\lambda_1 \sum_{i=1}^{n} V^{\prime}(\hat{b}^{\gamma}_{\tau_t}
        +\mathbf{K}_{i}^{\top}\hat{\bs{\alpha}}^{\gamma}_{\tau_t}
        -\hat{b}^{\gamma}_{\tau_{t+1}}-\mathbf{K}_{i}^{\top}
        \hat{\bs{\alpha}}^{\gamma}_{\tau_{t+1}})\mathbf{K}_{i}=\mathbf{0},\\
        -\frac{1}{n}\sum_{i=1}^{n} H^{\prime}_{\gamma,{}\tau}
        (y_{i}-\hat{b}^{\gamma}_{\tau_t}-\mathbf{K}_{i}^{\top}\hat{\bs{\alpha}}^{\gamma}_{\tau_t}
        ) \\
        \quad +
        \lambda_2 \sum_{i=1}^{n} V^{\prime}(\hat{b}^{\gamma}_{\tau_{t}}+\mathbf{K}_{i}^{\top}
        \hat{\bs{\alpha}}^{\gamma}_{\tau_{t}}-\hat{b}^{\gamma}_{\tau_{t+1}}
        -\mathbf{K}_{i}^{\top}\hat{\bs{\alpha}}^{\gamma}_{\tau_{t+1}})
        +\sum_{i\in\hat{S}_t} \xi_{i,t}=0,\\
        y_{i}=\hat{b}^{\gamma}_{\tau_t}+\mathbf{K}_{i}^{\top}\hat{\bs{\alpha}}^{\gamma}_{\tau_t},
        \enskip{}i\in\hat{S}_t.
      \end{array}
    \right.
  \end{equation}

\item Case 2. When $1<t<T$, we have

  \begin{equation}\label{eq:nckqr-kkt2}
    \left\{
      \begin{array}{l}
        -\frac{1}{n}\sum_{i=1}^{n}
        H^{\prime}_{\gamma,{}\tau}
        (y_{i}-\hat{b}^{\gamma}_{\tau_t}-\mathbf{K}_{i}^{\top}
        \hat{\bs{\alpha}}^{\gamma}_{\tau_t})\mathbf{K}_{i}+\lambda_2\mathbf{K}
        \hat{\alpha}^{\gamma}_{\tau_t} +\sum_{i\in \hat{S}_t} \xi_{i,t}\mathbf{K}_{i} \\
        \quad - \lambda_1 \sum_{i=1}^{n} V^{\prime}(\hat{b}^{\gamma}_{\tau_{t-1}}
        +\mathbf{K}_{i}^{\top}\hat{\bs{\alpha}}^{\gamma}_{\tau_{t-1}}
        -\hat{b}^{\gamma}_{\tau_t}-\mathbf{K}_{i}^{\top}
        \hat{\bs{\alpha}}^{\gamma}_{\tau_t})\mathbf{K}_{i} \\
        \quad +\lambda_1 \sum_{i=1}^{n} V^{\prime}(\hat{b}^{\gamma}_{\tau_{t}}
        +\mathbf{K}_{i}^{\top}\hat{\bs{\alpha}}^{\gamma}_{\tau_{t}}
        -\hat{b}^{\gamma}_{\tau_{t+1}}-\mathbf{K}_{i}^{\top}
        \hat{\bs{\alpha}}^{\gamma}_{\tau_{t+1}})\mathbf{K}_{i}=\mathbf{0},\\
        -\frac{1}{n}\sum_{i=1}^{n} H^{\prime}_{\gamma,{}\tau}
        (y_{i}-\hat{b}^{\gamma}_{\tau_t}-\mathbf{K}_{i}^{\top}\hat{\bs{\alpha}}^{\gamma}_{\tau_t})
        - \lambda_1 \sum_{i=1}^{n} V^{\prime}(\hat{b}^{\gamma}_{\tau_{t-1}}
        +\mathbf{K}_{i}^{\top}\hat{\bs{\alpha}}^{\gamma}_{\tau_{t-1}}
        -\hat{b}^{\gamma}_{\tau_{t}}-\mathbf{K}_{i}^{\top}\hat{\bs{\alpha}}^{\gamma}_{\tau_{t}}) \\
        \quad + \lambda_1 \sum_{i=1}^{n} V^{\prime}(\hat{b}^{\gamma}_{\tau_{t}}
        +\mathbf{K}_{i}^{\top}\hat{\bs{\alpha}}^{\gamma}_{\tau_{t}}
        -\hat{b}^{\gamma}_{\tau_{t+1}}-\mathbf{K}_{i}^{\top}\hat{\bs{\alpha}}^{\gamma}_{\tau_{t+1}})
        +\sum_{i\in\hat{S}_t} \xi_{i,t}=0\\
        y_{i}=\hat{b}^{\gamma}_{\tau_t}+ \mathbf{K}_{i}^{\top}\hat{\bs{\alpha}}^{\gamma}_{\tau_t},\,i\in\hat{S}_t.
      \end{array}
    \right.
  \end{equation}
\item When $t=T$, we have
  \begin{equation}\label{eq:nckqr-kkt3}
    \left\{
      \begin{array}{l}
        -\frac{1}{n}\sum_{i=1}^{n}
        H^{\prime}_{\gamma,{}\tau}
        (y_{i}-\hat{b}^{\gamma}_{\tau_t}-\mathbf{K}_{i}^{\top}
        \hat{\bs{\alpha}}^{\gamma}_{\tau_t})\mathbf{K}_{i}\\
        \quad +\lambda_2\mathbf{K}
        \hat{\alpha}^{\gamma}_{\tau_t}- \lambda_1 \sum_{i=1}^{n}
        V^{\prime}(\hat{b}^{\gamma}_{\tau_{t-1}}+\mathbf{K}_{i}^{\top}
        \hat{\bs{\alpha}}^{\gamma}_{\tau_{t-1}}-\hat{b}^{\gamma}_{\tau_t}
        -\mathbf{K}_{i}^{\top}\hat{\bs{\alpha}}^{\gamma}_{\tau_t})\mathbf{K}_{i}
        +\sum_{i\in \hat{S}_t} \xi_{l,i}\mathbf{K}_{i}=\mathbf{0},\\
        -\frac{1}{n}\sum_{i=1}^{n} H^{\prime}_{\gamma,{}\tau}
        (y_{i}-\hat{b}^{\gamma}_{\tau_t}-\mathbf{K}_{i}^{\top}
        \hat{\bs{\alpha}}^{\gamma}_{\tau_t})\\
        \quad - \lambda_1 \sum_{i=1}^{n} V^{\prime}(\hat{b}^{\gamma}_{\tau_{t-1}}
        +\mathbf{K}_{i}^{\top}\hat{\bs{\alpha}}^{\gamma}_{\tau_{t-1}}
        -\hat{b}^{\gamma}_{\tau_{t}}-\mathbf{K}_{i}^{\top}\hat{\bs{\alpha}}^{\gamma}_{\tau_{t}})
        +\sum_{i\in\hat{S}_t} \xi_{i,t}=0,\\
        y_{i}=\hat{b}^{\gamma}_{\tau_t}+ \mathbf{K}_{i}^{\top}\hat{\bs{\alpha}}^{\gamma}_{\tau_t},
        \,i\in\hat{S}_t.
      \end{array}
    \right.
  \end{equation}
\end{itemize}
Therefore, it follows from expression \eqref{eq:nckqr-kkt1} that
\begin{equation}
  \left\{
    \begin{array}{l}
      \mathbf{0}=-\frac1n\sum_{i\notin\hat{S}_t}
      H'_{\gamma,\tau}(y_{i}-\hat{b}^{\gamma}_{\tau_t}-\mathbf{K}_{i}^{\top}
      \hat{\bs{\alpha}}^{\gamma}_{\tau_t})\mathbf{K}_{i} -\frac1n\sum_{i\in\hat{S}_t}
      H'_{\gamma,\tau}(y_{i}-\hat{b}^{\gamma}_{\tau_t}
      -\mathbf{K}_{i}^{\top}\hat{\bs{\alpha}}^{\gamma}_{\tau_t})\mathbf{K}_{i} \\
      \quad \quad +\lambda_2\mathbf{K}\hat{\alpha}^{\gamma}_{\tau_t}+
      \sum_{i\in\hat{S}_t} \xi_{i,t}\mathbf{K}_{i}
      + \lambda_1 \sum_{i=1}^{n} V^{\prime}(\hat{b}^{\gamma}_{\tau_t}
      +\mathbf{K}_{i}^{\top}\hat{\bs{\alpha}}^{\gamma}_{\tau_t}
      -\hat{b}^{\gamma}_{\tau_{t+1}}-\mathbf{K}_{i}^{\top}
      \hat{\bs{\alpha}}^{\gamma}_{\tau_{t+1}})\mathbf{K}_{i}  \\
      \quad \in-\frac1n\sum_{i\notin \hat{S}_t}
      \partial\rho_{\tau}(y_{i}-\hat{b}^{\gamma}_{\tau_t}
      -\mathbf{K}_{i}^{\top}\hat{\bs{\alpha}}^{\gamma}_{\tau_t})
      \mathbf{K}_{i}-\frac1n\sum_{i\in\hat{S}_t}
      \partial\rho_{\tau}(y_{i}-\hat{b}^{\gamma}_{\tau_t}
      -\mathbf{K}_{i}^{\top}\hat{\bs{\alpha}}^{\gamma}_{\tau_t})\mathbf{K}_{i}\\
      \quad \quad +\lambda_2\mathbf{K}
      \hat{\alpha}^{\gamma}_{\tau_t}+\sum_{i\in\hat{S}_t}
      \xi_{i,t}\mathbf{K}_{i} + \lambda_1 \sum_{i=1}^{n}
      V^{\prime}(\hat{b}^{\gamma}_{\tau_t}+\mathbf{K}_{i}^{\top}
      \hat{\bs{\alpha}}^{\gamma}_{\tau_t}-\hat{b}^{\gamma}_{\tau_{t+1}}
      -\mathbf{K}_{i}^{\top}\hat{\bs{\alpha}}^{\gamma}_{\tau_{t+1}})\mathbf{K}_{i}  \\
      \quad =-\frac1n\sum_{i=1}^{n}\partial\rho_{\tau}
      (y_{i}-\hat{b}^{\gamma}_{\tau_t}-\mathbf{K}_{i}^{\top}
      \hat{\bs{\alpha}}^{\gamma}_{\tau_t})\mathbf{K}_{i}
      + \lambda_2\mathbf{K}
      \hat{\alpha}^{\gamma}_{\tau_t}+\sum_{i\in\hat{S}_t}
      \xi_{i,t}\mathbf{K}_{i}\\
      \quad \quad +\lambda_1 \sum_{i=1}^{n}
      V^{\prime}(\hat{b}^{\gamma}_{\tau_t}+\mathbf{K}_{i}^{\top}
      \hat{\bs{\alpha}}^{\gamma}_{\tau_t}-\hat{b}^{\gamma}_{\tau_{t+1}}
      -\mathbf{K}_{i}^{\top}\hat{\bs{\alpha}}^{\gamma}_{\tau_{t+1}})\mathbf{K}_{i},\\

      0=-\frac{1}{n}\sum_{i=1}^{n} H^{\prime}_{\gamma,{}\tau}
      (y_{i}-\hat{b}^{\gamma}_{\tau_t}-\mathbf{K}_{i}^{\top}
      \hat{\bs{\alpha}}^{\gamma}_{\tau_t})\\
      \quad\quad+ \lambda_1 \sum_{i=1}^{n}
      V^{\prime}(\hat{b}^{\gamma}_{\tau_{t}}+\mathbf{K}_{i}^{\top}
      \hat{\bs{\alpha}}^{\gamma}_{\tau_{t}}-\hat{b}^{\gamma}_{\tau_{t+1}}
      -\mathbf{K}_{i}^{\top}\hat{\bs{\alpha}}^{\gamma}_{\tau_{t+1}})
      +\sum_{i\in\hat{S}_t} \xi_{i,t} \\
      \quad \in -\frac{1}{n}\sum_{i=1}^{n} \partial \rho_{\tau}
      (y_{i}-\hat{b}^{\gamma}_{\tau_t}-\mathbf{K}_{i}^{\top}
      \hat{\bs{\alpha}}^{\gamma}_{\tau_t})\\
      \quad\quad+ \lambda_1 \sum_{i=1}^{n}
      V^{\prime}(\hat{b}^{\gamma}_{\tau_{t}}+\mathbf{K}_{i}^{\top}
      \hat{\bs{\alpha}}^{\gamma}_{\tau_{t}}-\hat{b}^{\gamma}_{\tau_{t+1}}
      -\mathbf{K}_{i}^{\top}\hat{\bs{\alpha}}^{\gamma}_{\tau_{t+1}})
      +\sum_{i\in\hat{S}_t} \xi_{i,t},\\
      y_{i}=\hat{b}^{\gamma}_{\tau_t}+\mathbf{K}_{i}^{\top}\hat{\bs{\alpha}}^{\gamma}_{\tau_t}
      ,\,i\in\hat{S}_t,
    \end{array}
  \right.
\end{equation}
Similarly, it can be shown for expressions~\eqref{eq:nckqr-kkt2} and
\eqref{eq:nckqr-kkt3}, we have
\begin{equation}
  \left\{
    \begin{array}{l}
      \mathbf{0} = -\frac{1}{n}\sum_{i=1}^{n}
      H^{\prime}_{\gamma,{}\tau}
      (y_{i}-\hat{b}^{\gamma}_{\tau_t}-\mathbf{K}_{i}^{\top}\hat{\bs{\alpha}}^{\gamma}_{\tau_t}
      )\mathbf{K}_{i}+\lambda_2\mathbf{K}
      \hat{\alpha}^{\gamma}_{\tau_t}
      +\sum_{i\in \hat{S}_t}
      \xi_{i,t}\mathbf{K}_{i} \\
      \quad\quad - \lambda_1 \sum_{i=1}^{n} V^{\prime}(\hat{b}^{\gamma}_{\tau_{t-1}}
      +\mathbf{K}_{i}^{\top}\hat{\bs{\alpha}}^{\gamma}_{\tau_{t-1}}
      -\hat{b}^{\gamma}_{\tau_t}-\mathbf{K}_{i}^{\top}
      \hat{\bs{\alpha}}^{\gamma}_{\tau_t})\mathbf{K}_{i}\\
      \quad\quad+\lambda_1 \sum_{i=1}^{n} V^{\prime}(\hat{b}^{\gamma}_{\tau_{t}}
      +\mathbf{K}_{i}^{\top}\hat{\bs{\alpha}}^{\gamma}_{\tau_{t}}
      -\hat{b}^{\gamma}_{\tau_{t+1}}-\mathbf{K}_{i}^{\top}
      \hat{\bs{\alpha}}^{\gamma}_{\tau_{t+1}})\mathbf{K}_{i} \\
      \quad \in -\frac{1}{n}\sum_{i=1}^{n} \partial \rho_{\tau}
      (y_{i}-\hat{b}^{\gamma}_{\tau_t}-\mathbf{K}_{i}^{\top}
      \hat{\bs{\alpha}}^{\gamma}_{\tau_t})\mathbf{K}_{i}+\lambda_2\mathbf{K}
      \hat{\alpha}^{\gamma}_{\tau_t} +\sum_{i\in \hat{S}_t} \xi_{i,t}\mathbf{K}_{i} \\
      \quad\quad - \lambda_1 \sum_{i=1}^{n} V^{\prime}(\hat{b}^{\gamma}_{\tau_{t-1}}
      +\mathbf{K}_{i}^{\top}\hat{\bs{\alpha}}^{\gamma}_{\tau_{t-1}}
      -\hat{b}^{\gamma}_{\tau_t}-\mathbf{K}_{i}^{\top}
      \hat{\bs{\alpha}}^{\gamma}_{\tau_t})\mathbf{K}_{i}\\
      \quad\quad+\lambda_1 \sum_{i=1}^{n} V^{\prime}(\hat{b}^{\gamma}_{\tau_{t}}
      +\mathbf{K}_{i}^{\top}\hat{\bs{\alpha}}^{\gamma}_{\tau_{t}}
      -\hat{b}^{\gamma}_{\tau_{t+1}}-\mathbf{K}_{i}^{\top}
      \hat{\bs{\alpha}}^{\gamma}_{\tau_{t+1}})\mathbf{K}_{i},\\

      0= -\frac{1}{n}\sum_{i=1}^{n}
      H^{\prime}_{\gamma,{}\tau}
      (y_{i}-\hat{b}^{\gamma}_{\tau_t}-\mathbf{K}_{i}^{\top}
      \hat{\bs{\alpha}}^{\gamma}_{\tau_t})-\lambda_1 \sum_{i=1}^{n}
      V^{\prime}(\hat{b}^{\gamma}_{\tau_{t-1}}+\mathbf{K}_{i}^{\top}
      \hat{\bs{\alpha}}^{\gamma}_{\tau_{t-1}}-\hat{b}^{\gamma}_{\tau_{t}}
      -\mathbf{K}_{i}^{\top}\hat{\bs{\alpha}}^{\gamma}_{\tau_{t}})\\
      \quad\quad + \lambda_1 \sum_{i=1}^{n} V^{\prime}(\hat{b}^{\gamma}_{\tau_{t}}
      +\mathbf{K}_{i}^{\top}\hat{\bs{\alpha}}^{\gamma}_{\tau_{t}}
      -\hat{b}^{\gamma}_{\tau_{t+1}}-\mathbf{K}_{i}^{\top}\hat{\bs{\alpha}}^{\gamma}_{\tau_{t+1}})
      +\sum_{i\in\hat{S}_t} \xi_{i,t} \\
      \quad \in -\frac{1}{n}\sum_{i=1}^{n} \partial \rho_{\tau}
      (y_{i}-\hat{b}^{\gamma}_{\tau_t}-\mathbf{K}_{i}^{\top}
      \hat{\bs{\alpha}}^{\gamma}_{\tau_t})- \lambda_1 \sum_{i=1}^{n}
      V^{\prime}(\hat{b}^{\gamma}_{\tau_{t-1}}+\mathbf{K}_{i}^{\top}
      \hat{\bs{\alpha}}^{\gamma}_{\tau_{t-1}}-\hat{b}^{\gamma}_{\tau_{t}}
      -\mathbf{K}_{i}^{\top}\hat{\bs{\alpha}}^{\gamma}_{\tau_{t}})\\
      \quad\quad + \lambda_1 \sum_{i=1}^{n} V^{\prime}(\hat{b}^{\gamma}_{\tau_{t}}
      +\mathbf{K}_{i}^{\top}\hat{\bs{\alpha}}^{\gamma}_{\tau_{t}}
      -\hat{b}^{\gamma}_{\tau_{t+1}}-\mathbf{K}_{i}^{\top}
      \hat{\bs{\alpha}}^{\gamma}_{\tau_{t+1}}) +\sum_{i\in\hat{S}_t} \xi_{i,t},\\
      y_{i}=\hat{b}^{\gamma}_{\tau_t}+\mathbf{K}_{i}^{\top}\hat{\bs{\alpha}}^{\gamma}_{\tau_t}
      ,i\in\hat{S}_t,
    \end{array}
  \right.
\end{equation}
and
\begin{equation}
  \left\{
    \begin{array}{l}
      \mathbf{0}=-\frac{1}{n}\sum_{i=1}^{n}
      H^{\prime}_{\gamma,{}\tau}
      (y_{i}-\hat{b}^{\gamma}_{\tau_t}-\mathbf{K}_{i}^{\top}\hat{\bs{\alpha}}^{\gamma}_{\tau_t}
      )\mathbf{K}_{i}+\lambda_2\mathbf{K}
      \hat{\alpha}^{\gamma}_{\tau_t}\\
      \quad \quad - \lambda_1 \sum_{i=1}^{n} V^{\prime}(\hat{b}^{\gamma}_{\tau_{t-1}}
      +\mathbf{K}_{i}^{\top}\hat{\bs{\alpha}}^{\gamma}_{\tau_{t-1}}
      -\hat{b}^{\gamma}_{\tau_t}-\mathbf{K}_{i}^{\top}\hat{\bs{\alpha}}^{\gamma}_{\tau_t})\mathbf{K}_{i}
      +\sum_{i\in \hat{S}_t} \xi_{i,t}\mathbf{K}_{i}\\
      \quad \in -\frac{1}{n}\sum_{i=1}^{n} \partial \rho_{\tau}
      (y_{i}-\hat{b}^{\gamma}_{\tau_t}-\mathbf{K}_{i}^{\top}\hat{\bs{\alpha}}^{\gamma}_{\tau_t}
      )\mathbf{K}_{i}+\lambda_2\mathbf{K}
      \hat{\alpha}^{\gamma}_{\tau_t}\\
      \quad \quad - \lambda_1 \sum_{i=1}^{n} V^{\prime}(\hat{b}^{\gamma}_{\tau_{t-1}}
      +\mathbf{K}_{i}^{\top}\hat{\bs{\alpha}}^{\gamma}_{\tau_{t-1}}-\hat{b}^{\gamma}_{\tau_t}
      -\mathbf{K}_{i}^{\top}\hat{\bs{\alpha}}^{\gamma}_{\tau_t})\mathbf{K}_{i}
      +\sum_{i\in \hat{S}_t} \xi_{i,t}\mathbf{K}_{i}\\

      0=-\frac{1}{n}\sum_{i=1}^{n} H^{\prime}_{\gamma,{}\tau}
      (y_{i}-\hat{b}^{\gamma}_{\tau_t}-\mathbf{K}_{i}^{\top}
      \hat{\bs{\alpha}}^{\gamma}_{\tau_t})\\
      \quad\quad-\lambda_1 \sum_{i=1}^{n}
      V^{\prime}(\hat{b}^{\gamma}_{\tau_{t-1}}+\mathbf{K}_{i}^{\top}
      \hat{\bs{\alpha}}^{\gamma}_{\tau_{t-1}}-\hat{b}^{\gamma}_{\tau_{t}}
      -\mathbf{K}_{i}^{\top}\hat{\bs{\alpha}}^{\gamma}_{\tau_{t}})
      +\sum_{i\in\hat{S}_t} \xi_{i,t} \\
      \quad \in -\frac{1}{n}\sum_{i=1}^{n} \partial \rho_{\tau}
      (y_{i}-\hat{b}^{\gamma}_{\tau_t}-\mathbf{K}_{i}^{\top}
      \hat{\bs{\alpha}}^{\gamma}_{\tau_t})\\
      \quad\quad-\lambda_1 \sum_{i=1}^{n}
      V^{\prime}(\hat{b}^{\gamma}_{\tau_{t-1}}+\mathbf{K}_{i}^{\top}
      \hat{\bs{\alpha}}^{\gamma}_{\tau_{t-1}}-\hat{b}^{\gamma}_{\tau_{t}}
      -\mathbf{K}_{i}^{\top}\hat{\bs{\alpha}}^{\gamma}_{\tau_{t}})
      +\sum_{i\in\hat{S}_t} \xi_{i,t},\\
      y_{i}=\hat{b}^{\gamma}_{\tau_{t}}+\mathbf{K}_{i}^{\top}
      \hat{\bs{\alpha}}^{\gamma}_{\tau_{t}}, i \in \hat{S}_t.
    \end{array}
  \right.
\end{equation}
Now consider the constrained problem
\begin{equation}\label{eq:constr-nckqr}
  \begin{split}
    \min_{\{b_{\tau_t}, \bs{\alpha}_{\tau_t}\}^T_{t=1}}\enskip{}
    & \sum^{T}_{t=1}\left[\frac1n\sum_{i=1}^{n}
      \rho_{\tau}(y_{i}-b_{\tau_t}-\mathbf{K}_{i}^{\top}\bs{\alpha}_{\tau_t})
      +\frac{\lambda_2}{2}\bs{\alpha}_{\tau_t}^{\top}
      \mathbf{K}\bs{\alpha}_{\tau_t}\right]\\
    & + \lambda_1 \sum^{T-1}_{t=1} V(b_{\tau_t}+\mathbf{K}_{i}^{\top}
      \bs{\alpha}_{\tau_t}-b_{\tau_{t+1}}-\mathbf{K}_{i}^{\top}\bs{\alpha}_{\tau_{t+1}})\\
    \text{subject to}\enskip{}
    & y_{i}=b_{\tau_t}+\mathbf{K}_{i}^{\top}\bs{\alpha}_{\tau_t},
      \,i\in\hat{S}_t, t=1, \ldots, T.
  \end{split}
\end{equation}
We see that $\{\hat{b}^\gamma_{\tau_t},
\hat{\bs{\alpha}}^\gamma_{\tau_t}\}^T_{t=1}$ satisfies the KKT conditions of the
convex problem \eqref{eq:constr-nckqr}. Therefore, $\{\hat{b}^\gamma_{\tau_t},
\hat{\bs{\alpha}}^\gamma_{\tau_t}\}^T_{t=1}$ is the minimizer of problem
\eqref{eq:constr-nckqr}. Moreover, it can be easily seen that
$\{\hat{b}_{\tau_t}, \hat{\bs{\alpha}}_{\tau_t}\}^T_{t=1}$ is a
feasible point of \eqref{eq:constr-nckqr} since $\hat{S}_t\subseteq{}S_{0}$.
This implies that
\[
  Q\left(\{\hat{b}^\gamma_{\tau_t}, \hat{\bs{\alpha}}^\gamma_{\tau_t}\}^T_{t=1}\right)
  \leq{}Q\left(\{\hat{b}_{\tau_t}, \hat{\bs{\alpha}}_{\tau_t}\}^T_{t=1}\right)\leq{}
  Q\left(\{\hat{b}^\gamma_{\tau_t}, \hat{\bs{\alpha}}^\gamma_{\tau_t}\}^T_{t=1}\right),
\]
where the second inequality follows from the fact that $(\{\hat{b}_{\tau_t},
\hat{\bs{\alpha}}_{\tau_t}\}^T_{t=1})$ is the unconstrained minimizer of
$Q\left(\{b_{\tau_t}, \bs{\alpha}_{\tau_t}\}^T_{t=1}\right)$. Thus, we have
$Q\left(\{\hat{b}^\gamma_{\tau_t},
  \hat{\bs{\alpha}}^\gamma_{\tau_t}\}^T_{t=1}\right)=Q\left(\{\hat{b}_{\tau_t},
  \hat{\bs{\alpha}}_{\tau_t}\}^T_{t=1}\right)$ and by the uniqueness of
$\{\hat{b}_{\tau_t}, \hat{\bs{\alpha}}_{\tau_t}\}^T_{t=1}$, we conclude that
$\{\hat{b}^\gamma_{\tau_t},
\hat{\bs{\alpha}}^\gamma_{\tau_t}\}^T_{t=1}=\{\hat{b}_{\tau_t},
\hat{\bs{\alpha}}_{\tau_t}\}^T_{t=1}$.

\end{document}